\documentclass{article} 
\usepackage{iclr2027_conference,times}

\usepackage{amsmath,amsfonts,bm}

\def\eqref#1{equation~\ref{#1}}

\def\1{\bm{1}}

\def\ra{{\textnormal{a}}}

\def\rx{{\textnormal{x}}}

\def\rva{{\mathbf{a}}}

\def\erva{{\textnormal{a}}}

\def\ervx{{\textnormal{x}}}

\def\rmA{{\mathbf{A}}}

\def\vmu{{\bm{\mu}}}
\def\vtheta{{\bm{\theta}}}
\def\va{{\bm{a}}}

\def\ve{{\bm{e}}}

\def\vx{{\bm{x}}}

\def\eva{{a}}

\def\mA{{\bm{A}}}

\def\mH{{\bm{H}}}
\def\mI{{\bm{I}}}
\def\mJ{{\bm{J}}}

\def\mX{{\bm{X}}}

\def\mSigma{{\bm{\Sigma}}}

\DeclareMathAlphabet{\mathsfit}{\encodingdefault}{\sfdefault}{m}{sl}
\SetMathAlphabet{\mathsfit}{bold}{\encodingdefault}{\sfdefault}{bx}{n}
\newcommand{\tens}[1]{\bm{\mathsfit{#1}}}
\def\tA{{\tens{A}}}

\def\tX{{\tens{X}}}

\def\gG{{\mathcal{G}}}

\def\sA{{\mathbb{A}}}
\def\sB{{\mathbb{B}}}

\def\sS{{\mathbb{S}}}

\def\emA{{A}}

\newcommand{\etens}[1]{\mathsfit{#1}}

\def\etA{{\etens{A}}}

\newcommand{\E}{\mathbb{E}}

\newcommand{\R}{\mathbb{R}}

\newcommand{\KL}{D_{\mathrm{KL}}}
\newcommand{\Var}{\mathrm{Var}}

\newcommand{\Cov}{\mathrm{Cov}}

\newcommand{\normltwo}{L^2}
\newcommand{\normlp}{L^p}

\newcommand{\parents}{Pa} 

\usepackage{graphicx}
\usepackage{booktabs}
\usepackage{overpic}
\usepackage[table]{xcolor}
\usepackage{amsfonts}
\usepackage{amsmath,amssymb}
\usepackage{amsthm}
\usepackage[dvipsnames]{xcolor}
\definecolor{orangehighlight}{RGB}{255, 230, 200} 
\usepackage{wrapfig}
\usepackage{graphicx}
\usepackage{subcaption}
\usepackage{mathtools}
\usepackage{enumitem}
\usepackage[utf8]{inputenc}
\usepackage[T1]{fontenc}
\usepackage{amsmath, amssymb}
\usepackage[skins]{tcolorbox}
\tcbuselibrary{skins}
\usepackage{graphicx}
\usepackage{forest}
\usepackage{makecell}

\usepackage[skins]{tcolorbox}
\usepackage{tikz}
\usetikzlibrary{arrows.meta} 
\usetikzlibrary{decorations.pathmorphing, backgrounds, calc, shapes.arrows}
\definecolor{elegantBlue}{RGB}{65, 105, 225} 
\definecolor{elegantOrange}{RGB}{237, 125, 49} 

\usepackage[utf8]{inputenc} 
\usepackage[T1]{fontenc}    
\usepackage{hyperref}       
\usepackage{url}            
\usepackage{booktabs}       
\usepackage{amsfonts}       
\usepackage{nicefrac}       
\usepackage{microtype}      
\usepackage{xcolor}         
\usepackage{tcolorbox}
\tcbuselibrary{skins} 

\title{BFMT: Enhancing Search Capabilities of Tree Sampler via \underline{B}ootstrap \underline{F}low-\underline{M}ap \underline{T}ree}

\author{
    Binglin Ji$^{1}$\thanks{Authors Made Equal and Core Contributions to This Paper.} ,~~~Anindya Sarkar$^{1}$\footnotemark[1] ,~~~~Hengchang Lu$^{1}$,~~~~Kunyu Wang$^{3}$,~~~~Alessandro Tibo$^{3}$\\~~~~\textbf{Jens Sjölund}$^{2}$,~~~~\textbf{Yevgeniy Vorobeychik}$^{1}$ \\
    \texttt{\{binglin.j,~anindya,~yvorobeychik\}@wustl.edu,}\\${}^1$Department of CSE, Washington University in St.Louis, USA\\
    ${}^2$Department of Information Technology, Uppsala University, Sweden\\
    ${}^3$Molecular AI, Discovery Sciences, R\&D, AstraZeneca, Gothenburg, Sweden}

\iclrfinalcopy 
\begin{document}

\maketitle

\begin{abstract}
The ability to efficiently navigate high-dimensional spaces to identify optimal candidates—whether designing targeted molecular structures or generating images with preferred attributes—stands as a central pillar of modern scientific progress. 
Recent tree-based samplers~\citep{jain2025diffusion} enable highly scalable inference-time search by eliminating the reward-gradient bottleneck inherent to particle-based methods.
Despite their potential, existing tree-based samplers are fundamentally constrained by the massive number of function evaluations (NFEs) required for node valuation, crippling their utility in exploration-heavy tasks. Furthermore, their reliance on small, uniform transitions at each depth precludes dynamic, adaptive search capabilities.
To address this, we introduce \textbf{B}ootstrap \textbf{F}low-\textbf{M}ap-\textbf{T}ree (a.k.a BFMT), a computationally efficient sampling framework that enables full tree-path construction from any tree depth using a single function evaluation, drastically reducing computational overhead while providing critical foresight for sequential sampling. By enabling dynamic transition time-step scheduling, BFMT efficiently allocates its sampling budget, smoothly transitioning from broad global exploration to fine-grained local refinement of high-utility modes discovered through exploration. Extensive experiments and ablation studies across diverse domains—from high-dimensional images to large-scale molecules—demonstrate BFMT's superiority over baseline approaches in both search and alignment tasks.
\end{abstract}
\vspace{-6pt}
\vspace{-4pt}
\section{Introduction}
\vspace{-8pt}
In many scientific and engineering applications, ranging from the discovery of life-saving therapeutics to the optimization of interactive visual recommendation systems, progress is often constrained by the high cost of evaluating candidate solutions. This creates a need for \emph{feedback-efficient search}, which requires uncovering high-quality targets through a limited sequence of expensive queries. For example, in personalized medicine, finding an effective treatment for a rare disease may involve iteratively proposing drug candidates and refining future choices using patient or experimental feedback. Since each evaluation can require substantial time, resources, and wet-lab experimentation, the objective is to uncover optimal targets with as few feedback rounds as possible. Addressing it demands a framework that can efficiently explore complex, high-dimensional spaces—such as molecular design or image spaces—while strategically harnessing sequential feedback to optimize the search process under a strict evaluation budget.

Current approaches are fundamentally ill-equipped for this challenge. For instance,~\citep{uehara2024feedback} relies on online RL fine-tuning of diffusion models to approximately minimize the KL divergence to the target distribution. This inherently mode-seeking objective fails to capture diverse, high-utility regions of the target distribution. Moreover, these methods hinge on an online-trained reward model, whose early-phase bias can misdirect the fine-tuning process and degrade sample quality. Recent Sequential Monte Carlo–based inference-time scaling methods~\citep{singhal2025general,skreta2025feynman,kim2025test} circumvent direct model fine-tuning by operating on fixed generative priors. However, they exhibit severe weight degeneracy, which effectively collapses the proposals and suppresses sample diversity~\citep{lee2025debiasing}. Moreover, these approaches rely on reward values or gradients at every resampling step; in practice, these signals are unavailable and must be approximated, introducing systematic bias into the sampler and degrading sample quality. 

To circumvent these barriers, recent tree-based samplers~\citep{guo2025training,jain2025diffusion} evaluate only terminal rewards, thereby obviating the need to compute the reward gradient at every intermediate denoising step. However, their practical utility remains severely hindered by the prohibitive number of function evaluations (NFEs) required for node valuation. Moreover, their reliance on uniform transitions at each tree depth stifles adaptive search, while their resource-blind exploration strategies make them unsuitable for budget-constrained applications. To overcome these bottlenecks, we propose a principled framework centered on the following question:
\vspace{-7pt}
\begin{tcolorbox}[colback=blue!5!white, colframe=gray!80!black]
\emph{How can we develop a principled, computationally efficient, training-free tree sampler that maximizes search efficiency while remaining highly effective under strict sampling budgets?}
\end{tcolorbox}
\vspace{-7pt}
To this end, we introduce the Bootstrapped Flow-Map Tree (BFMT), a novel sampling framework
\begin{wrapfigure}{r}{0.62\textwidth}
    \centering
    \includegraphics[width=\linewidth]{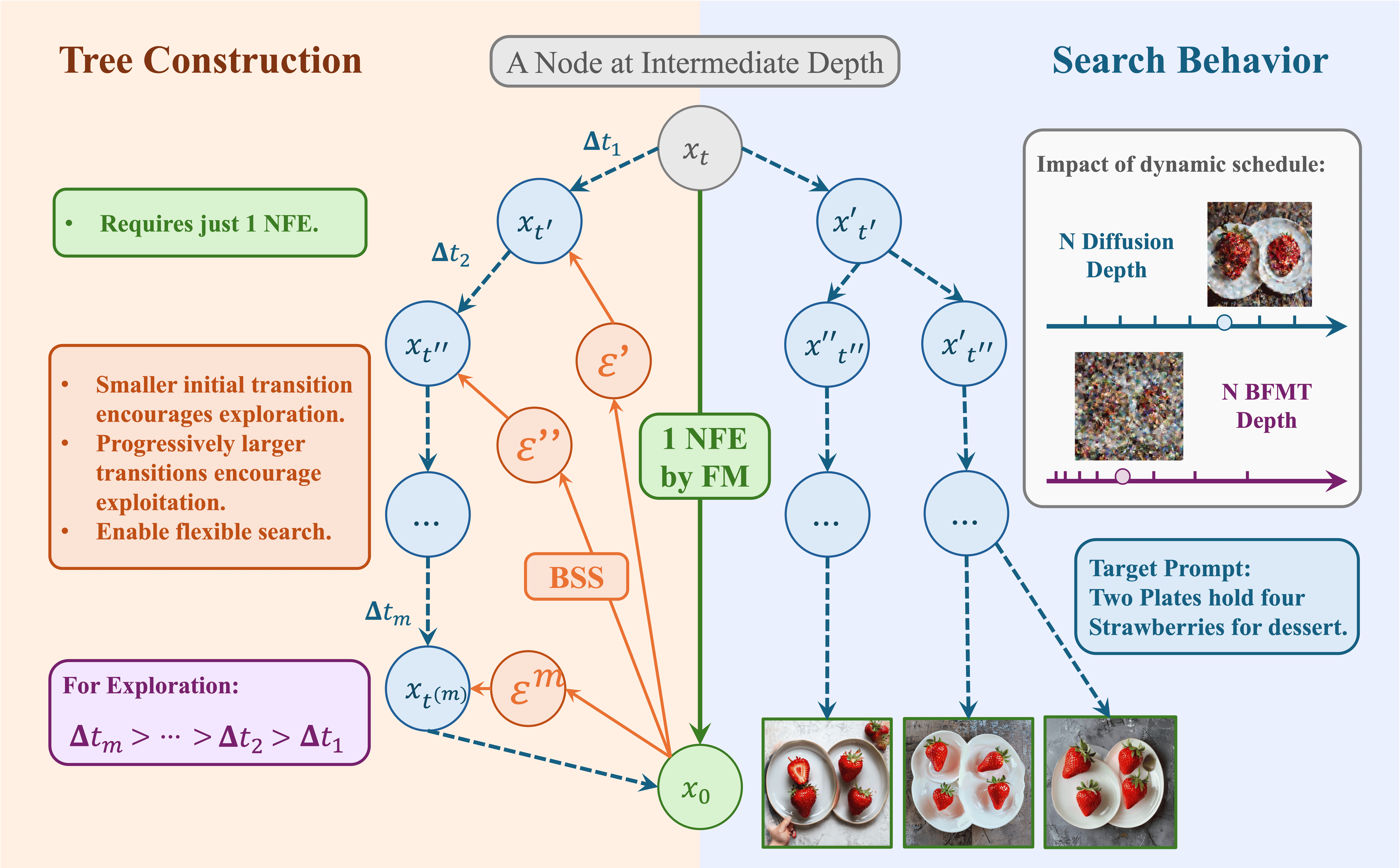}
    \caption{Overview of BFMT.}
    \label{fig:bfmt_trajectory}
     \vspace{-7pt}
\end{wrapfigure}
  that synergizes history-aware tree search with highly efficient Flow Map dynamics. While standard tree search requires computationally expensive rollouts to estimate intermediate node values, BFMT utilizes Flow Maps to collapse this evaluation into a single NFE. Although the deterministic ODE nature of Flow Maps typically bottlenecks the stochastic, DDPM-like transitions required for child-node sampling, we circumvent this by deploying a \emph{bootstrapped sufficient-statistic} scheme. This mechanism enables the construction of the \emph{entire stochastic tree path without incurring any additional NFEs}. Furthermore, because Flow Maps permit state estimation across arbitrary time intervals, BFMT accommodates non-uniform transition steps. Breaking from rigid diffusion schedules unlocks dynamic, granular control over the search process and the exploration-exploitation trade-off. Inherently, BFMT executes a smooth hierarchical search: early stages near the root prioritize broad global exploration to uncover diverse modes, while deeper layers progressively exploit local neighborhoods of discovered high-utility regions—guided by sequential feedback and Bellman backups—to maximize the overall search efficiency. Finally, we augment BFMT with a novel budget-aware node selection strategy that outperforms standard UCT in resource-constrained environments. We provide a conceptual depiction of the BFMT framework in Figure~\ref{fig:bfmt_trajectory}. We summarize our primary contributions as follows:

\vspace{-6pt}
\begin{tcolorbox}[colback=blue!5!white, colframe=gray!80!black, title={\textbf{Contributions}}]
\begin{itemize}[itemsep=1pt,topsep=0pt, leftmargin=*]
\item \textbf{Flow-Based Tree Sampler}: We introduce the Bootstrapped Flow-Map Tree, a principled, tree-based sampling framework for feedback-efficient search.
\item \textbf{Stochastic Path Construction with single NFE}: We formulate a bootstrapped sufficient-statistics scheme that synthesizes complete, DDPM-like stochastic tree trajectories from deterministic ODEs using only a single NFE.
\item \textbf{Dynamic Hierarchical Search}: By leveraging Flow Maps for non-uniform time transitions, BFMT unlocks adaptive search capabilities, seamlessly shifting from broad global mode exploration to fine-grained local refinement.
\item \textbf{Budget-Aware Node Selection}: We propose a custom budget-aware tree search strategy designed specifically for resource-constrained environments.  
\item \textbf{Rigorous Empirical Validation across Diverse Domains}: Comprehensive quantitative and qualitative ablations validate each component of BFMT across challenging tasks, from high-dimensional image synthesis to large-scale molecular design.
\end{itemize}
\end{tcolorbox}

\section{Methodology: BFMT Framework}
\textbf{Background:} Feedback-efficient search requires strategic exploration of the target landscape. This can be achieved by a policy that efficiently samples the target distribution:
\vspace{-3pt}
\begin{equation}
    \small{\pi^{*}(x) = \frac{1}{Z} p^{\text{pre}}_{\theta}(x) \exp(\beta \> r(x))}
    \label{eq:reward_tilted}
\end{equation}
In this formulation, $p^{\text{pre}}_{\theta}(x)$ denotes the pretrained generative prior, $r(x)$ represents the reward function, and $\beta$ is the inverse temperature parameter, with $Z$ serving as the normalizing constant. The target distribution can be seen as the optimal policy for the following objective~\citep{uehara2025inference}:
\vspace{-3pt}
\begin{equation}
    \small{\pi^*(x) := \text{arg\,max}_{\pi} \mathbb{E}_{x \sim \pi(\cdot)} [r(x)] - \frac{1}{\beta} D_{\text{KL}}(\pi \parallel p^{\text{pre}}_\theta)}
    \label{eq:optimal_policy}
\end{equation}
Efficiently optimizing $\pi^*(x)$ requires the estimation of the soft value function at time $t$~\citep{jain2025diffusion}, defined as:
\begin{equation}
    \small{V_t(x_t) := \frac{1}{\beta} \log \mathbb{E}_{\colorbox{blue!20}{$p^{\text{pre}}_{\theta}(x_{0:t-1}|x_t)$}} \left[ \exp \left( \beta r(x_0) \right) \right]}
    \label{eq:soft_value_func}
\end{equation}

Estimating the value of $x_t$ via Equation~\ref{eq:soft_value_func} introduces a computational bottleneck, as it necessitates simulating the entire denoising trajectory from $x_t$. Because these stochastic transition steps must be infinitesimal to minimize discretization error, they demand a prohibitively large number of transition steps and a correspondingly large NFE cost. 
Standard Tweedie-based posterior estimators are inadequate, as their pronounced inaccuracy at high noise levels critically undermines accurate value estimation. Moreover, a fundamental tension exists: while ODE sampling is highly efficient, enabling search strictly requires stochastic transitions. 
To resolve this conflict,~\citep{holderrieth2026diamond} introduce an ODE-based flow model that enables sampling from $x_0 \sim p(x_0|x_t)$ in a single step. 
Theoretically, given $x_t$, there exists an ODE that transports the prior $p_1$ to the conditional posterior $p_{0|t}(\cdot|x_t)$. The associated drift ${b}_s(\cdot;x_t)$ for this flow can be defined as the solution to the standard flow matching (FM) problem~\citep{lipman2022flow}, targeting the conditional posterior $p_{0|t}(\cdot|x_t)$ rather than the marginal data distribution $p_0$:
\begin{equation}
\small{{b}_s(x; x_t) = \mathbb{E}\left[\dot{\alpha}_sx_1 + \dot{\beta}_sx_0 \mid x_s = x\right], \quad x_s = \alpha_sx_1 + \beta_sx_0, \quad x_1 \sim \mathcal{N}(0, I_d) , \quad x_0 \sim p_{0|t}(\cdot|x_t)}
\end{equation}
Consequently, the probability flow associated with 
$b_s(\cdot; x_t)$ satisfies:
\vspace{-3pt}
\begin{equation}
\small{\frac{d}{ds}x_s = b_s(x_s; x_t), \quad x_1 \sim \mathcal{N}(0, I_d) \implies \text{Law}(x_0) = p_{0|t}(\cdot|x_t)}
\label{eq:ode_cond}
\end{equation}
Here $\alpha_t, \beta_t$ are time-dependent coefficients satisfying $\alpha_0 = \beta_1 = 0$ and $\alpha_1 = \beta_0 = 1$.
Assuming a Gaussian prior $p_1$, the conditional drift $b_s$ can be derived analytically~\citep{holderrieth2025glass}. This is achieved through a reparameterization of the unconditional drift $b_t$ of the pretrained flow model:
$$b_s(x_s; x_t) = w_1x_s + w_2 b_{t^*}(S(x_s, x_t)) + w_3 x_t$$
 Where, $w_1, w_2,$ and $w_3$ represent scalar weights. $S$ serves as a linear \emph{sufficient statistic}, and $t^*$ denotes a reparametrized time parameter:
\vspace{-7pt}
\begin{equation}
\small{S_{s,t}(x_s, x_t) = \frac{\alpha_s \sigma^2_t x_s + \alpha_t \sigma^2_s x_t}{\sigma^2_t \alpha^2_s + \alpha^2_t \sigma^2_s}, \qquad t^*(s, t) = g^{-1}\left( \frac{\sigma^2_t \sigma^2_s}{\sigma^2_t \alpha^2_s + \alpha^2_t \sigma^2_s} \right), \quad g(t) = \frac{\sigma^2_t}{\alpha^2_t}}.
\label{eq:combined_reparam}
\end{equation}
Here $\sigma_t$ represents the diffusion schedule. Although this reparameterization ensures the accessibility of $b_s$, the process of generating posterior samples from $p_{0|t}(\cdot|x_t)$ via the integration of ODE trajectories remains computationally demanding.
A standard approach to overcome this is teacher-distillation~\citep{boffi2026build}, in which the exact analytical field $b_s(x_s; x_t)$ serves as the teacher for the distillation process. The student model's velocity $\hat{v}$ is optimized using a combined objective, where the instantaneous loss directly minimizes the mean squared error between the student's instantaneous velocity, $\hat{v}_{s,s}$, and the analytical teacher field:
\vspace{-7pt}
\begin{equation}
\small{\mathcal{L}_{\text{inst}}^{\text{distill}}(\hat{v}) := \int_0^1 \int_0^1 \mathbb{E} ||\hat{v}_{s,s}(x; x_t) - b_s(x; x_t)||^2 ds dt}
\end{equation}
Coupled with a consistency loss ($\mathcal{L}^{\text{distill}}_{\text{cons}}$)~\citep{boffi2025build} that ensures the student accurately integrates this velocity over varying time intervals. $\mathcal{L}^{\text{distill}}_{\text{cons}}(\hat{v})$ is defined as follows:
\vspace{-6pt}
\[
    \small{\int_{0}^{1} \int_{0}^{u} \mathbb{E} \left\| \hat{v}_{s,u}(x_{s}; x'_t) - \text{sg}\left( b_u (\hat{X}_{s,u}(x_{s}; x'_t)); x'_t \right) - (u - s)\partial_u \hat{v}_{s,u}(x_{s}; x'_t) \right\|^2 ds \, du}
\]
Here, $\hat{X}_{s,u}(\cdot \, ; x_t) : \mathbb{R}^d \to \mathbb{R}^d$ acts as the solution operators for the context-dependent ODEs as in~\ref{eq:ode_cond}.
The distillation compresses the iterative multi-step ODE into a single amortized map, \emph{enabling computationally efficient, one-step sampling directly from the conditional posterior} $p_{0|t}(\cdot|x_t)$.

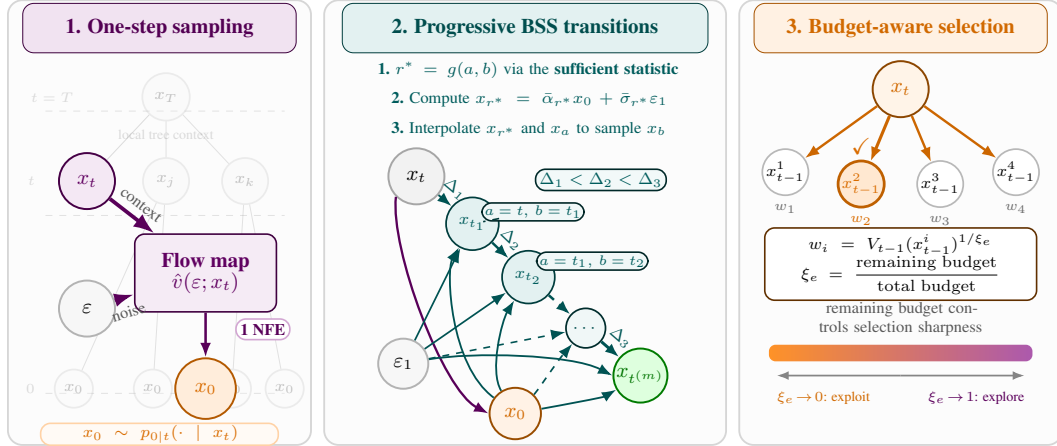
\begin{figure*}[t]
\resizebox{\textwidth}{!}{%
\begin{tikzpicture}[
    x=1cm,y=0.9cm,
    >=Latex,
    font=\footnotesize,
    panel/.style={draw=black!16, rounded corners=2.2mm, line width=0.8pt, fill=black!1},
    header/.style={rounded corners=1.7mm, minimum height=7.5mm, align=center,
                  font=\bfseries\footnotesize, inner sep=2pt},
    state/.style={circle, draw=black!35, fill=white, minimum size=8.5mm,
                  inner sep=0pt, line width=0.85pt, font=\bfseries\footnotesize},
    smallstate/.style={circle, draw=black!28, fill=white, minimum size=7mm,
                       inner sep=0pt, line width=0.75pt, font=\scriptsize},
    card/.style={draw=black!17, fill=white, rounded corners=1.5mm,
                 line width=0.7pt, align=center, inner sep=2.5pt},
    label/.style={font=\scriptsize, text=black!65, align=center},
    note/.style={font=\tiny, text=black!58, align=center},
    purplearrow/.style={-{Latex[length=2.2mm]}, violet!70!black, line width=1.1pt},
    tealarrow/.style={-{Latex[length=2.2mm]}, teal!65!black, line width=1.05pt},
    orangearrow/.style={-{Latex[length=2.2mm]}, orange!80!black, line width=1.05pt}
]

\draw[panel] (0.15,0.25) rectangle (4.75,7.75);
\draw[panel] (4.95,0.25) rectangle (11.05,7.75);
\draw[panel] (11.25,0.25) rectangle (16.15,7.75);

\node[header, draw=violet!48!black, fill=violet!11, text=violet!65!black,
      minimum width=4.15cm] at (2.45,7.28) {1. One-step sampling};
\node[header, draw=teal!48!black, fill=teal!11, text=teal!60!black,
      minimum width=5.55cm] at (8.00,7.28) {2. Progressive BSS transitions};
\node[header, draw=orange!65!black, fill=orange!11, text=orange!82!black,
      minimum width=4.45cm] at (13.70,7.28) {3. Budget-aware selection};

\begin{scope}[opacity=0.24]
\draw[black!45, dashed] (0.72,5.88) -- (4.35,5.88);
\draw[black!45, dashed] (0.72,4.12) -- (4.35,4.12);
\draw[black!45, dashed] (0.72,1.12) -- (4.35,1.12);

\node[smallstate, minimum size=7.2mm, fill=black!4] (tree-root)
      at (2.55,6.12) {$x_T$};
\node[smallstate, minimum size=7.0mm, fill=violet!5] (tree-left)
      at (1.35,4.70) {$x_t$};
\node[smallstate, minimum size=7.0mm, fill=black!3] (tree-mid)
      at (2.55,4.70) {$x_j$};
\node[smallstate, minimum size=7.0mm, fill=black!3] (tree-right)
      at (3.75,4.70) {$x_k$};

\node[smallstate, minimum size=5.8mm, fill=black!2] (tree-b1)
      at (1.15,1.20) {$x_0$};
\node[smallstate, minimum size=5.8mm, fill=black!2] (tree-b2)
      at (2.35,1.20) {$x_0$};
\node[smallstate, minimum size=5.8mm, fill=black!2] (tree-c1)
      at (3.55,1.20) {$x_0$};
\node[smallstate, minimum size=5.8mm, fill=black!2] (tree-c2)
      at (4.35,1.20) {$x_0$};

\draw[black!55] (tree-root) -- (tree-left);
\draw[black!45] (tree-root) -- (tree-mid);
\draw[black!45] (tree-root) -- (tree-right);
\draw[black!45] (tree-mid) -- (tree-b1);
\draw[black!45] (tree-mid) -- (tree-b2);
\draw[black!45] (tree-right) -- (tree-c1);
\draw[black!45] (tree-right) -- (tree-c2);

\node[note, anchor=west] at (0.38,6.12) {$t=T$};
\node[note, anchor=east] at (0.68,4.70) {$t$};
\node[note, anchor=east] at (0.68,1.20) {$0$};
\node[note] at (2.55,5.48) {local tree context};
\end{scope}

\node[state, draw=violet!55!black, fill=violet!9, text=violet!75!black]
      (xt) at (1.35,4.70) {$x_t$};
\node[state, draw=black!30, fill=black!3] (eps) at (1.35,2.55) {$\varepsilon$};
\node[card, draw=violet!50!black, fill=violet!8, text=violet!75!black,
      minimum width=2.15cm, minimum height=1.18cm] (flow) at (3.15,3.15)
      {\textbf{Flow map}\\[-1pt] $\hat v(\varepsilon;x_t)$};
\node[state, draw=orange!70!black, fill=orange!14, text=orange!85!black,
      minimum size=9.2mm] (xzero) at (3.15,1.20) {$x_0$};
\draw[purplearrow, line width=1.65pt,
      -{Latex[length=3.2mm,width=2.4mm]}]
      (xt) -- node[label, above, sloped] {context} (flow);
\draw[purplearrow, line width=1.45pt,
      -{Latex[length=3.0mm,width=2.1mm]}]
      (eps) -- node[label, below, sloped] {noise} (flow);
\draw[purplearrow] (flow) -- (xzero);
\node[card, draw=violet!35, fill=white, inner sep=2pt, text=violet!70!black]
      at (4.02,2.20) {\bfseries\scriptsize 1 NFE};
\node[card, draw=orange!35, fill=orange!4, text=orange!85!black,
      text width=3.55cm, minimum height=0.30cm, inner sep=0.8pt,
      font=\scriptsize] at (2.45,0.43)
      {$x_0\sim p_{0\mid t}(\cdot\mid x_t)$};

\node[label, anchor=west, text=teal!72!black, text width=5.30cm]
      at (5.28,6.58)
      {\textbf{1.}\ $r^*=g(a,b)$ via the \textbf{sufficient statistic}};
\node[label, anchor=west, text=teal!72!black, text width=5.30cm]
      at (5.28,6.08)
      {\textbf{2.}\ Compute
       $x_{r^*}=\bar\alpha_{r^*}x_0+
       \bar\sigma_{r^*}\varepsilon_1$};
\node[label, anchor=west, text=teal!72!black, text width=5.30cm]
      at (5.28,5.58)
      {\textbf{3.}\ Interpolate $x_{r^*}$ and $x_a$ to sample $x_b$};

\node[state, draw=black!35, fill=black!5] (bss-xt)
      at (6.35,4.78) {$x_t$};
\node[smallstate, draw=teal!48!black, fill=teal!11, text=teal!72!black,
      minimum size=8mm] (bss-t1) at (7.20,3.98) {$x_{t_1}$};
\node[smallstate, draw=teal!48!black, fill=teal!11, text=teal!72!black,
      minimum size=8mm] (bss-t2) at (8.05,3.08) {$x_{t_2}$};
\node[smallstate, draw=teal!35!black, fill=teal!6, text=teal!60!black,
      minimum size=6mm] (bss-dots) at (8.92,2.22) {$\cdots$};
\node[state, draw=green!48!black, fill=green!13, text=green!45!black,
      minimum size=8.2mm] (bss-final) at (9.75,1.42) {$x_{t^{(m)}}$};

\draw[tealarrow, line width=0.85pt] (bss-xt) --
      node[label, above, sloped, text=teal!68!black] {$\Delta_1$} (bss-t1);
\draw[tealarrow, line width=1.05pt] (bss-t1) --
      node[label, above, sloped, text=teal!68!black] {$\Delta_2$} (bss-t2);
\draw[tealarrow, dashed, line width=1.10pt] (bss-t2) -- (bss-dots);
\draw[tealarrow, line width=1.30pt] (bss-dots) --
      node[label, above, sloped, text=teal!68!black] {$\Delta_3$} (bss-final);
\node[card, draw=teal!32!black, fill=teal!5, text=teal!72!black,
      inner sep=1.7pt, font=\scriptsize] at (9.15,4.72)
      {$\Delta_1<\Delta_2<\Delta_3$};

\node[card, draw=teal!30!black, fill=teal!6, text=teal!72!black,
      inner sep=1.2pt, font=\tiny] (bss-ab1) at (8.12,4.18)
      {$a=t,\ b=t_1$};
\node[card, draw=teal!30!black, fill=teal!6, text=teal!72!black,
      inner sep=1.2pt, font=\tiny] (bss-ab2) at (9.02,3.35)
      {$a=t_1,\ b=t_2$};
\draw[teal!45!black, dashed, line width=0.7pt] (bss-t1) -- (bss-ab1);
\draw[teal!45!black, dashed, line width=0.7pt] (bss-t2) -- (bss-ab2);

\node[state, draw=orange!60!black, fill=orange!10, text=orange!82!black,
      minimum size=7.8mm] (bss-base) at (7.85,0.78) {$x_0$};
\node[state, draw=black!30, fill=black!3, minimum size=7.6mm]
      (bss-eps) at (6.15,1.62) {$\varepsilon_1$};
\draw[purplearrow, line width=1.0pt] (bss-xt.south west) .. controls
      (5.95,3.55) and (6.75,1.18) .. (bss-base.west);
\draw[tealarrow, line width=0.8pt] (bss-base) .. controls
      (6.85,1.60) and (6.80,3.20) .. (bss-t1);
\draw[tealarrow, line width=0.8pt] (bss-base) .. controls
      (7.45,1.50) and (7.65,2.35) .. (bss-t2);
\draw[tealarrow, dashed, line width=0.75pt] (bss-base) .. controls
      (8.18,1.15) and (8.48,1.65) .. (bss-dots);
\draw[tealarrow, line width=0.8pt] (bss-base) .. controls
      (8.55,0.92) and (9.18,1.05) .. (bss-final);
\draw[tealarrow, line width=0.75pt] (bss-eps) -- (bss-t1);
\draw[tealarrow, line width=0.75pt] (bss-eps) -- (bss-t2);
\draw[tealarrow, dashed, line width=0.70pt] (bss-eps) -- (bss-dots);
\draw[tealarrow, line width=0.80pt] (bss-eps) .. controls
      (7.20,1.82) and (8.70,1.78) .. (bss-final.west);

\node[state, draw=orange!55!black, fill=orange!9, text=orange!82!black]
      (root3) at (13.70,6.25) {$x_t$};
\node[smallstate] (c1) at (11.95,4.85) {$x_{t-1}^{1}$};
\node[smallstate, draw=orange!75!black, fill=orange!18, text=orange!85!black,
      line width=1.2pt] (c2) at (13.10,4.65) {$x_{t-1}^{2}$};
\node[smallstate] (c3) at (14.30,4.65) {$x_{t-1}^{3}$};
\node[smallstate] (c4) at (15.45,4.85) {$x_{t-1}^{4}$};
\draw[orangearrow] (root3) -- (c1);
\draw[orangearrow, line width=1.35pt] (root3) -- (c2);
\draw[orangearrow] (root3) -- (c3);
\draw[orangearrow] (root3) -- (c4);
\node[text=orange!85!black, font=\bfseries\normalsize] at (13.10,5.30)
      {$\checkmark$};
\node[note] at (11.95,4.24) {$w_1$};
\node[note, text=orange!85!black, font=\bfseries\tiny] at (13.10,4.04) {$w_2$};
\node[note] at (14.30,4.04) {$w_3$};
\node[note] at (15.45,4.24) {$w_4$};

\node[card, draw=orange!38!black, fill=white, text width=3.95cm,
      minimum height=0.82cm, font=\scriptsize] at (13.70,3.30)
      {$w_i=V_{t-1}(x_{t-1}^{i})^{1/\xi_e}$\\[-1pt]
       $\displaystyle \xi_e=
       \frac{\mathrm{remaining\ budget}}{\mathrm{total\ budget}}$};

\node[label, text width=4.0cm] at (13.70,2.38)
      {remaining budget controls selection sharpness};
\shade[left color=orange!85, right color=violet!65, rounded corners=1mm]
      (11.70,1.66) rectangle (15.70,1.94);
\draw[-{Latex[length=1.8mm]}, black!48, line width=0.75pt]
      (13.70,1.42) -- (11.82,1.42);
\draw[-{Latex[length=1.8mm]}, black!48, line width=0.75pt]
      (13.70,1.42) -- (15.58,1.42);
\node[note, anchor=west, text=orange!82!black] at (11.70,1.05)
      {$\xi_e\!\to\!0$: exploit};
\node[note, anchor=east, text=violet!72!black] at (15.70,1.05)
      {$\xi_e\!\to\!1$: explore};
\end{tikzpicture}%
}
\caption{Key components of BFMT.}
\label{fig:tree-steps-global}
\end{figure*}

\textbf{Proposed Approach:} Armed with the distilled flow-map model $\hat{v}(x;x_t)$, our next challenge is:

\begin{tcolorbox}[colback=blue!5!white, colframe=gray!80!black]
How can we exploit $\hat{v}(x;x_t)$ to orchestrate an entire sequence of Markovian stochastic transitions that synthesizes the full tree trajectory from $x_t$ to $x_0$ with just a single NFE?
\end{tcolorbox}
\vspace{-2pt}
As a first step, we directly generate $x_0$ from $x_t$ by sampling noise $\epsilon \sim \mathcal{N}(0, I)$ and passing it through the distilled flow map $\hat{v}(\epsilon; x_t)$. By definition, this operation yields exact samples from the true posterior, $x_0 \sim p_{0|t}(\cdot | x_t)$. Having anchored the trajectory's endpoints (i.e., $\epsilon$ and $x_0$), we realize the complete DDPM-like stochastic trajectory from $x_t$ using only a single NFE by our proposed Bootstrap sufficient statistics (BSS) approach.
 Concretely, given the initial state $x_t$ and the terminal prediction $x_0$ computed via $\hat{v}(\epsilon; x_t)$, we construct the trajectory recursively as follows:
\begin{equation}
\left.
\begin{aligned}
    \epsilon \sim N (0, I) \rightarrow  x_{r^\ast} &= \bar\alpha_{r^\ast} x_0 + \bar\sigma_{r^\ast} \varepsilon \rightarrow
    x_{t'} = \alpha_{t'} \frac{ \bar{\alpha}_{r^*} \sigma_t^2 x_{r^*} + \alpha_t \bar{\sigma}_{r^*}^2 x_t}{\bar{\alpha}_{r^*}^2 \sigma_t^2 + \alpha_t^2 \bar{\sigma}_{r^*}^2}, \\[6pt]
    r^*(t, t') &= g^{-1}\left( \frac{g(t)g(t')}{g(t) - g(t')} \right), \quad g(t) = \frac{\sigma^2_t}{\alpha^2_t}
\end{aligned}
\ \right\} 
\begin{array}{l}
    \textbf{Bootstrap Loop} \\
    \text{Update } t \leftarrow t' \\
    \text{until } t'=0
\end{array}
\label{eq:bss}
\end{equation}

$\bar{\alpha}$ and  $\bar{\sigma}^2$ are defined in the App.~\ref{sec:prop2-a}. Crucially, \emph{Equation~\ref{eq:bss} (i.e., BSS) imposes no restrictions on $t^{\prime}$, granting the flexibility to use dynamic transition steps for adaptive search.} In the following proposition, we show that this realized stochastic trajectory perfectly matches trajectories with DDPM-like transitions, theoretically guaranteeing convergence to the target distribution at the terminal step.
\begin{tcolorbox}[colback=blue!5!white, colframe=gray!80!black, title={\textbf{Proposition 4.1} (Bootstrap Sufficient Statistic Realizes DDPM-Like Stochastic Trajectory)}]
Let $0< t''<t' < t$ be any two consecutive transition time steps in a trajectory, and $p^{\text{DDPM}}_{t''|t'}$ be the DDPM transition kernel. Then for $x_{t''}$ obtained via~\ref{eq:bss}, starting from $x_t$, satisfies:
\begin{equation}
    x_{t''} \sim p^{\text{DDPM}}_{t''|t'}(\cdot \mid x_t)
\end{equation}
\end{tcolorbox}

Equipped with the distilled flow map, the proposed bootstrap sufficient statistic mechanism synthesizes a complete DDPM-like stochastic trajectory in a single NFE. Furthermore, the Markovian structure of this stochastic chain naturally induces a finite-horizon search tree over the $d$-dimensional diffusion space.
In this search tree, the nodes at depth $t$ represent noisy states $x_t$, and the edges represent a denoising step. Each node $x_t$ can be stochastically denoised into multiple children $x_{t-1} \sim \mathrm{BSS}^{1}\left( x_0 \sim p_{0|t}(\cdot \mid x_t) \right)$. $\mathrm{BSS}^1$ represents a single iteration of bootstrap sufficient statistic loop defined in~\ref{eq:bss}.
Consistent with standard tree samplers~\citep{jain2025diffusion}, each node maintains its state and timestep $(x_t, t)$, visit count $N(x_t)$, and a Monte Carlo estimate of the soft value function ${V}_t(x_t)$.
Analogous to the soft Bellman equation, ${V}_t(x_t)$ satisfies the following recursive relation:
\begin{equation}\label{eq:value}
V_t(x_t) = \frac{1}{\beta} \log \mathbb{E}_{x_{t-1} \sim \mathrm{BSS}^{1}\left( x_0 \sim p_{0|t}(\cdot \mid x_t) \right)} \left[ \exp (\beta \>\>\> V_{t-1}(x_{t-1})) \right].
\end{equation}
Moreover, ${V}_t(x_t)$ characterizes the optimal policy $\pi^{*} := \operatorname*{argmax}_{\pi} \mathbb{E}_{x \sim \pi(\cdot)} [r(x)] - \frac{1}{\beta} D_{\text{KL}} (\pi \parallel p_\theta)$:
\begin{equation}\label{eq:policy}
\small{\pi^*_t(x_{t-1} \mid x_t) = \frac{\mathrm{BSS}^{1}\left( x_0 \sim p_{0|t}(\cdot \mid x_t) \right) \exp (\beta V_{t-1}(x_{t-1}))}{\int \mathrm{BSS}^{1}\left( x_0 \sim p_{0|t}(\cdot \mid x_t) \right) \exp (\beta V_{t-1}(x_{t-1})) \, dx_{t-1}}.} 
\end{equation}
Here, $V_0(x_0) = r(x_0)$ is the terminal clean ground-truth reward feedback. We present the derivation of equation~\ref{eq:policy} in the Appendix~\ref{sec:proof-dts}. 

\paragraph{Budget-Aware Node Selection} However, choosing children solely by their soft value estimates starves the search of trajectories that reach diverse modes. Prior approaches~\citep{jain2025diffusion}, relying on Upper Confidence Tree~\citep{couetoux2011continuous}, are well-suited for asymptotic exploration but fundamentally assume unbounded horizons, failing to adapt to strict, depleting resource constraints. To address this and prevent catastrophic budget exhaustion on unpromising paths, we introduce a \emph{Budget-Aware Stochastic Exploration} mechanism ($\textrm{BASE}$) and sample each child (we use $\mathcal{C}(x_t)$ to denote the set of children of node $x_t$) according to the following normalized probability distribution:
\begin{equation}\label{eq:explore}
\small{\mathbf{p}^{{\textrm{BASE}}}(x_{t-1}) := \frac{w_{x_{t-1}}}{\sum_{x_{j} \in \mathcal{C}(x_t)} w_{x_j}}, \quad w_{x_{t-1}} = V_{t-1}(x_{t-1})^{\frac{1}{\zeta_t}}, \quad 
\zeta_t = \left(\frac{\text{\small{Remaining Budget}}}{\text{\small{Total Budget}}}\right)}
\end{equation}
Specifically, Starting from the root $x_1$, sample a child $x_{t-1} \in \mathcal{C}(x_t)$ following the $\mathbf{p}^{\textrm{BASE}}(\cdot)$ distribution recursively until either an unexpanded node is reached or $t = 0$.
$\textrm{BASE}$ approach shifts the search behavior based on the remaining budget. When resources are plentiful ($\zeta_t \approx 1$), the scaling exponent $\alpha_t = \frac{1}{\zeta_t}$ remains near $1$, creating a sampling distribution that follows the raw node values and encourages broad exploration. However, as the budget is depleted ($\zeta_t \to 0$), $\alpha_t$ grows, amplifying the differences between node values and focusing the probability on the most promising paths. Consequently, the agent transitions from an exploratory strategy to a strictly exploitative one as it nears its resource limit.

During each online interaction step, we traverse from the root to a terminal node by recursively applying the selection rule (Eqn.~\ref{eq:explore}). Upon sampling this terminal node, 
 we evaluate it via $V_{0}(x_0) = r(x_0)$, we recursively propagate this signal up the tree using a Monte-Carlo soft Bellman backup (Equation~\ref{eq:value}) and increment the visit counts for all nodes along the path. 
During tree traversal, if an intermediate node $x_t$ ($t > 0$) has not yet reached its branching capacity (i.e., $|\mathcal{C}(x_t)| < B(x_t) = \lceil C \cdot N(x_t)^\zeta \rceil, \quad C > 0, \zeta \in (0, 1)$), we actively expand it. Following \citep{couetoux2011continuous,jain2025diffusion}, to systematically exploit the most promising trajectories, we prioritize the expansion of highly visited nodes. We instantiate a new child $x_{t-1} \sim \mathrm{BSS}^{1}\left( x_0 \sim p_{0|t}(\cdot \mid x_t) \right)$ and initialize its statistics to $V_{t-1}(x_{t-1}) = 0$ and $N(x_{t-1}) = 1$. 
From each newly instantiated node, we execute a rollout to a terminal state $x_0$ via the recursive BSS loop (Eqn.~\ref{eq:bss}), seamlessly integrating this new path into the search tree. We term this framework BFMT, with its key features visualized in Fig.~\ref{fig:tree-steps-global}. 

As feedback accumulates, the generated samples via the BFMT approach strictly converge to the true target distribution, mirroring the behavior of the prior tree sampler.
We also show that the sampling error decays polynomially with respect to the total sampling budget. Crucially, \emph{BFMT’s efficiency hinges on its quadratic dependence of the convergence rate on the tree horizon: by realizing a long horizon with a single NFE or a single step, BFMT can substantially accelerate convergence.} 
\begin{tcolorbox}[colback=blue!5!white, colframe=gray!80!black, title={\textbf{Proposition 4.3} (BFMT Convergence Rate) (Informal)}]
Let $r(x_0)$ be bounded and $L$-Lipschitz continuous, $M$ be the feedback budget, and $T$ be the maximum tree depth. Under standard regularity assumptions, the Total Variation distance between the BFMT sampling policy $\hat{q}_M(x_0)$ and the target optimal policy $\pi^*(x_0)$ is bounded:$$D_{TV}(\hat{q}_M \parallel \pi^*) \le \mathcal{O}\left( \beta T^2 M^{-1/4} \right)$$
\end{tcolorbox}
\vspace{-5pt}
Importantly, BFMT seamlessly generalizes to domains lacking an off-the-shelf stochastic flow map by integrating different variants of distilled models, including standard deterministic flow maps—such as consistency models (detailed derivations in App.~\ref{sec:trigflow_adaptation}). Extensive experiments with state-of-the-art consistency models confirm BFMT's superior performance and broad applicability (see Results).
\vspace{-8pt}
\paragraph{Generalizability to Discrete Domain} 
The key motivation for extending inference-time search with BFMT samplers to the discrete domain is the elegant continuous relaxation introduced by Categorical Flow Maps~\citep{roos2026categorical} (CFMs). Rather than navigating the complexities of discrete Markov chains, CFMs model the generative process as a smooth, continuous probability path constrained directly on the probability simplex $\Delta^K$. The core "trick" relies on using a simplex-constrained endpoint parameterization. Because the entire flow trajectory unfolds in this continuous space, we can plug in continuous-domain inference-time search algorithms—like BFMT samplers—to categorical settings. The only modification required is a simple argmax projection at the very end of the generated path to yield the discrete output. Our empirical findings on a large-scale molecule generation task highlight BFMT's effectiveness for efficient search in the discrete domain.
\vspace{-12pt}
\section{Experiments and Results}
\begin{figure*}[h]
\centering
\begin{minipage}[t]{0.48\textwidth}
    \centering
    {\scriptsize\itshape "A dog in the snow."\par}
    \vspace{1ex}
    \begin{minipage}[t]{0.32\linewidth}
        \centering
        {\scriptsize\textbf{BFMT-ISI}}\par
        \IfFileExists{figs/bss_i.png}{\includegraphics[width=\linewidth]{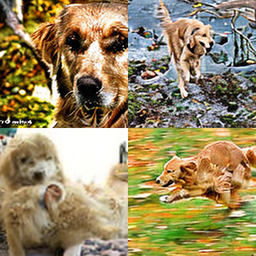}}{\fbox{\begin{minipage}[c][28mm][c]{\linewidth}\centering\scriptsize Add\\\detokenize{figs/bss_i.png}\end{minipage}}}%
    \end{minipage}\hfill
    \begin{minipage}[t]{0.32\linewidth}
        \centering
        {\scriptsize\textbf{BFMT}}\par
        \IfFileExists{figs/bss_s.png}{\includegraphics[width=\linewidth]{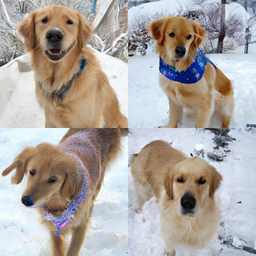}}{\fbox{\begin{minipage}[c][28mm][c]{\linewidth}\centering\scriptsize Add\\\detokenize{figs/bss_s.png}\end{minipage}}}%
    \end{minipage}\hfill
    \begin{minipage}[t]{0.32\linewidth}
        \centering
        {\scriptsize\textbf{Reward}}\par
        \IfFileExists{figs/BSS.png}{\includegraphics[width=\linewidth]{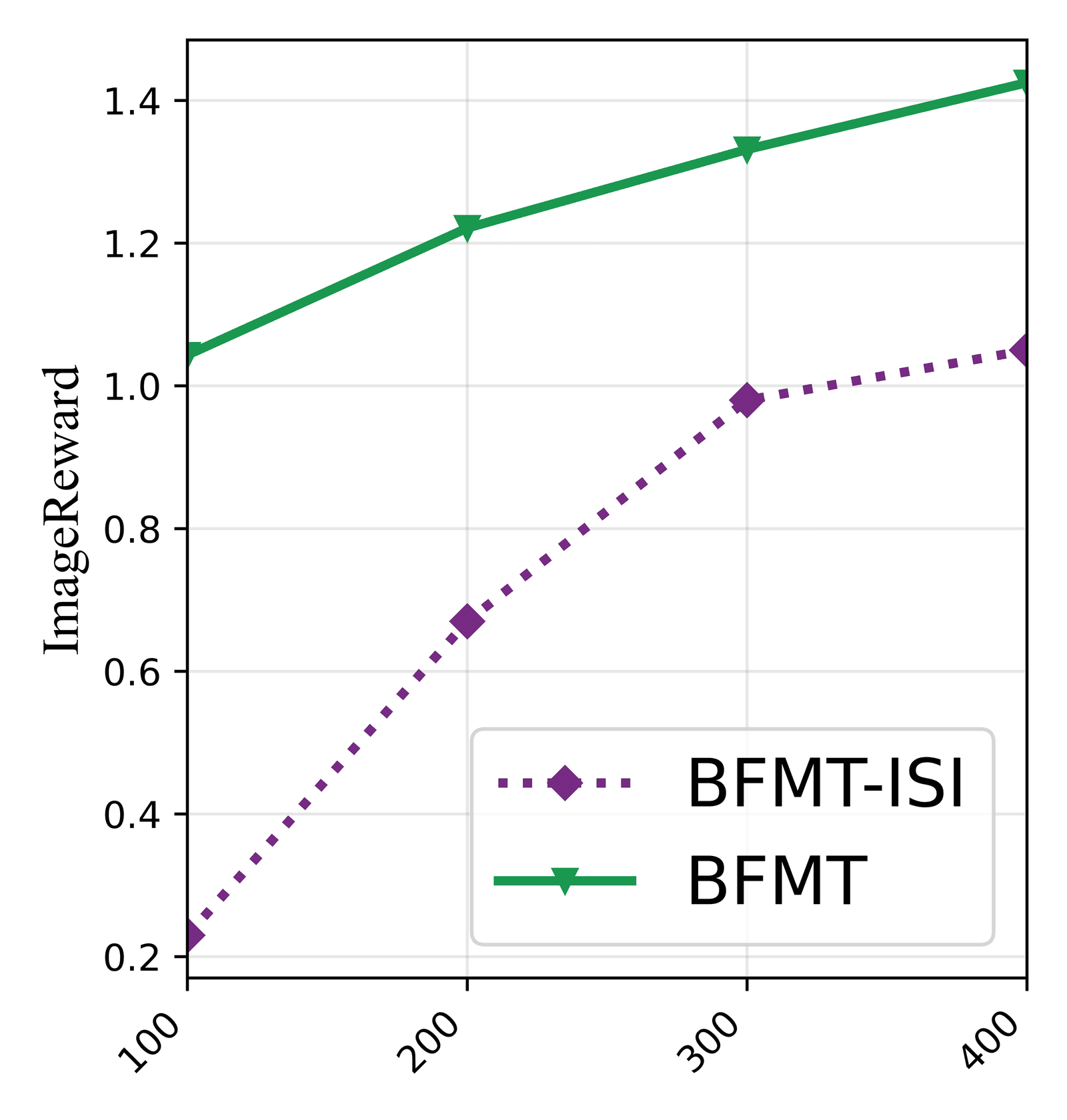}}{\fbox{\begin{minipage}[c][28mm][c]{\linewidth}\centering\scriptsize Add\\\detokenize{figs/BSS.png}\end{minipage}}}\par
        {\scriptsize\textbf{NFE}}
    \end{minipage}
    \vspace{-4pt}
    \captionof{figure}{\small{Impact of BSS.}}
    \label{fig:BSS}
\end{minipage}%
\hfill
\begin{minipage}[t]{0.48\textwidth}
    \centering
    {\scriptsize\itshape "A red sports car in the showroom"\par}
    \vspace{1ex}
    \begin{minipage}[t]{0.32\linewidth}
        \centering
        {\scriptsize\textbf{BFMT-US}}\par
        \IfFileExists{figs/SS.png}{\includegraphics[width=\linewidth]{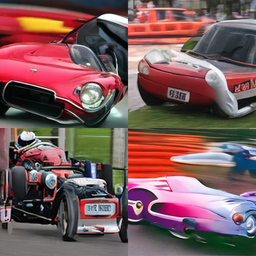}}{\fbox{\begin{minipage}[c][28mm][c]{\linewidth}\centering\scriptsize Add\\\detokenize{figs/SS.png}\end{minipage}}}%
    \end{minipage}\hfill
    \begin{minipage}[t]{0.32\linewidth}
        \centering
        {\scriptsize\textbf{BFMT}}\par
        \IfFileExists{figs/DS.png}{\includegraphics[width=\linewidth]{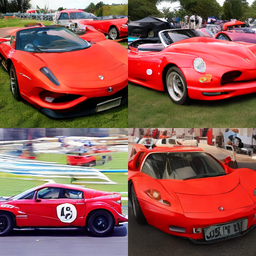}}{\fbox{\begin{minipage}[c][28mm][c]{\linewidth}\centering\scriptsize Add\\\detokenize{figs/DS.png}\end{minipage}}}%
    \end{minipage}\hfill
    \begin{minipage}[t]{0.32\linewidth}
        \centering
        {\scriptsize\textbf{Reward}}\par
        \IfFileExists{figs/DS_plot.png}{\includegraphics[width=\linewidth]{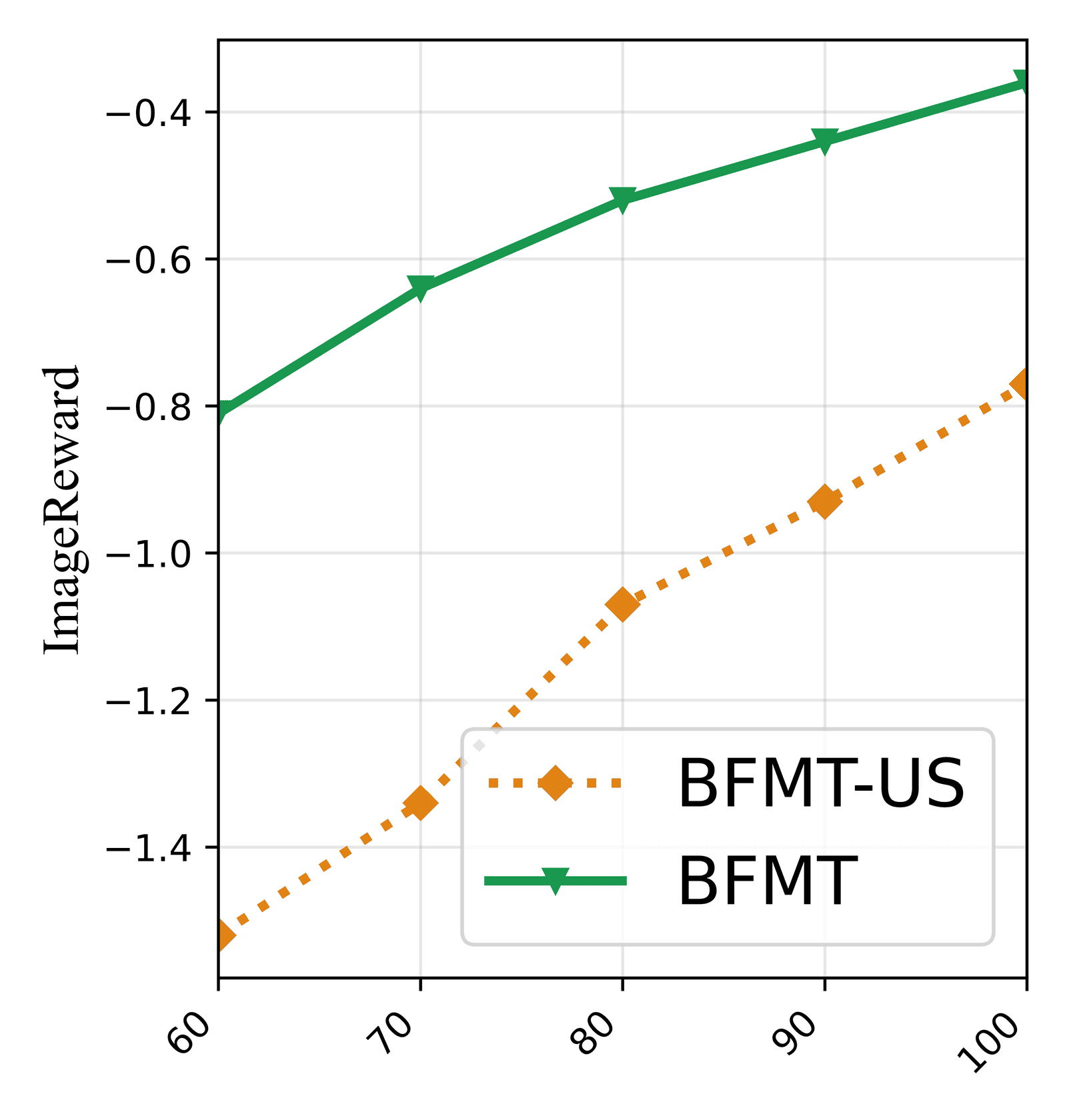}}{\fbox{\begin{minipage}[c][28mm][c]{\linewidth}\centering\scriptsize Add\\\detokenize{figs/DS_plot.png}\end{minipage}}}\par
        {\scriptsize\textbf{Budget}}
    \end{minipage}
    \captionof{figure}{\small{Impact of Dynamic Transition step.}}
    \label{fig:dynamic_transition}
    \vspace{-4pt}
\end{minipage}
\end{figure*}

\vspace{-5pt}
\paragraph{Task Overview and Baselines} We assess BFMT in two settings: \emph{search} and \emph{alignment}. In the \emph{search} setting, an unconditional prior model must generate samples that satisfy fine-grained, class-specific objectives while operating under strict sampling and computational budgets. In the \emph{alignment} setting, the goal is to guide a conditional prior model toward generating samples that satisfy complex, fine-grained constraints involving compositional and quantitative reasoning within fixed computational and sampling limits.
We compare BFMT with a diverse collection of competitive, state-of-the-art methods, including the tree-based Diffusion Tree Search (DTS)~\citep{jain2025diffusion}, Feynman--Kac Steering (FKS)~\citep{singhal2025general}, the SMC-based DAS method~\citep{kim2025test}, and the recently introduced Meta-Flow-Map (MFM) approach~\citep{potaptchik2026meta}. To simulate feedback, we use two widely adopted reward models: the VQA score obtained with CLIP-FlanT5~\citep{lin2024evaluating} and the ImageReward score~\citep{xu2023imagereward}.
\vspace{-8pt}
\paragraph{Search Experimental Setting:} To evaluate the search capabilities of the BFMT, we utilize ImageNet-1K. We employ a distilled flow map model optimized jointly using the $\mathcal{L}^{\text{distill}}_{\text{inst}}$ and $\mathcal{L}^{\text{distill}}_{\text{const}}$ loss objectives as the prior. We evaluate BFMT and all competing methods on 200 randomly selected target classes, each paired with its corresponding fine-grained specifications. Examples of the target classes and prompts are provided in the Appendix~\ref{sec:prompt-a}. We report results under two complementary metrics: mean reward and max reward. Mean reward captures the overall alignment efficacy of the sampler, reflecting how well it performs across all generated samples. Max reward, on the other hand, measures optimal alignment performance — a critical metric in downstream tasks where producing a single high-quality, target-aligned sample is more valuable than broad generation quality. For a fair evaluation, we include MFM (BoN) as a baseline only in the max reward comparison, and exclude it from the mean reward comparison. This choice reflects its design: MFM (BoN) optimizes for a single, highly aligned target sample rather than broad, population-wide alignment efficiency.
\vspace{-4pt}
\begin{figure*}[!h]
\centering

\newcommand{\legenditem}[4]{%
  \begin{tikzpicture}[baseline=-0.6ex]
    \draw[#1, line width=1.8pt, #2] (0,0) -- (0.9,0);
    \node[text=#1, fill=white, inner sep=0.6pt] at (0.45,0) {\scriptsize $#3$};
  \end{tikzpicture}\,#4%
}

\newcommand{\methodlegend}{%
  {\footnotesize
  \setlength{\tabcolsep}{5pt}%
  \begin{tabular}{@{}ccccc@{}}
    \legenditem{cyan!70!black}{solid}{\circ}{DAS} &
    \legenditem{red!75!black}{dashed}{\square}{FKS} &
    \legenditem{orange!90!black}{dash dot}{\triangle}{MFM (BoN)} &
    \legenditem{purple!80!black}{dotted}{\diamond}{DTS} &
    \legenditem{green!60!black}{solid}{\blacktriangledown}{\textbf{BFMT (Our)}}
  \end{tabular}%
  }%
}

\newcommand{\barpanelL}[2]{%
  \begin{minipage}[t]{0.235\linewidth}
    \centering
    \includegraphics[width=\linewidth,height=0.58\textheight,keepaspectratio]{#1}\par
    \vspace{-0.3ex}
    {\scriptsize #2}
  \end{minipage}%
}

\newcommand{\barpanelR}[2]{%
  \begin{minipage}[t]{0.485\linewidth}
    \centering
    \IfFileExists{#1}{\includegraphics[width=\linewidth,height=0.40\textheight,keepaspectratio]{#1}}{\fbox{\begin{minipage}[c][0.40\textheight][c]{\linewidth}\centering\scriptsize Add\\\detokenize{#1}\end{minipage}}}\par
    \vspace{-0.3ex}
    {\scriptsize #2}
  \end{minipage}%
}

\methodlegend
\vspace{0.3ex}

\begin{minipage}[t]{0.655\textwidth}
  \centering
  {\small\textbf{Mean Reward}\par}
  \vspace{-0.5ex}
  \noindent\rule{\linewidth}{0.5pt}
  \vspace{-1.4ex}

  \barpanelL{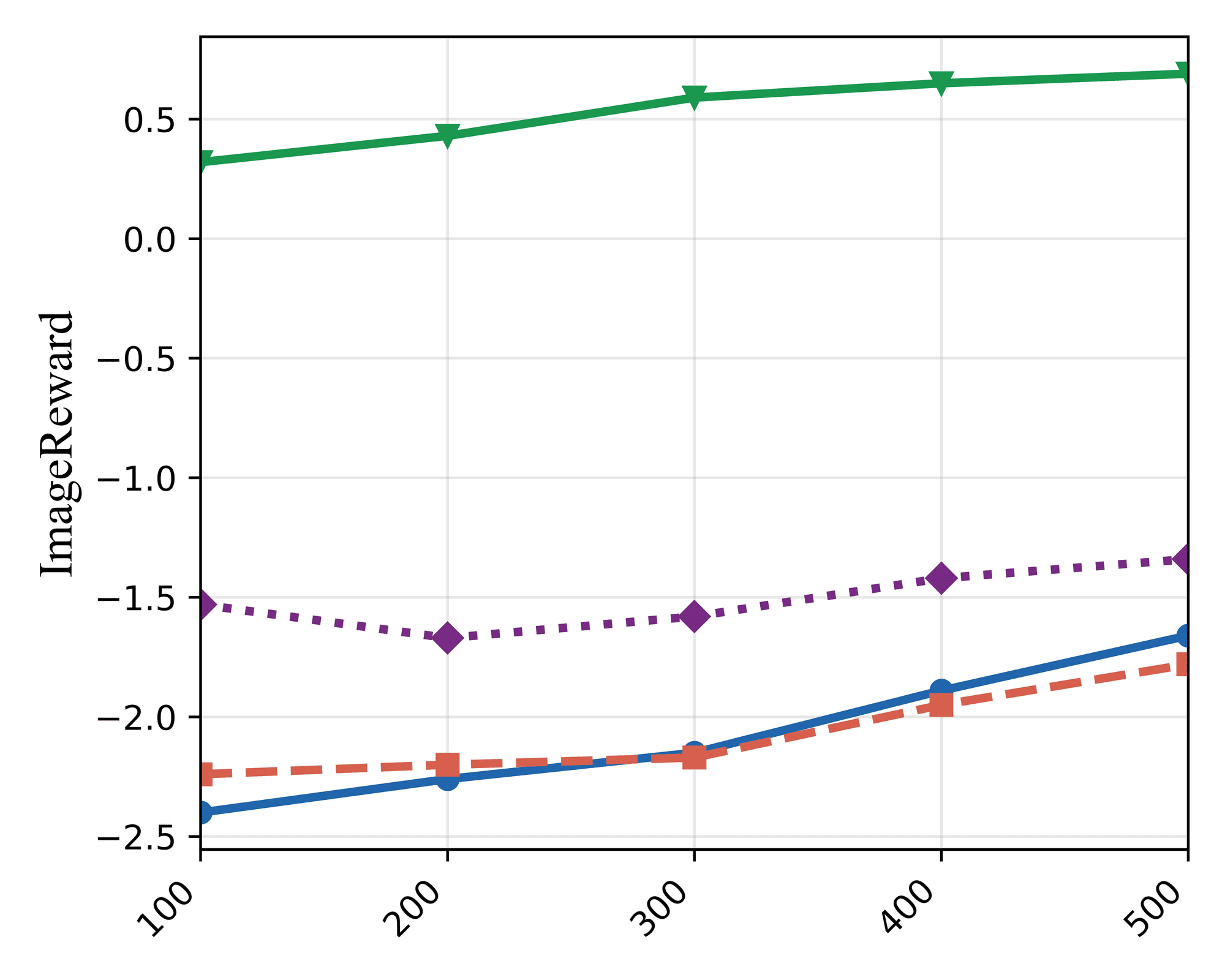}{NFE}\hspace{0.012\linewidth}%
  \barpanelL{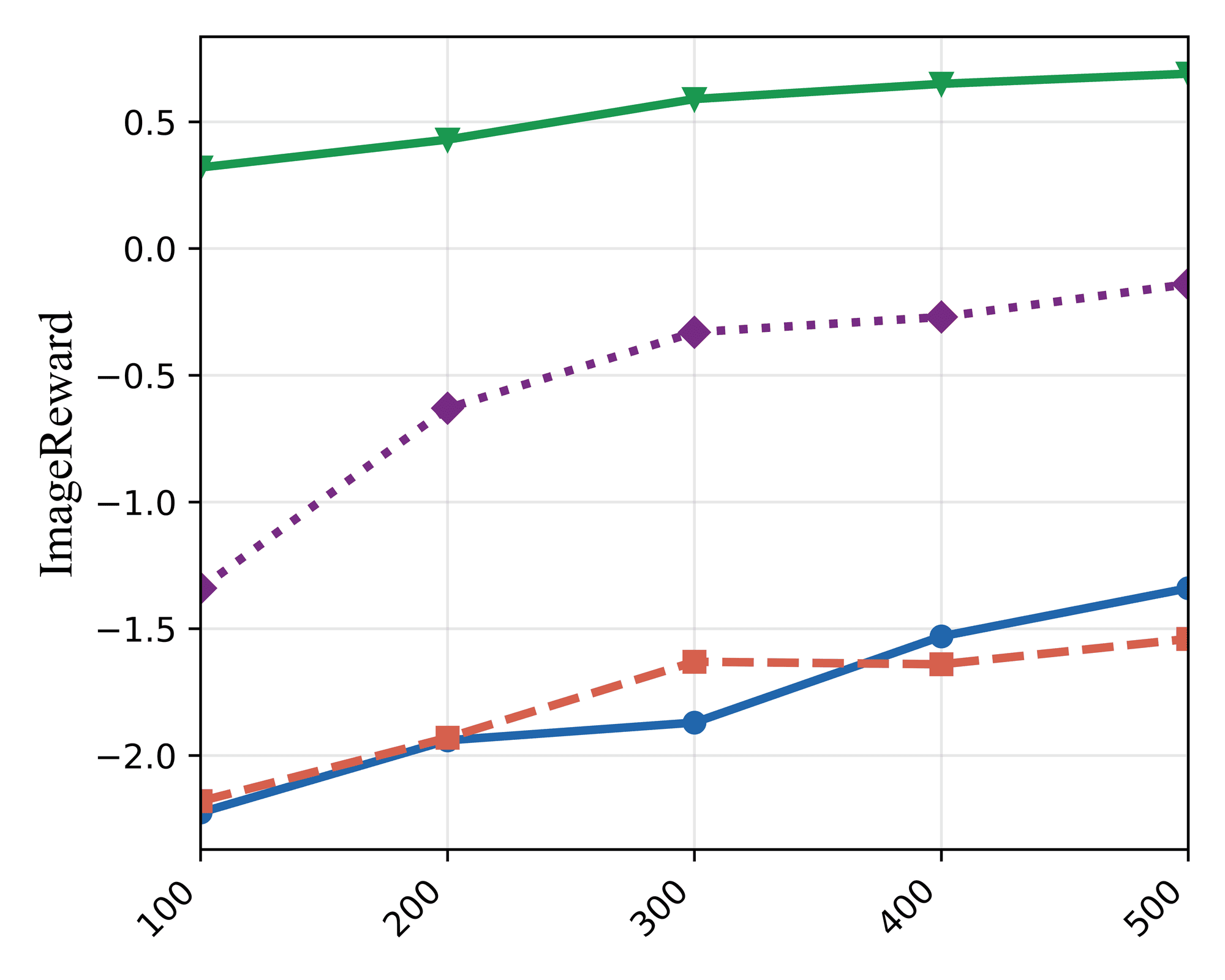}{Feedback Budgets}\hspace{0.012\linewidth}%
  \barpanelL{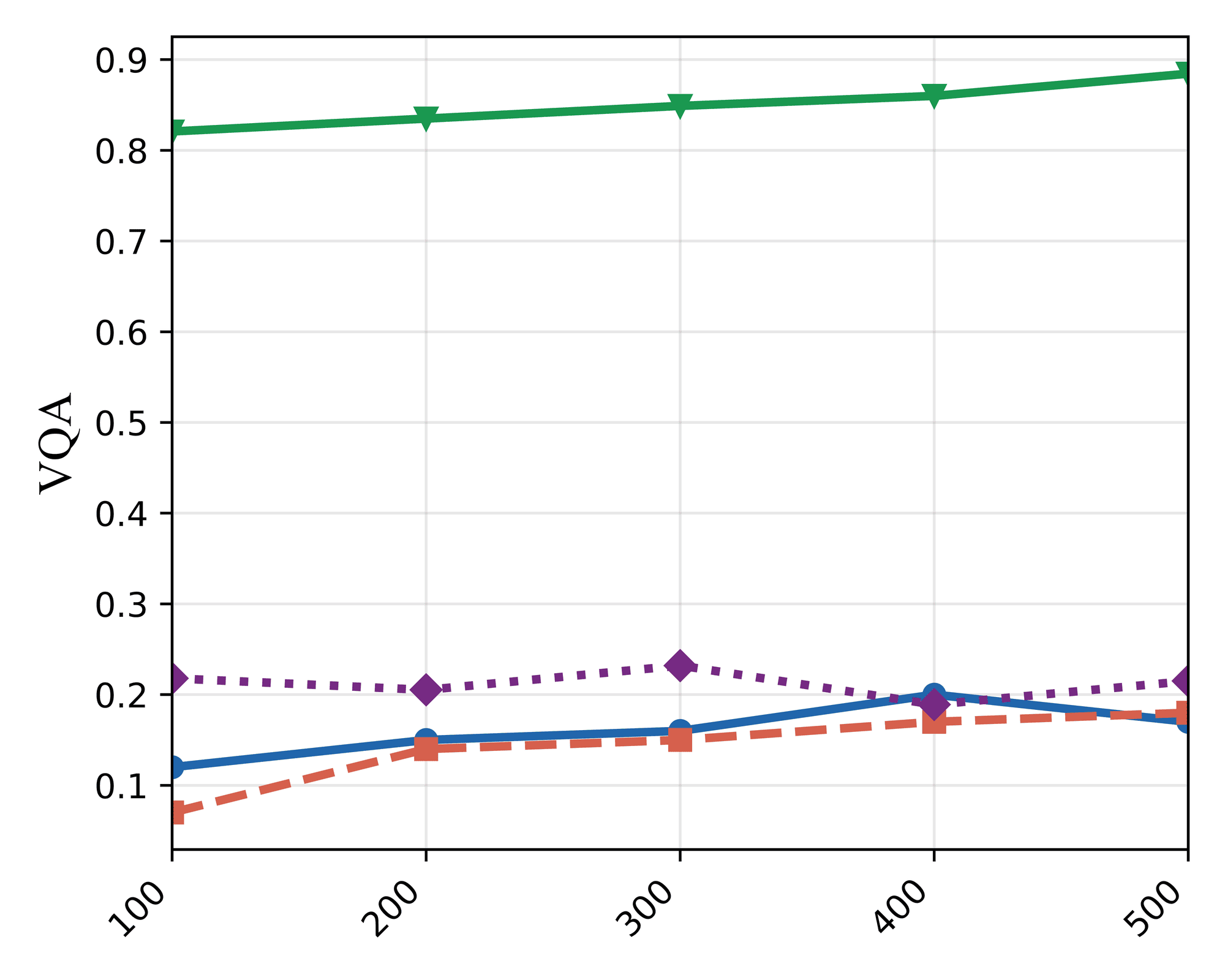}{NFE}\hspace{0.012\linewidth}%
  \barpanelL{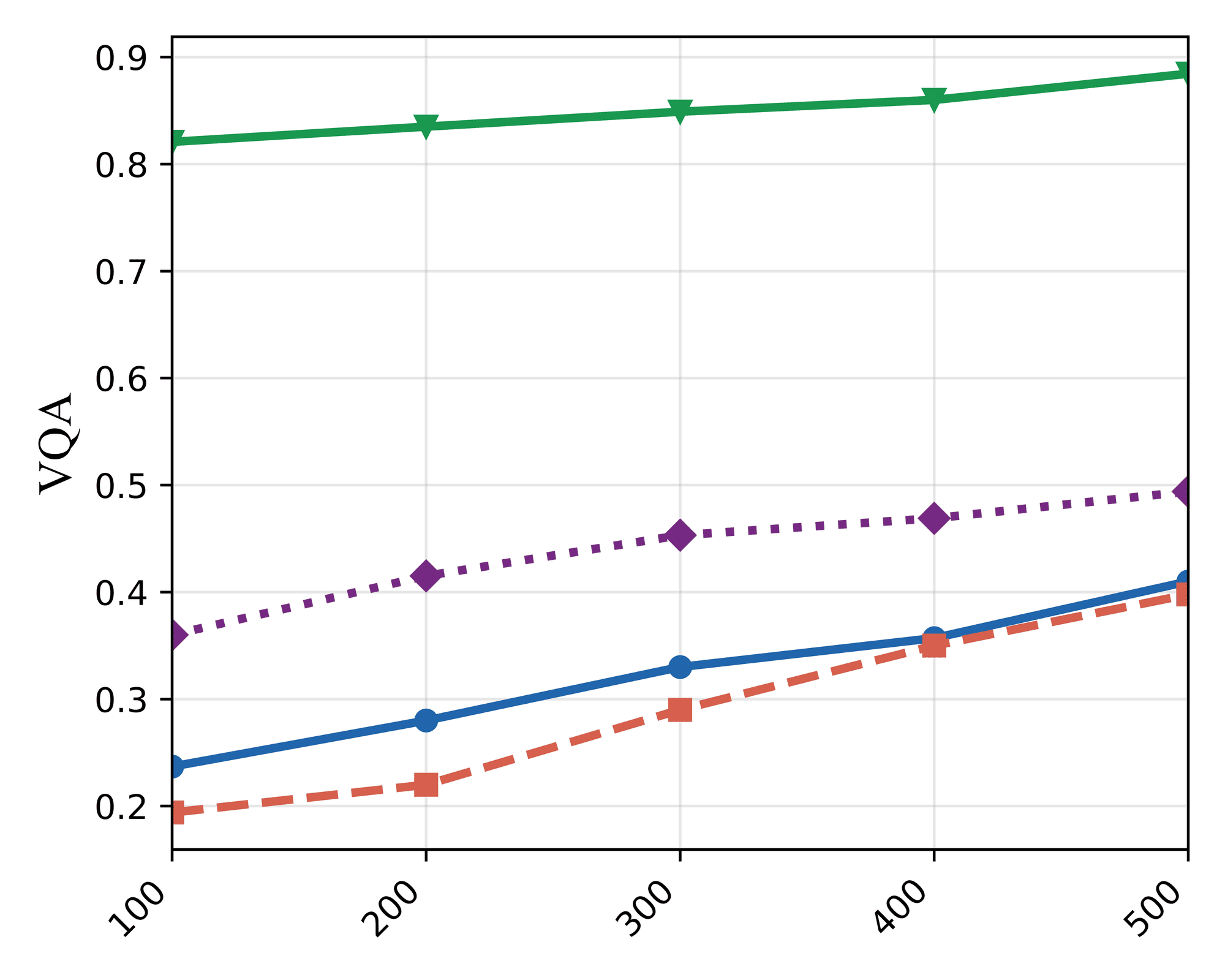}{Feedback Budgets}
\end{minipage}%
\hspace{0.006\textwidth}%
\begin{minipage}[t]{0.335\textwidth}
  \centering
  {\small\textbf{Max Reward}\par}
  \vspace{-0.5ex}
  \noindent\rule{\linewidth}{0.5pt}
  \vspace{-1.4ex}

  \barpanelR{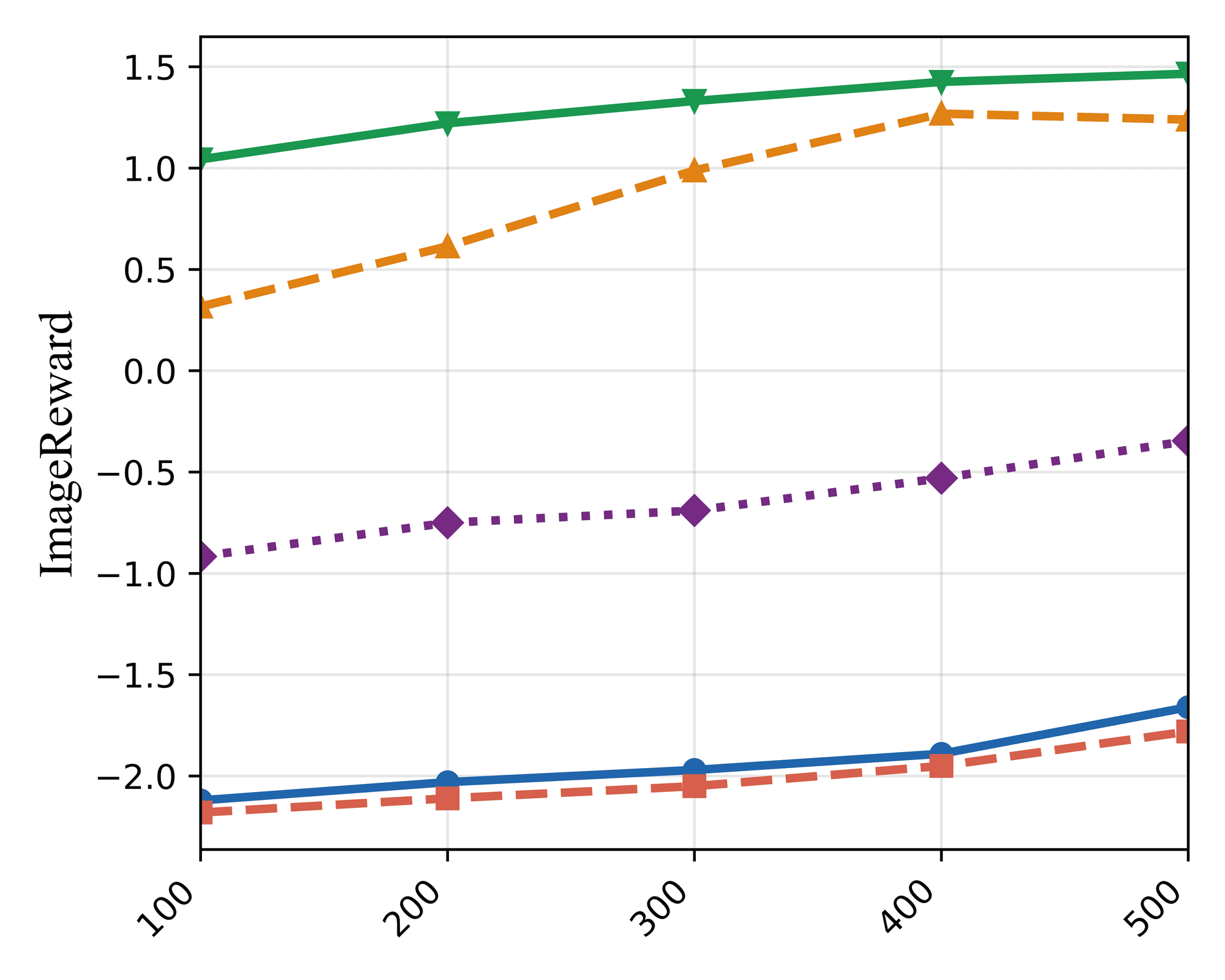}{NFE}\hspace{0.012\linewidth}%
  \barpanelR{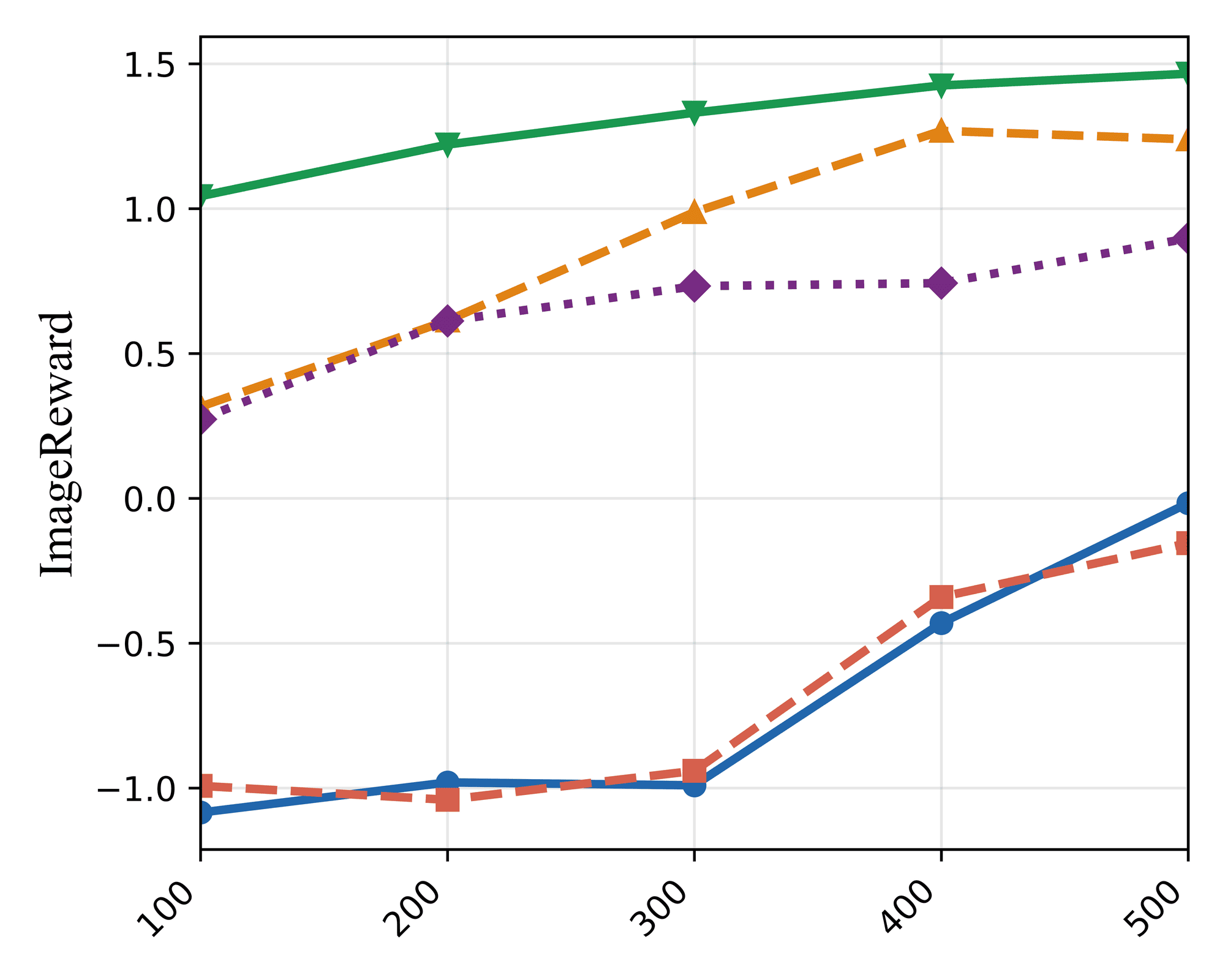}{Feedback Budgets}
\end{minipage}

\vspace{-3pt}
\caption{\small{Search Performance Analysis on ImageNet Target Classes. (left) Mean reward; (right) Max reward.}}
\label{fig:search-result}
\label{fig:search-result-max}
\end{figure*}
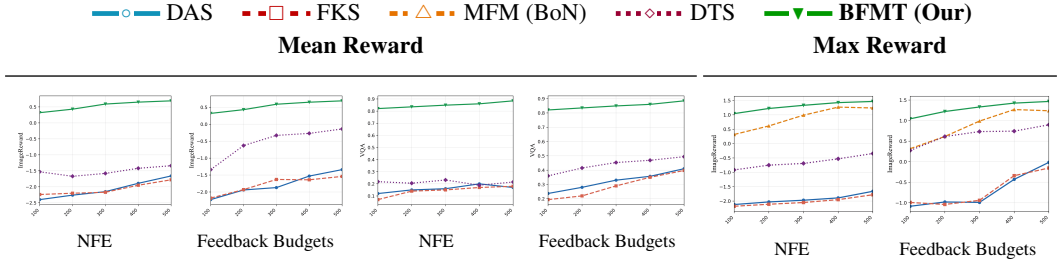
\vspace{-16pt}
\paragraph{Search Result:} 
Fig.~\ref{fig:search-result} demonstrates that BFMT consistently achieves superior reward scores using substantially fewer NFEs and feedback queries than all baselines. This extreme computational efficiency is driven by the bootstrap sufficient statistic mechanism, which synthesizes complete trajectories in a single NFE. Grounded in Proposition 4.3, this allows BFMT to cheaply simulate extensive diffusion horizons, directly exploiting quadratic convergence scaling to accelerate target alignment. Furthermore, BFMT maximizes feedback efficiency through the synergy of arbitrary-duration Flow-Map transitions and budget-aware node selection. This unlocks dynamic, hierarchical search: early stages use small transitions and exploration-biased selection to broadly discover global macro-structures, while deeper stages leverage larger transitions and exploitation-focused selection to fine-tune granular details (Several detailed visualizations illustrating BFMT's hierarchical search strategies are presented in the Appendix~\ref{sec:vis-hi},~\ref{sec:exp-a},~\ref{sec:hie-vis-b},~\ref{sec:hie-vis-a}). Qualitative comparisons (Fig.~\ref{fig:search-visu} and additional visualizations are in the Appendix~\ref{sec:search-vis-a}) further confirm this precise target alignment.
\begin{figure*}[h]
\centering
\setlength{\tabcolsep}{2pt}
\resizebox{\textwidth}{!}{%
{\fontsize{9.5}{10.5}\selectfont
\begin{tabular}{@{}ccc@{\hspace{16pt}}ccc@{}}
\multicolumn{3}{c}{\itshape ``a red sports car.''} & \multicolumn{3}{c}{\itshape ``A dog in the snow.''} \\[4pt]
\textbf{DTS} & \textbf{MFM} & \textbf{BFMT} & \textbf{DTS} & \textbf{MFM} & \textbf{BFMT} \\[4pt]
\hline
\includegraphics[width=3.6cm]{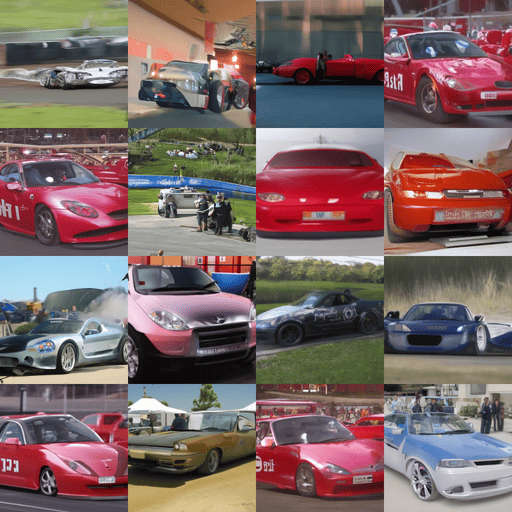} &
\includegraphics[width=3.6cm]{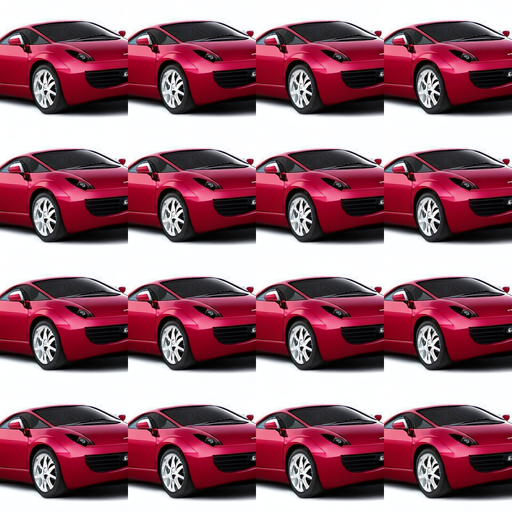} &
\includegraphics[width=3.6cm]{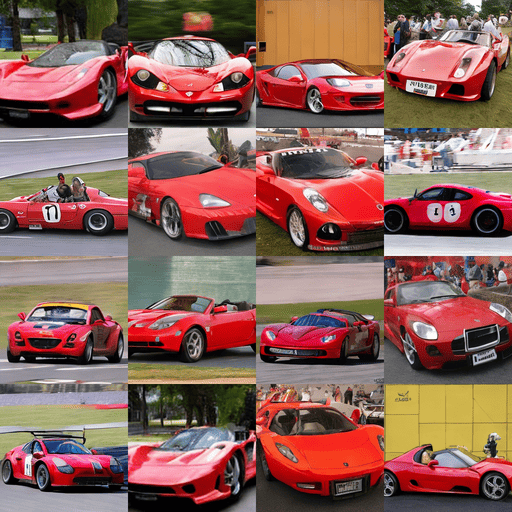} &
\includegraphics[width=3.6cm]{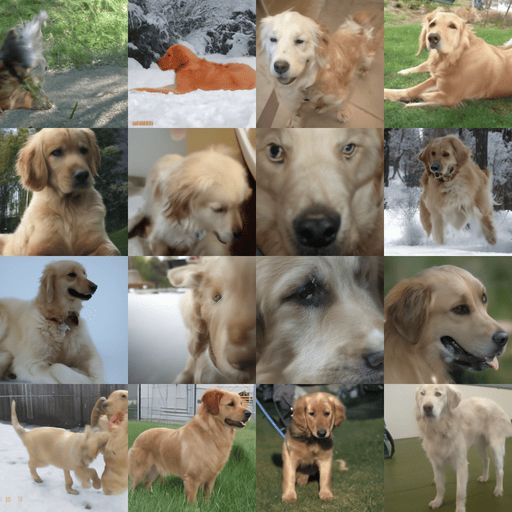} &
\includegraphics[width=3.6cm]{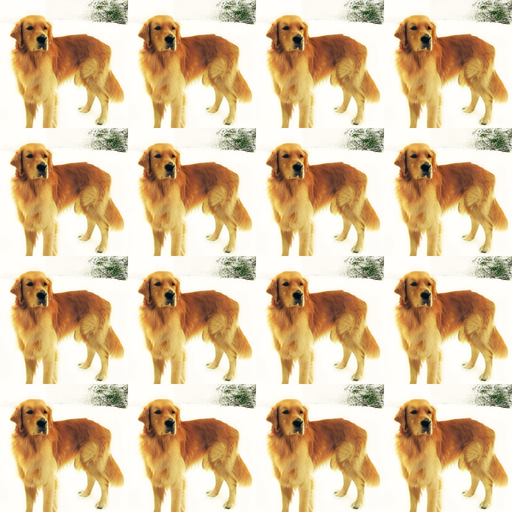} &
\includegraphics[width=3.6cm]{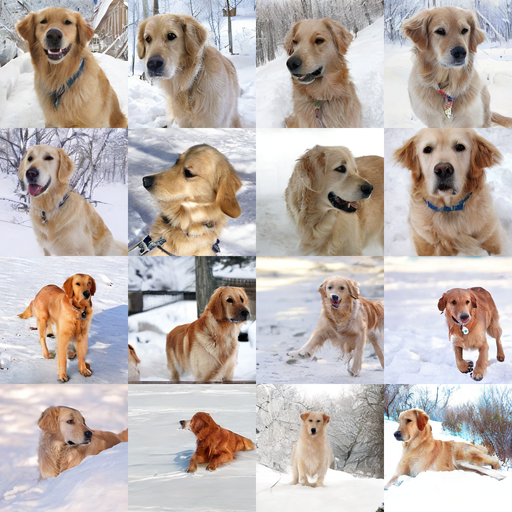} \\
\end{tabular}%
}%
}
\caption{\small{Search Visualizations.}}
\label{fig:search-visu}
\vspace{-8pt}
\end{figure*}
\vspace{-10pt}
\paragraph{Search Performance Evaluation with Best Reward}
  In this section, we evaluate the search performance of BFMT against several baseline methods, including MFM Best-Of-N (BoN), which is specifically optimized for best-reward evaluation.  Detailed comparative results are presented in the figure ~\ref{fig:search-result-max}. Consistent with our earlier analysis, we use ImageReward for evaluation. Our experiments demonstrate that BFMT outperforms all baselines, including BoN. These results underscore a key strength of BFMT: beyond producing better target-aligned samples on average — as reflected in higher mean reward — it also excels at generating the single best target-aligned sample. This capability is particularly valuable in applications where producing an optimal output matters more than overall alignment capability of the sampler.
\paragraph{Impact of BASE}
\vspace{-9pt}
\begin{wrapfigure}{r}{0.34\textwidth}
\vspace{-8pt}
\centering
\caption{\small{BASE vs. UCT.}}
\vspace{-3pt}
\includegraphics[width=\linewidth]{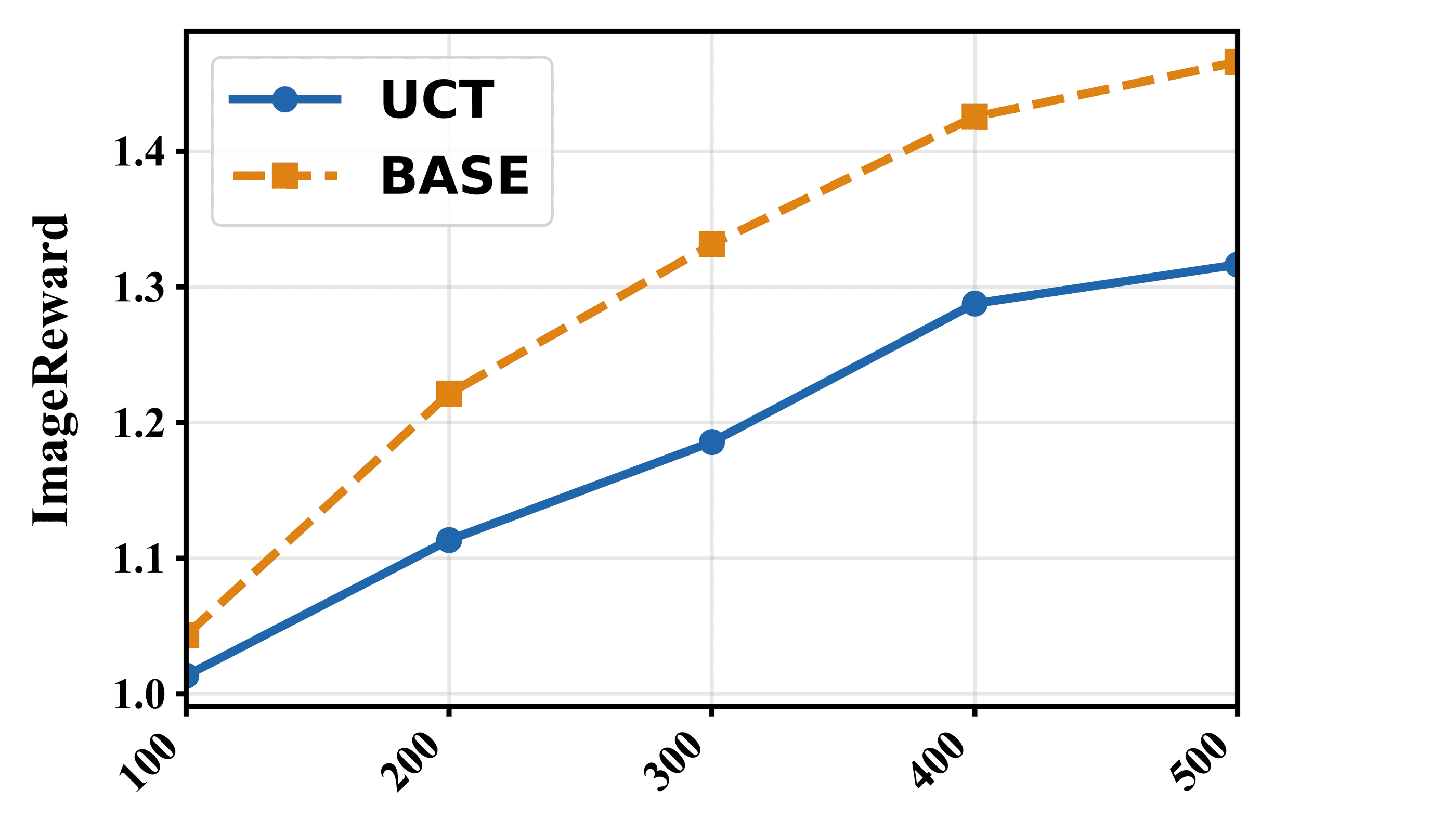}
\par\vspace{-0.5ex}
{\small Feedback Budgets}
\vspace{-0.6ex}
\label{fig:base-uct}
\vspace{-2.0ex}
\end{wrapfigure}
To evaluate the effectiveness of the BASE, we compare its performance against a standard UCT-based mechanism across varying feedback budgets. As shown in Fig.~\ref{fig:base-uct}, BASE achieves significantly higher reward scores under any given fixed budget. This demonstrates that incorporating budget awareness into the node selection process provides far more efficient control over the exploration-exploitation tradeoff than standard UCT. These results underscore the critical role of the BASE node selection strategy within the BFMT framework in enabling feedback-efficient search.
\vspace{-9pt}
\paragraph{The Impact of Bootstrap Sufficient Statistic (BSS):}\label{pg:bss}
To evaluate the impact of the BSS mechanism within BFMT, we compare our approach against BFMT-ISI, a variant that replaces BSS with an alternative ODE-based, single-NFE trajectory generation scheme: mapping $x_t$ to $x_0$ via a pretrained flow-map, followed by an Iterative Stochastic Interpolant (ISI) between $x_0$ and $\epsilon \sim \mathcal{N}(0, I)$ to construct intermediate nodes. As shown in Fig.~\ref{fig:BSS}, this alternative degrades sample quality. Furthermore, because it lacks the DDPM-like transitions guaranteed by BSS, it fails to construct valid trajectories, which hinders convergence to the target distribution and leads to lower reward scores (Fig.~\ref{fig:BSS}). We visualize the impact of BSS in Appendix~\ref{sec:BSS-a}.
\vspace{-4pt}

\begin{figure*}[!h]
\centering
\newlength{\molh}\setlength{\molh}{2.0cm}
\newlength{\rowgap}\setlength{\rowgap}{4pt}

\begin{minipage}[c]{0.40\linewidth}
\centering
\begin{minipage}[t]{0.48\linewidth}
\centering
\begin{overpic}[width=0.86\linewidth, height=\molh]{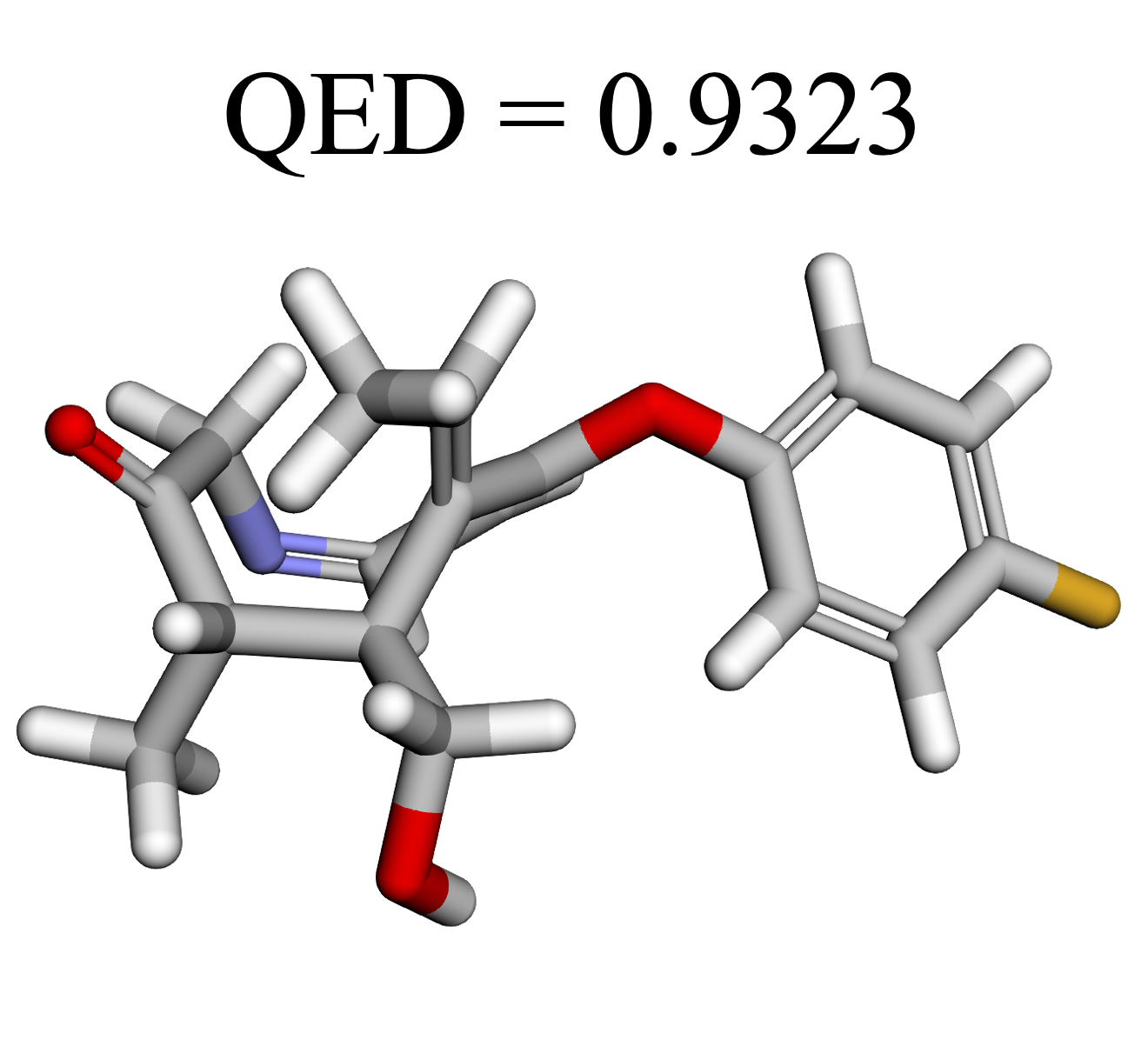}
\end{overpic}
\end{minipage}\hfill
\begin{minipage}[t]{0.48\linewidth}
\centering
\begin{overpic}[width=0.86\linewidth, height=\molh]{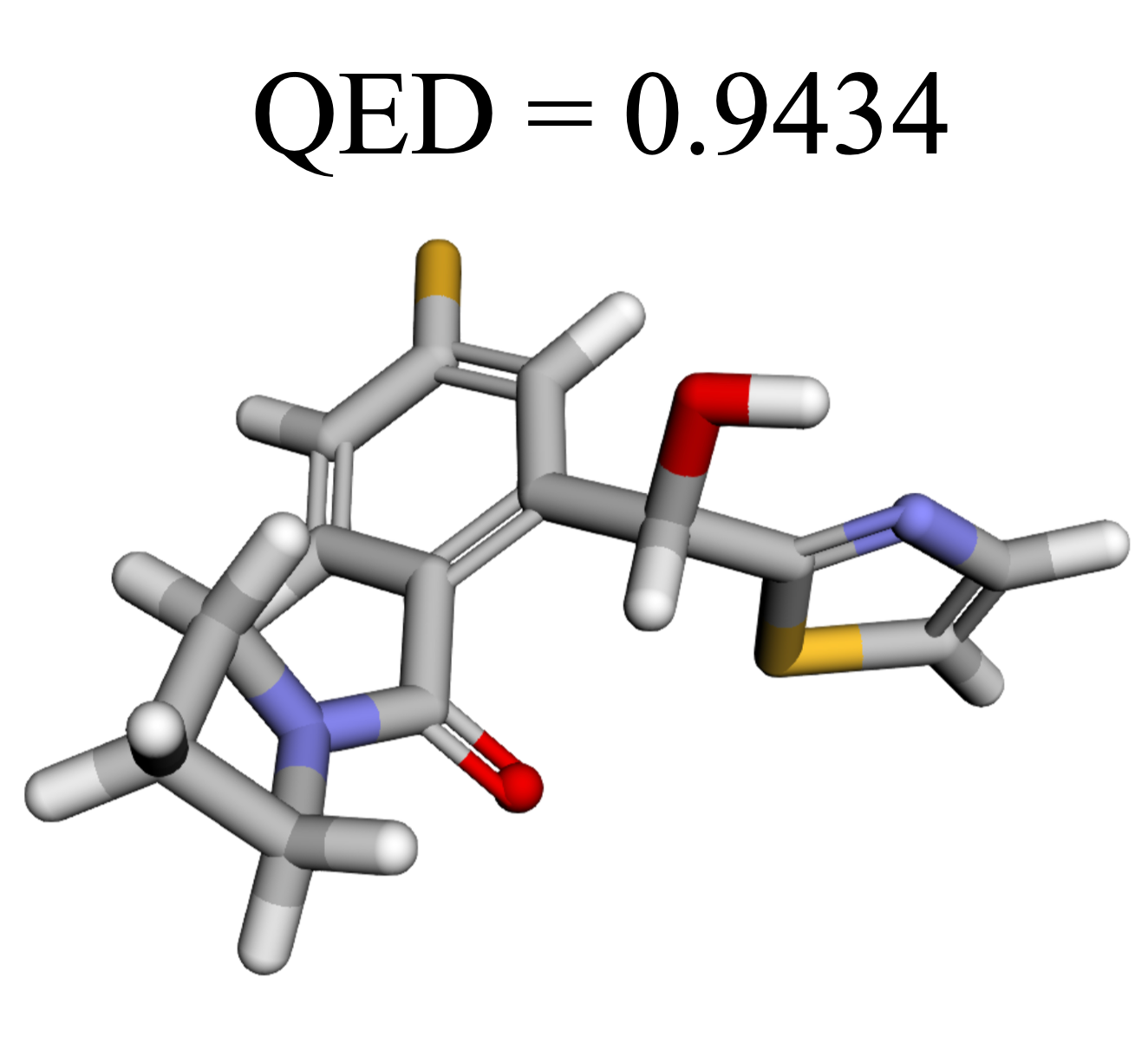}
\end{overpic}
\end{minipage}

\vspace{\rowgap}

\begin{minipage}[t]{0.48\linewidth}
\centering
\begin{overpic}[width=0.86\linewidth, height=\molh]{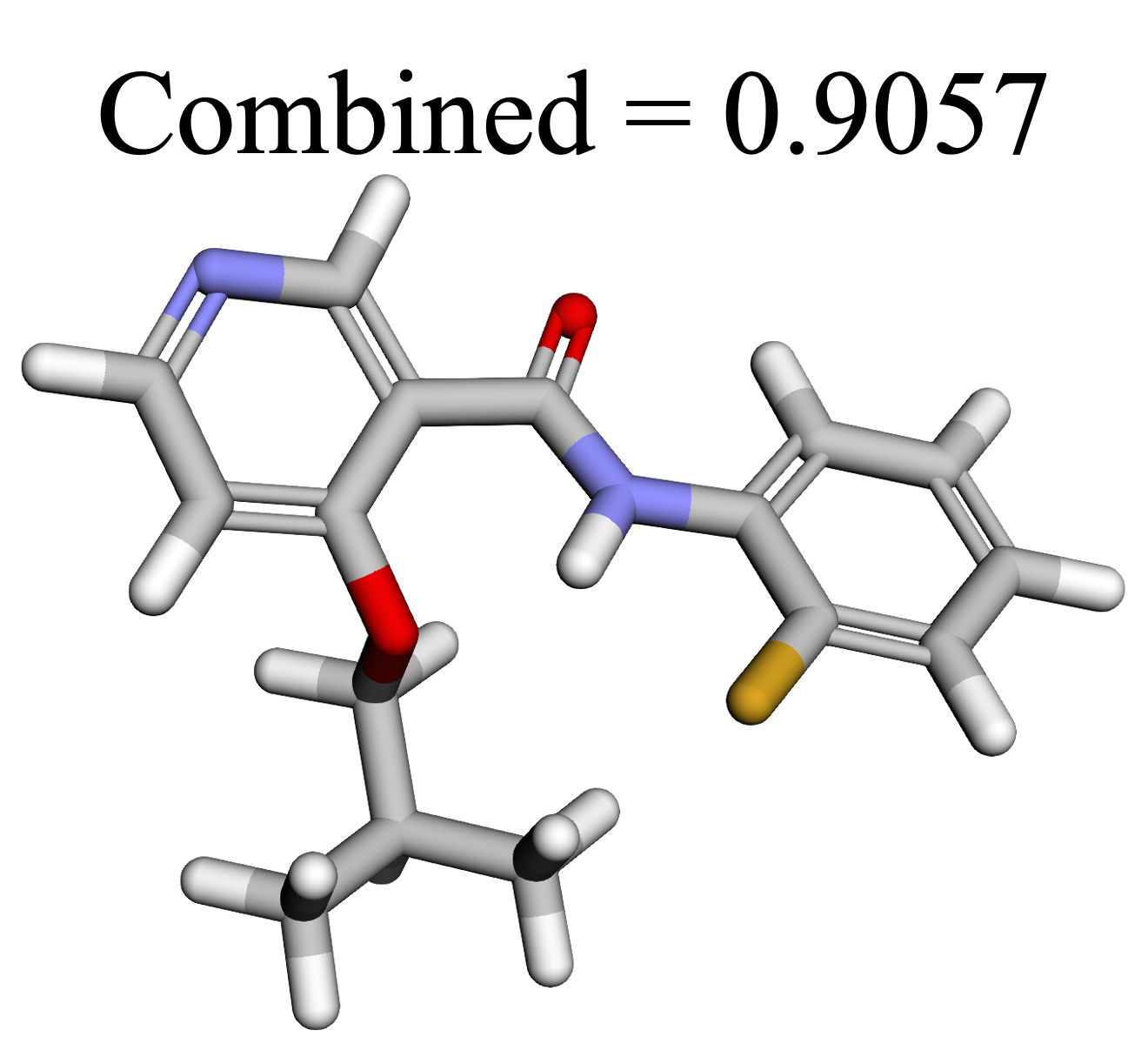}
\end{overpic}
\end{minipage}\hfill
\begin{minipage}[t]{0.48\linewidth}
\centering
\begin{overpic}[width=0.8\linewidth, height=\molh]{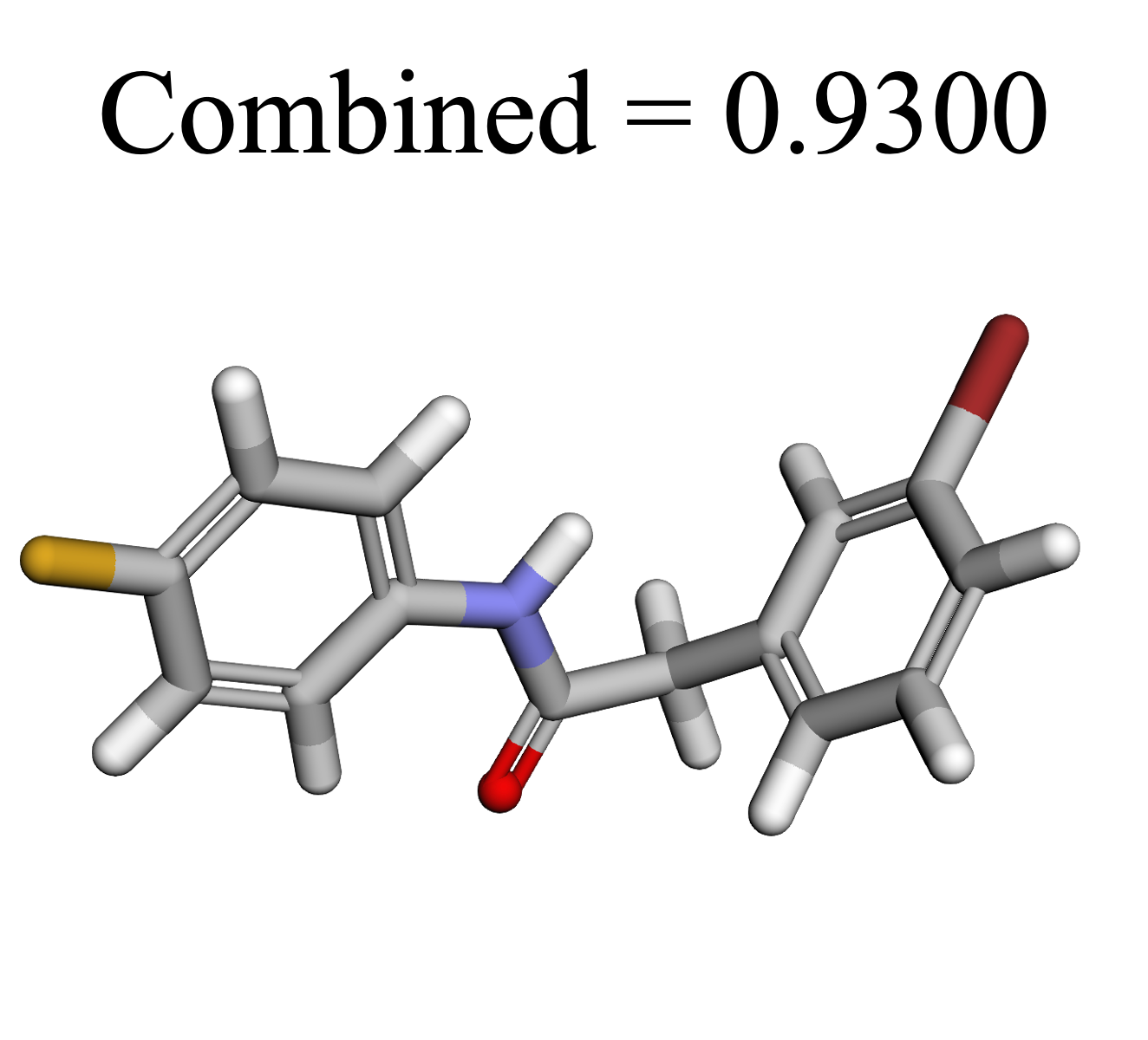}
\end{overpic}
\end{minipage}
\end{minipage}%
\hfill
\begin{minipage}[c]{0.30\linewidth}
\centering
\begin{overpic}[width=\linewidth, height=\dimexpr 2\molh+\rowgap\relax]{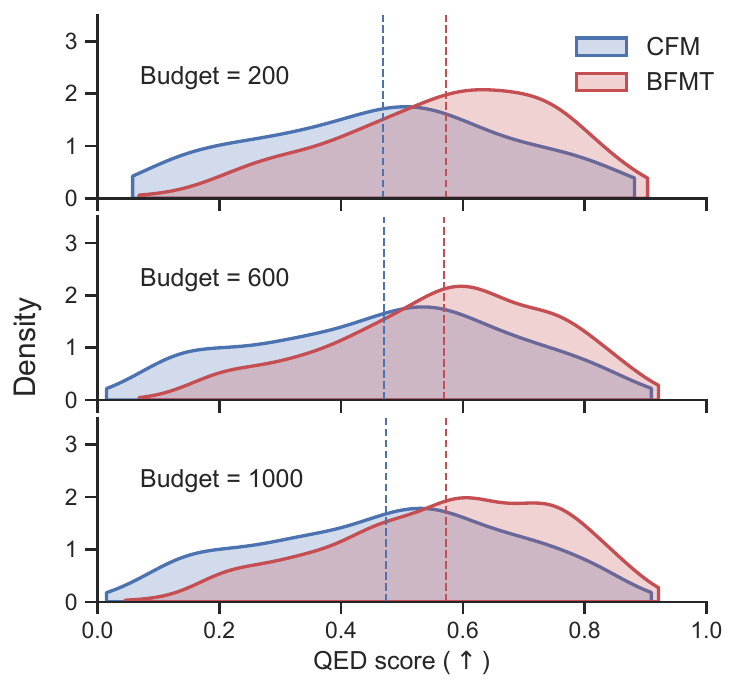}
\end{overpic}
\end{minipage}%
\hfill
\begin{minipage}[c]{0.30\linewidth}
\centering
\begin{overpic}[width=\linewidth, height=\dimexpr 2\molh+\rowgap\relax]{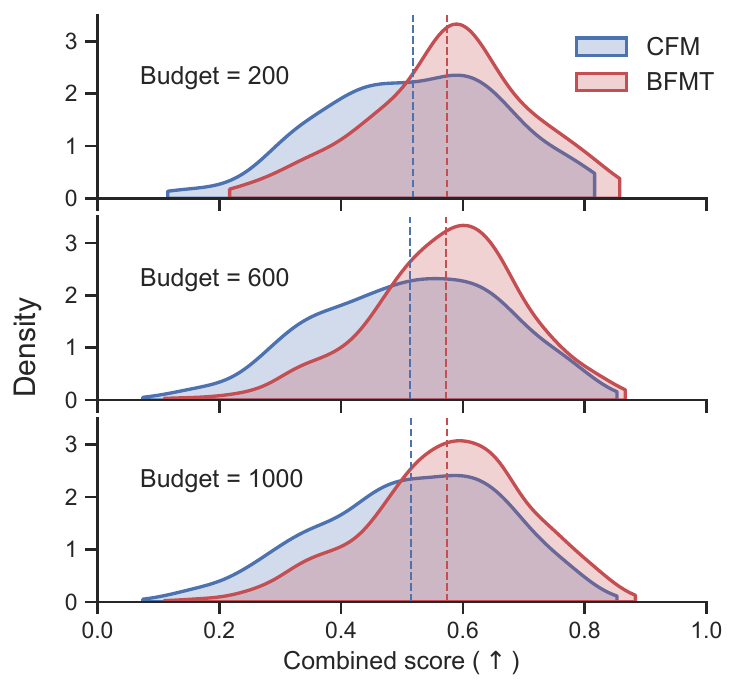}
\end{overpic}
\end{minipage}
\caption{Effectiveness of BFMT on target-aligned feedback-constrained Molecule Generation Task.}
\label{fig:mol-res}
\end{figure*}

\vspace{-5pt}
\vspace{-8pt}
\paragraph{Effect of Dynamic Transition Schedule on Search}\label{pg:dynamic}
BFMT leverages Flow-Maps to enable dynamic transitions during trajectory rollout, providing precise control over the exploration-exploitation tradeoff, key to enable efficient search. By employing smaller transition steps during the early denoising phase, BFMT achieves broader exploration. Conversely, larger steps in the final phases promote localized exploitation. Given a fixed NFE budget, this non-uniform transition scheduling enables BFMT to conduct a more efficient search compared to its uniform-step counterpart, BFMT-US. This enhanced exploratory capacity directly yields a more diverse set of generated samples aligning with the target prompt, while resulting in higher reward across varying NFE (see Fig.~\ref{fig:dynamic_transition}). 
\vspace{-10pt}
\paragraph{Results with Large-Scale Molecular Structure Generation Task}
We consider $10$M molecules extracted from PubChem~\citep{kim2023pubchem}. The molecules are represented as SELFIES~\citep{krenn2020self}, a robust string representation for molecular structures. These molecules are used to train Categorical Flow Maps (CFM), a generative flow-matching model developed by~\citep{roos2026categorical}. We trained the CFM model for $16$ epochs using the default hyperparameters and then used the trained model for inference-time search. We subsequently used BFMT to steer generation toward regions of chemical space with high (i) Quantitative Estimate of Drug-likeness (QED) and (ii) a  combination of QED and synthetic accessibility (SA) scores. QED measures a molecule's drug-likeness, while SA estimates how easily it can be synthesized. Our goal is to generate molecules that satisfy these objectives and assess whether BFMT produces a higher proportion of desirable molecules than vanilla CFM sampling, while maintaining comparable validity and uniqueness.
\begin{wraptable}{r}{0.72\textwidth}
\centering
\vspace{-10pt}
\caption{Validity Check for the Generated Molecules.}
\vspace{-4pt}
\label{tab:mole-res}
\footnotesize
\setlength{\tabcolsep}{2.6pt}
\renewcommand{\arraystretch}{0.80}
\begin{tabular}{l ccc|ccc}
\toprule
Method & Validity & Unique. & Mean QED & Validity & Unique. & Mean QED$\&$SA \\
\midrule
CFM & 0.9396 & 0.9968 & 0.4892 & 0.9396 & 0.9968 & 0.5173 \\
\rowcolor{blue!20}BFMT & 0.9558 & 0.9644 & \textbf{0.5610} & 0.9540 & 0.9746 & \textbf{0.5746} \\
\bottomrule
\end{tabular}
\vspace{-8pt}
\end{wraptable}
Fig.~\ref{fig:mol-res} illustrates the target property distributions alongside representative visualizations of the generated molecules. Empirically, BFMT consistently outperforms CFM by generating samples that more effectively satisfy the target criteria, maintaining this superior search performance across all evaluated metrics and feedback budgets. 
Table~\ref{tab:mole-res} reports the mean validity, uniqueness, QED, and the combination of QED and SA scores for samples generated under feedback budget 1000. The results indicate that BFMT effectively optimizes for target properties without compromising chemical validity. Ultimately, these results establish BFMT as a highly feedback-efficient sampler capable of navigating complex discrete spaces such as molecular design. Additional results across varying feedback budgets are presented in the Appendix~\ref{sec:mole-add}.
\vspace{-22pt}
\paragraph{Alignment Experimental Setting:}
We assess BFMT's alignment performance across two challenging tasks. First, for the compositional alignment task, we utilize 100 randomly selected prompts from GenAI-Bench~\citep{jiang2024genai} that necessitate at least three advanced compositional skills. 
Second, for the quantity-aware alignment task, we evaluate using 100 prompts randomly sampled from the numeracy category of T2I-CompBench++~\citep{huang2025t2i}. Here, the held-out reward is defined as the negative Residual Sum of Squares (RSS) between the target object counts and the detected counts, where detections are obtained using DINO~\citep{liu2024grounding} and SAM~\citep{kirillov2023segment}. 
We employ an off-the-shelf consistency model (SANA)~\citep{chen2025sana} as the prior for these experiments. We provide a detailed derivation of the reparametrization of the noise schedule used in SANA (see App.~\ref{sec:trigflow_adaptation}) to facilitate its integration within the BFMT framework.
\vspace{-8pt}

\begin{figure*}[h]
\centering
\newcommand{\triplepanel}[3]{%
  \setlength{\tabcolsep}{2pt}%
  \begin{tabular}{@{}c@{\hspace{1pt}}c@{\hspace{1pt}}c@{}}
    \includegraphics[width=0.305\linewidth]{#1} &
    \includegraphics[width=0.305\linewidth]{#2} &
    \includegraphics[width=0.305\linewidth]{#3}
  \end{tabular}%
}
\newcommand{\panelblock}[4]{%
  \begin{minipage}[t]{\linewidth}
    \centering
    {\itshape ``#1''\par}
    \vspace{0pt}
    \triplepanel{#2}{#3}{#4}
  \end{minipage}%
}
\newcommand{\topheaders}{%
  {\fontsize{9.5}{10.5}\selectfont
  \begin{tabular}{@{}c@{\hspace{1pt}}c@{\hspace{1pt}}c@{}}
    \makebox[0.305\linewidth][c]{\textbf{DTS}} &
    \makebox[0.305\linewidth][c]{\textbf{FKS}} &
    \makebox[0.305\linewidth][c]{\textbf{BFMT}}
  \end{tabular}%
  }
}
\noindent
\begin{minipage}[t]{0.48\textwidth}
\centering
\topheaders
\end{minipage}\hfill%
\begin{minipage}[t]{0.48\textwidth}
\centering
\topheaders
\end{minipage}
\vspace{-0.8ex}
\noindent\rule{\textwidth}{0.6pt}
\vspace{-1.2ex}
\begin{minipage}[t]{0.48\textwidth}
\centering
\panelblock{Eight Guitars.}{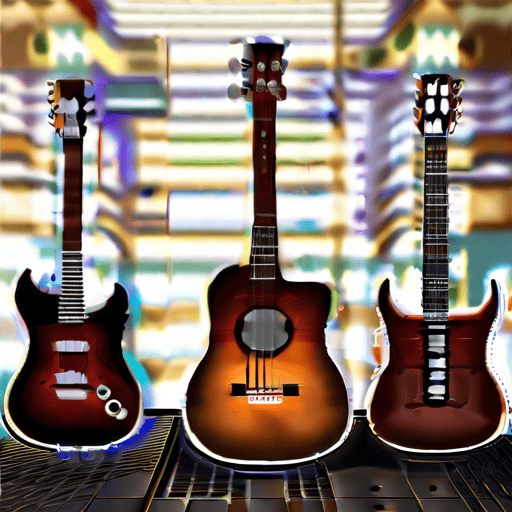}{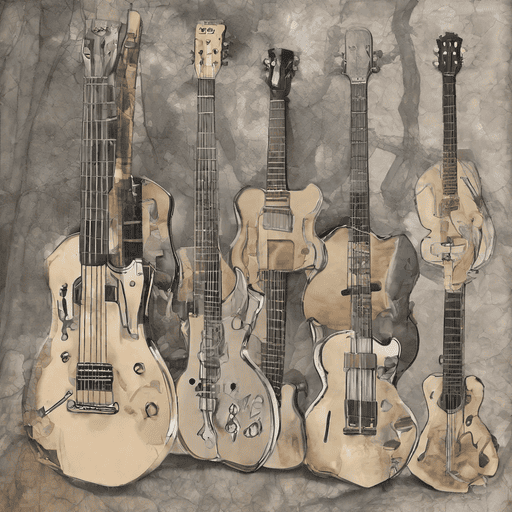}{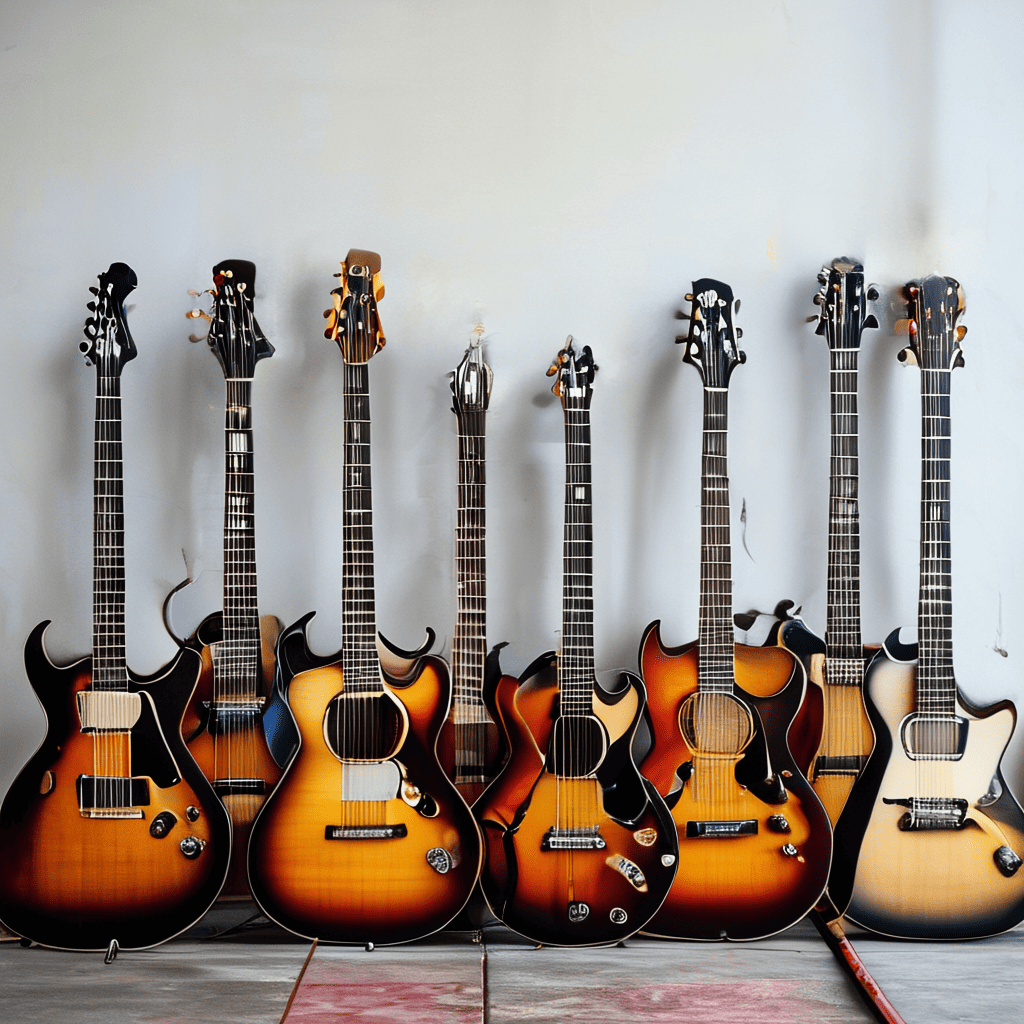}
\end{minipage}\hfill%
\begin{minipage}[t]{0.48\textwidth}
\centering
\panelblock{\small{A crystal tree shimmering under a twilit, starry sky.}}{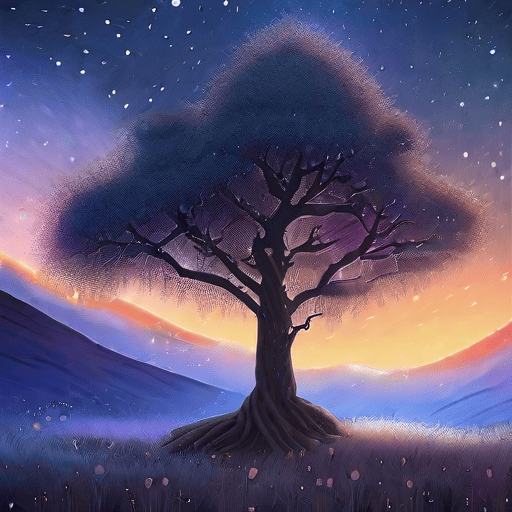}{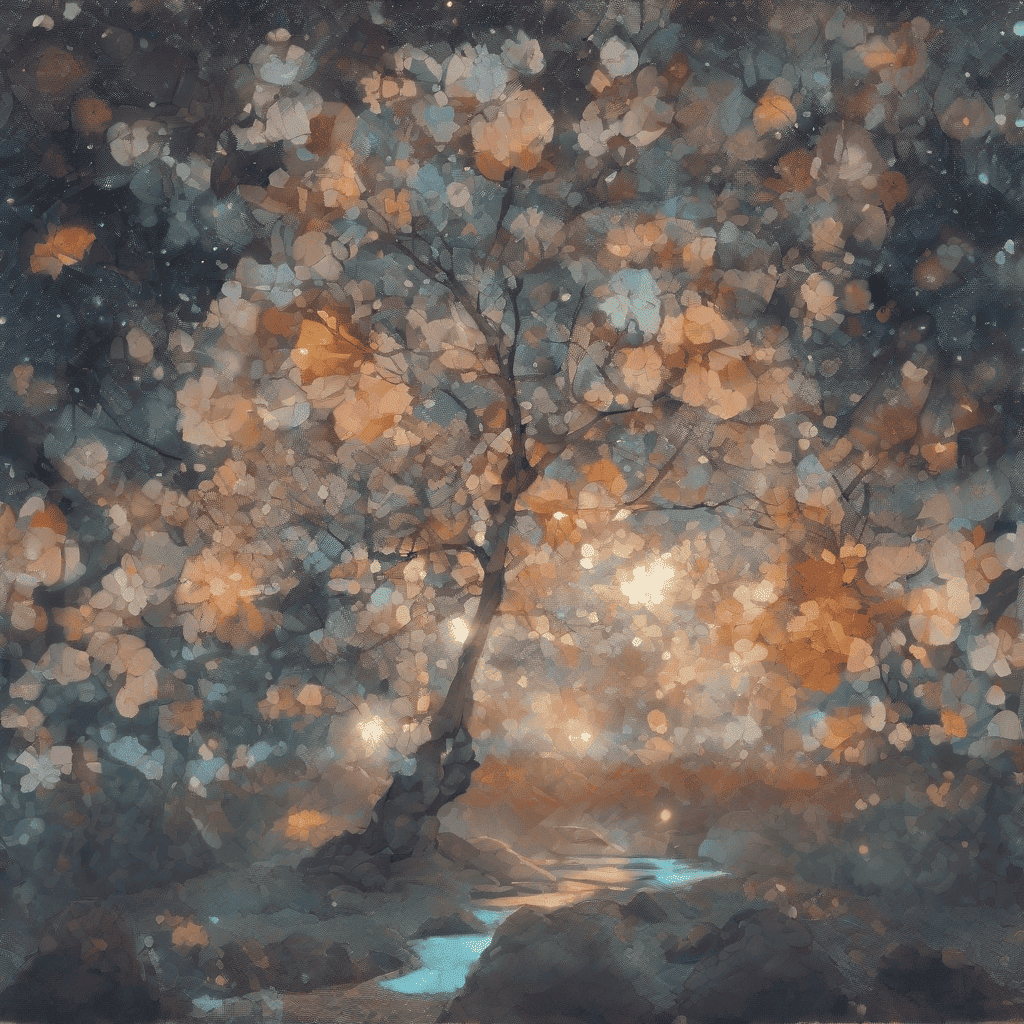}{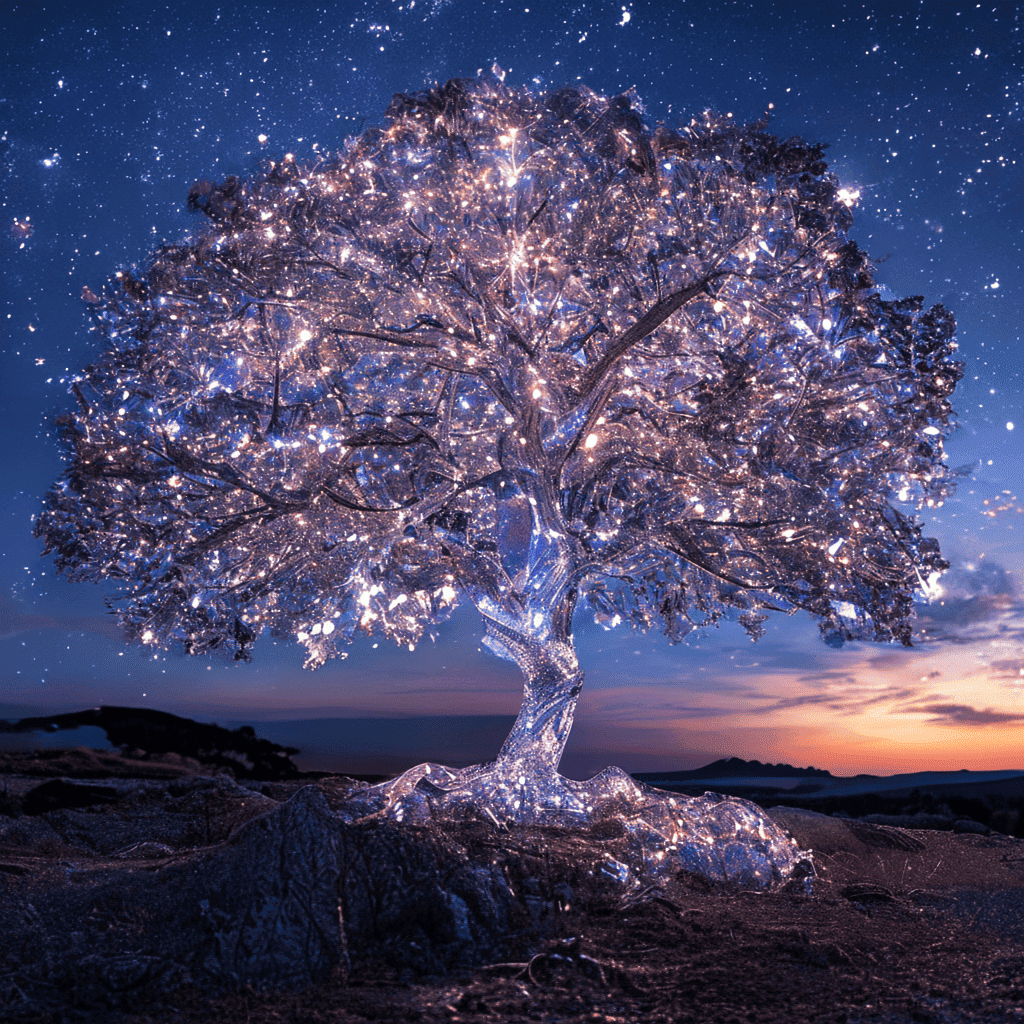}
\end{minipage}
\vspace{-0.45ex}
\noindent\rule{\textwidth}{0.6pt}
\caption{\small{Alignment Visualization. (left) Quantity Prompt. (right) Compositional Prompt. }}
\label{fig:align-visu}
\vspace{-9pt}
\end{figure*}

\begin{figure*}[h]
\centering

\newcommand{\legenditem}[4]{%
  \begin{tikzpicture}[baseline=-0.6ex]
    \draw[#1, line width=1.6pt, #2] (0,0) -- (0.9,0);
    \node[text=#1, fill=white, inner sep=0.6pt] at (0.45,0) {\scriptsize $#3$};
  \end{tikzpicture}\,#4%
}

\newcommand{\methodlegend}{%
  {\footnotesize
  \setlength{\tabcolsep}{4pt}%
  \begin{tabular}{@{}ccccc@{}}
    \legenditem{blue}{solid}{\circ}{DAS} &
    \legenditem{red}{dashed}{\square}{FKS} &
    \legenditem{orange}{dotted}{\triangle}{MFM (BoN)} &
    \legenditem{purple}{dotted}{\diamond}{DTS} &
    \legenditem{green!60!black}{solid}{\triangledown}{BFMT (Our)}
  \end{tabular}%
  }%
}

\newcommand{\barpanel}[2]{%
  \begin{minipage}[t]{0.145\textwidth}
    \centering
    \includegraphics[width=\linewidth]{#1}\par
    \vspace{-0.3ex}
    {\scriptsize #2}
  \end{minipage}%
}

\methodlegend
\vspace{0.2ex}

\noindent
\begin{minipage}[t]{0.64\textwidth}
  \centering
  {\small\textbf{Mean Reward}}
\end{minipage}%
\hfill
\begin{minipage}[t]{0.28\textwidth}
  \centering
  {\small\textbf{Best Reward}}
\end{minipage}

\noindent\rule{\textwidth}{0.5pt}

\barpanel{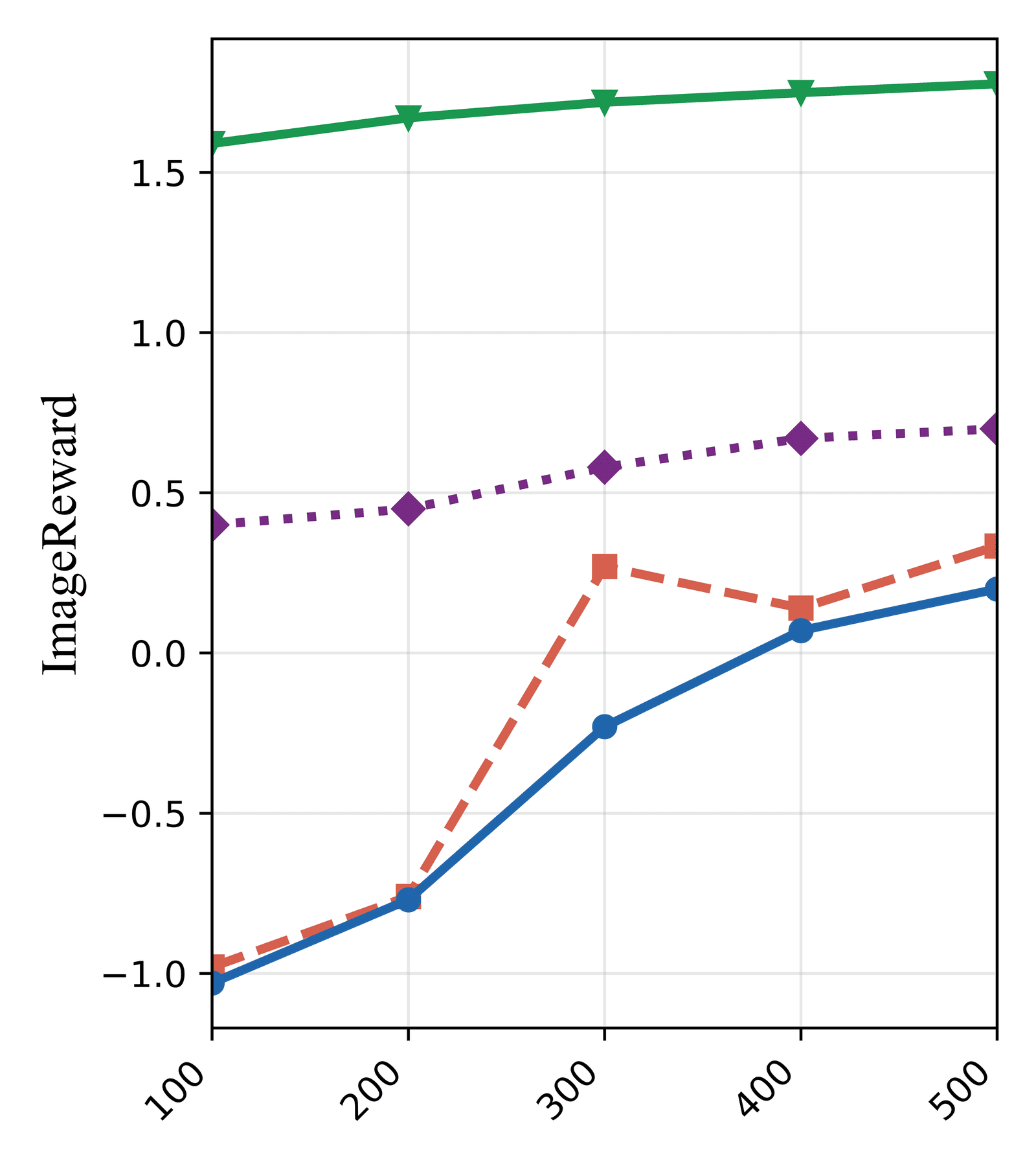}{NFE}\hspace{0.01\textwidth}%
\barpanel{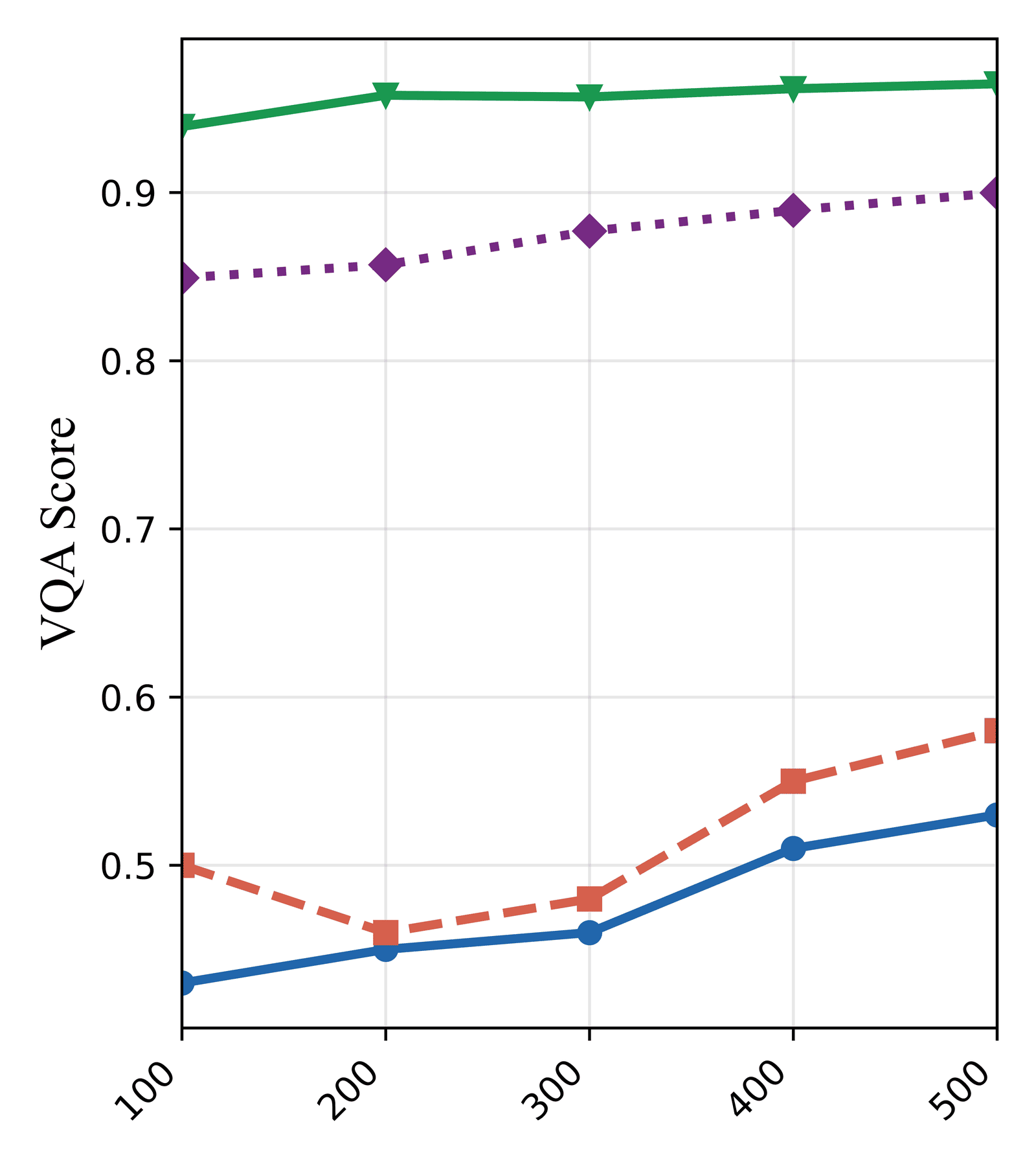}{NFE}\hspace{0.01\textwidth}%
\barpanel{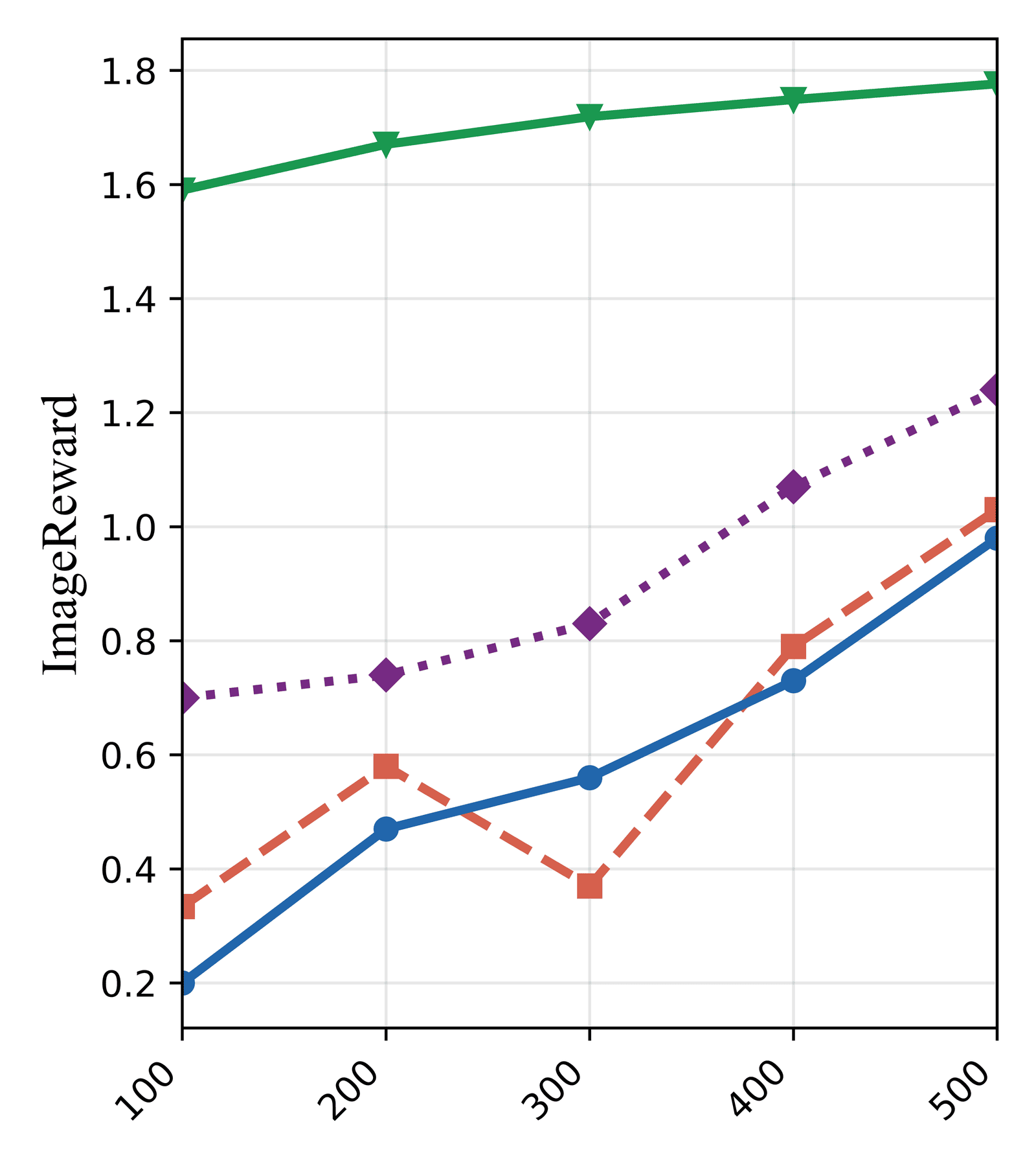}{Feedback Budgets}\hspace{0.01\textwidth}%
\barpanel{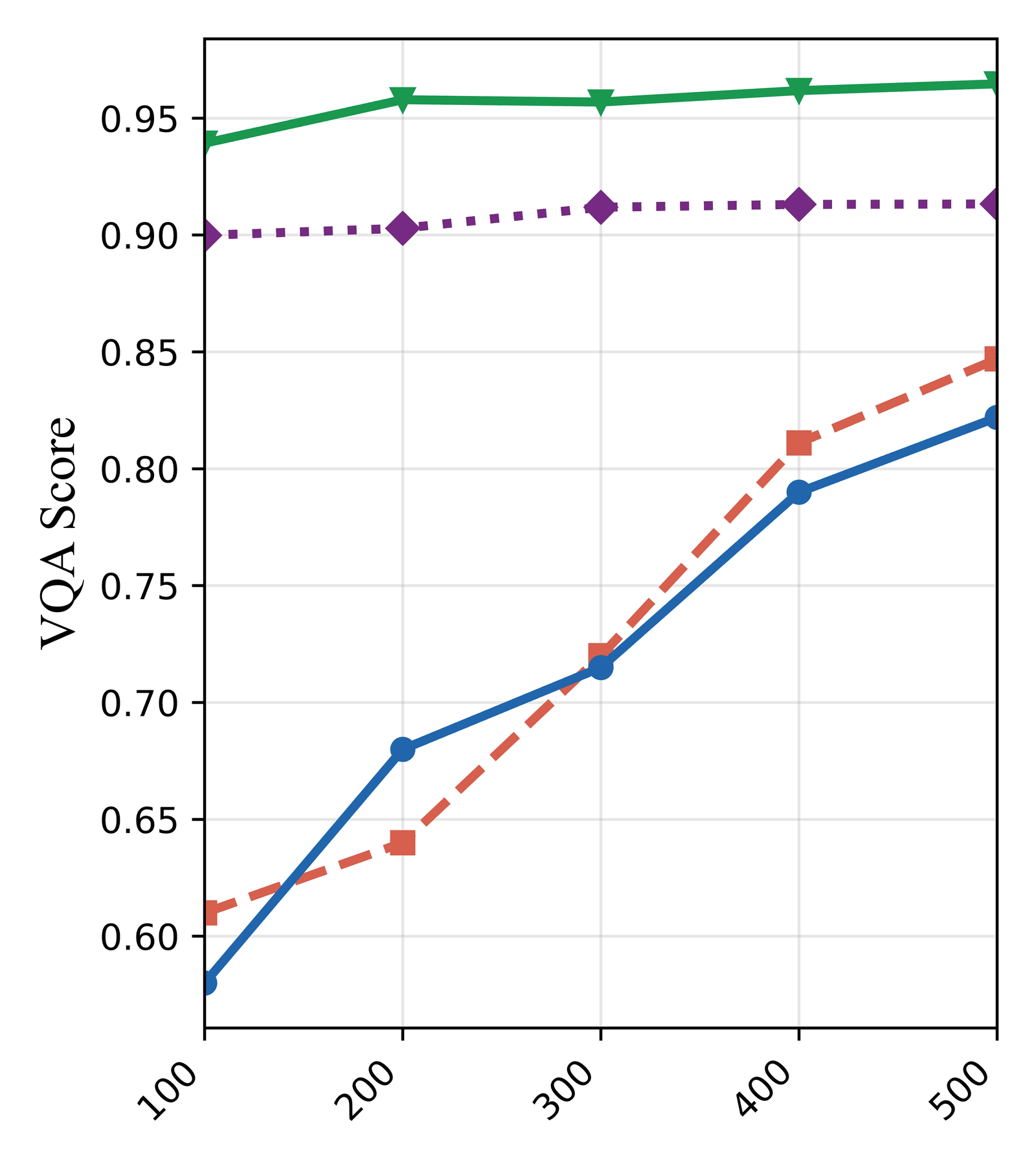}{Feedback Budgets}\hspace{0.012\textwidth}%
{\color{black}\rule{0.3pt}{2.2cm}}\hspace{0.012\textwidth}%
\barpanel{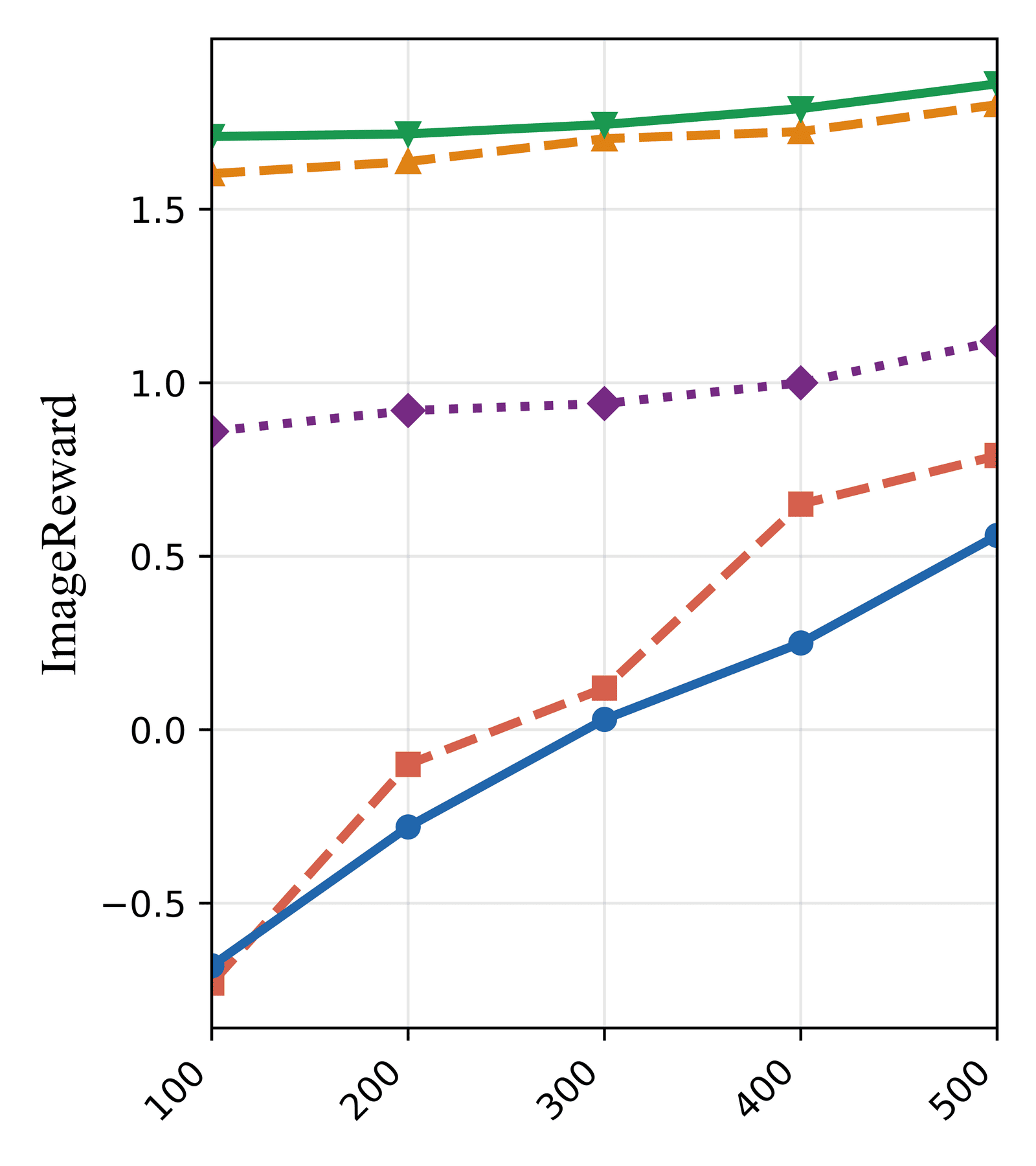}{NFE}\hspace{0.01\textwidth}%
\barpanel{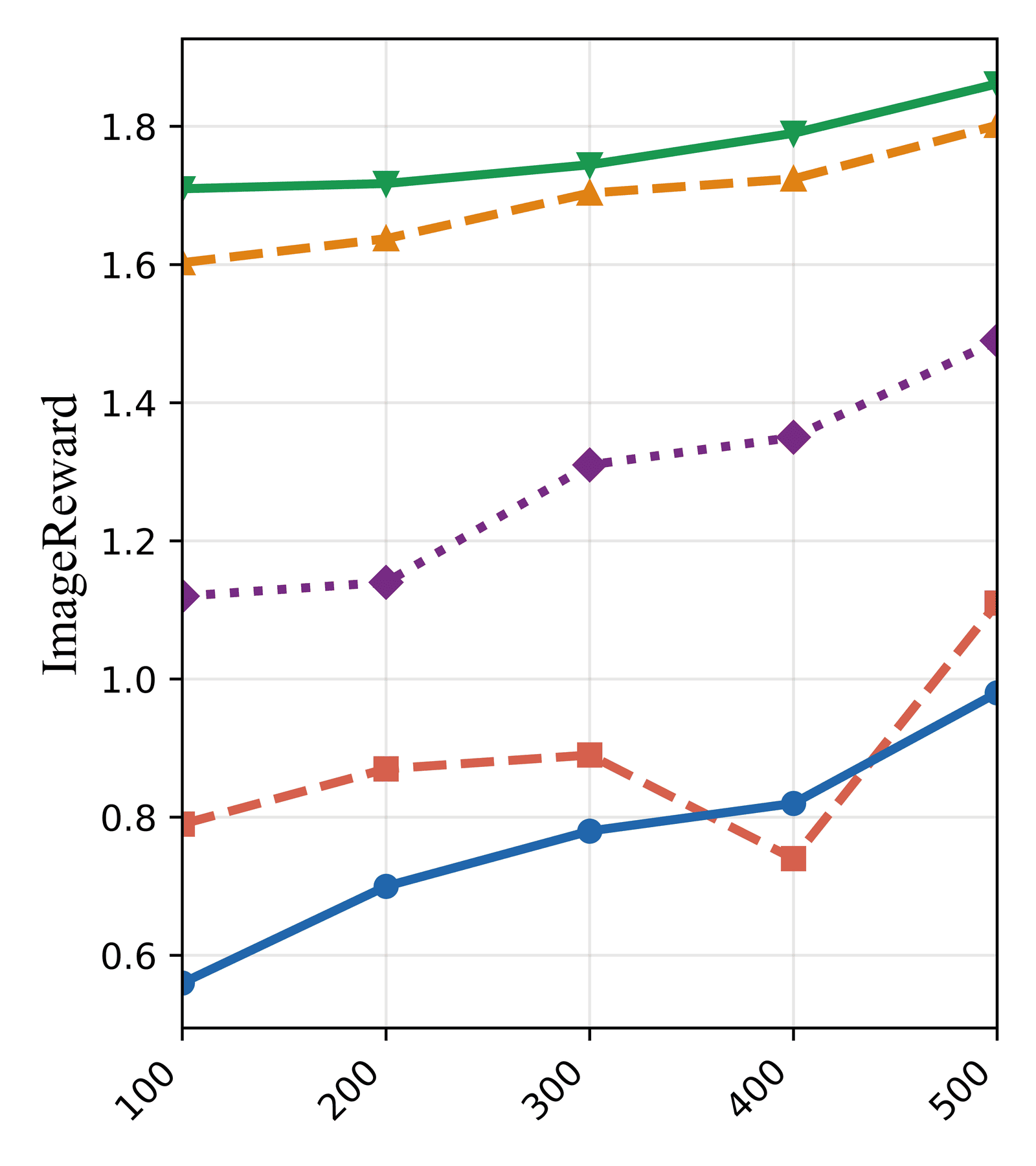}{Feedback Budgets}

\vspace{-3pt}
\caption{\small{Analysis of Competitive Approaches on Compositional Alignment.}}
\label{fig:alignment-result}
\end{figure*}
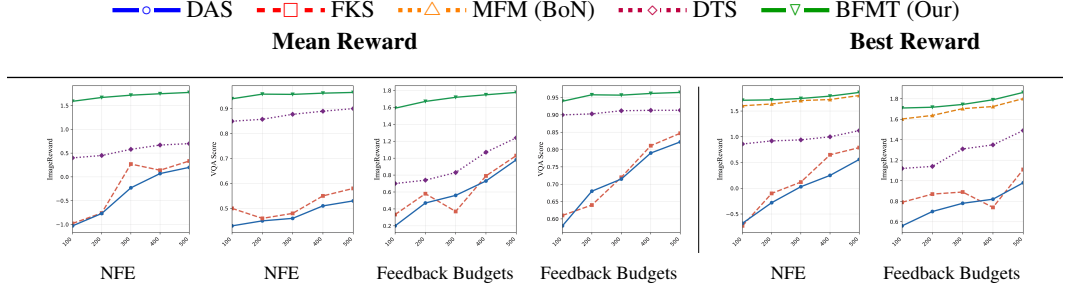

\paragraph{Alignment Result:}
Consistent with the search setting, Fig.~\ref{fig:alignment-result} demonstrates that BFMT achieves superior reward scores compared to competing methods, using significantly fewer NFEs and excelling under a fixed feedback budget.
These findings reinforce the efficacy of BFMT’s highly effective, feedback-driven alignment capability—powered by its dynamic transition schedule and a budget-aware node selection mechanism. Qualitative comparisons are presented in Fig.~\ref{fig:align-visu} (additional viz. in App.~\ref{sec:align-vis-a}). The results on the quantity-aware alignment task are reported in the Appendix~\ref{sec:se-quant}.

\vspace{-7pt}

\vspace{-9pt}
\paragraph{Related Work}\label{sec:rel_work}
\vspace{-8pt}
\smallskip
\paragraph{Tree-Based Sampler For Search}
\vspace{-6pt}
Tree-based samplers have been introduced for effective inference-time search and alignment. DTS~\citep{jain2026diffusion}  casts the denoising process as a finite-horizon tree, where, similar to MCTS, rollouts are used to continuously improve value estimates for intermediate noisy states. DTS thus enables a principled search in the space of denoising trajectories to identify the modes within high-volume regions of the target density. Unlike SMC-style samplers, tree-based frameworks backpropagate only terminal rewards, eliminating the need to evaluate rewards at every intermediate denoising step. While promising, current tree-based samplers~\citep{jain2026diffusion,guo2026training} are severely bottlenecked by the massive number of function evaluations (NFEs) needed for node valuation, rendering them impractical for exploration-heavy tasks. Moreover, their dependence on rigid, uniform transitions restricts adaptive search, and heuristic workarounds like beam search~\citep{li2025dynamic} fail to outperform standard UCT selection~\citep{kocsis2006bandit}. 
\vspace{-6pt}
\paragraph{SMC-based Sampler for Search}
Inference-time scaling has seen rapid adoption, with SMC serving as a primary framework. SMC methods approximate the target distribution by maintaining an evolving population of sample particles. 
SMC~\citep{del2006sequential,kim2025test,singhal2025general} utilize potential functions—which typically approximate the soft value function—to iteratively weight and resample particles at each step. While numerous variants have recently been adapted for diffusion model alignment~\citep{kim2025test,wu2023practical}, classical SMC theoretically guarantees both exact target sampling and optimal value estimation only in the limit of infinite particles.
In practice, repeated resampling inherently diminishes sample diversity due to high weight variance and imprecise value estimation. 
To combat particle weight degeneracy and debias particle-based guidance, several correction techniques~\citep{skreta2025feynman,lee2025debiasing} have recently been proposed. However, these methods falter in high-dimensional domains, such as image synthesis, due to their reliance on dense reward gradients at every resampling step. Because exact gradients are rarely available in practice, relying on approximations inevitably injects systematic bias into the sampling trajectory.
Although numerous gradient-free alignment methods circumvent the need for reward gradients via best-of-N sampling~\citep{stiennon2020learning,beirami2024theoretical}, this generate-and-filter paradigm incurs severe computational overhead. Furthermore, because intervention occurs exclusively post-hoc at the final output state, these approaches fundamentally fail to actively steer the intermediate generative trajectory—a capability essential for feedback-efficient sampling.
\vspace{-3pt}
\paragraph{Conclusion}
We introduce BFMT, a principled Flow-Map Tree sampler tailored for efficient feedback-efficient search and alignment. Crucially, BFMT enables compute-efficient dynamic search by constructing arbitrary-transition search trajectories with only a single NFE. Extensive experiments show BFMT's effectiveness and scalability across a diverse range of modalities.

\newpage

\subsection*{AI use statement}
We used large language models (LLMs) solely as a writing assistant for language polishing and improving clarity of presentation. The LLMs were not involved in research ideation, methodological design, experimental execution, or result analysis. All scientific contributions and substantive writing were carried out by the authors.

\subsection*{Reproducibility statement}
To facilitate reproducibility, we have provided full implementation details, hyperparameter settings, and architectural specifications. Our complete source code, including training scripts and evaluation pipelines, is available via the zip file included in the supplementary material. Upon acceptance, we will make all code publicly available to facilitate easy reproduction.

\bibliography{main}
\bibliographystyle{iclr2027_conference}

\newpage
\appendix
\newcommand{\ldotsfill}{\leavevmode\leaders\hbox to .5em{\hss.\hss}\hfill\kern0pt}

\section*{BFMT: Enhancing Search Capabilities of Tree Sampler via \underline{B}ootstrap \underline{F}low-\underline{M}ap \underline{T}ree (Appendix)}
\vspace{3pt}

\section*{Appendix Table of Contents}
\hrule
\vspace{0.4cm}

\begin{description}
    \item[\textbf{Appendix: Theoretical Derivations}] \hfill \textbf{Page}
    \begin{itemize}
        \item[1] Derivation of BFMT Convergence Rate \dotfill \pageref{sec:prop1-a}
        \item[2] Bootstrap Sufficient Statistic Realizes DDPM-Like Stochastic Path \dotfill \pageref{sec:prop2-a}
        \item[3] Derivation of Equation~\ref{eq:policy} \dotfill \pageref{sec:proof-dts}
    \end{itemize}

    \item[\textbf{Appendix: Quantitative Results on the Quantity-Aware Alignment Task}] \hfill
    \begin{itemize}
        \item[1] Detailed Results on the Quantity Aware Alignment Task \dotfill \pageref{sec:se-quant}
        \item[2] Additional Quantitative Comparisons with Baselines on the Quantity-aware Alignment Task\dotfill \pageref{sec:mean-quant}
    \end{itemize}

    \item[\textbf{Appendix: Additional Ablation Studies and Qualitative Analysis of BFMT}] \hfill
    \begin{itemize}
        \item[1] Visualizing the Impact of Bootstrap Sufficient Statistic \dotfill \pageref{sec:BSS-a}
        \item[2] A Visual Illustration of BFMT's Exploratory Search Strategy \dotfill \pageref{sec:vis-hi}
        \item[3] Additional Visual Illustration of BFMT’s Exploratory Search Strategy
        \dotfill 
        \pageref{sec:exp-a}
        \item[4] Hierarchical Exploration Strategy of BFMT
        \dotfill \pageref{sec:hie-vis-b}
        \item[5] Additional Visualizations of Hierarchical Exploration Strategy of BFMT
        \dotfill \pageref{sec:hie-vis-a}
    \end{itemize}

    \item[\textbf{Appendix: Additional Implementation Details, Results, and Analysis}] \hfill
    \begin{itemize}
        \item[1] Derivation of Noise Schedule to Adapt IMPFM on Alignment Tasks \dotfill \pageref{sec:mean-a}
        \item[2] Details of Class and Prompt Detail in the Search Experiment \dotfill \pageref{sec:prompt-a}
        \item[3] Details about the Evaluation Dataset for Alignment Tasks \dotfill \pageref{sec:align-eval-data}
        \item[4] Additional Results With Molecule Generation Task \dotfill \pageref{sec:mole-add}
        \item[5] Implementation Details \dotfill \pageref{sec:imp-a}
    \end{itemize}

    \item[\textbf{Appendix: Additional  Qualitative Visualizations on Search and Alignment Task}] \hfill
    \begin{itemize}
        \item[1] Additional Comparative Search Visualizations \dotfill \pageref{sec:search-vis-a}
        \item[2] Additional Competitive Alignment Visualization \dotfill \pageref{sec:align-vis-a}
    \end{itemize}
\end{description}

\vspace{0.2cm}
\hrule

\newpage

\section{Proof of Proposition 4.3}\label{sec:prop1-a}
\begin{tcolorbox}[colback=blue!5!white, colframe=gray!80!black, title={\textbf{Proposition 4.3} (BFMT Convergence Rate)}]
Let $r(x_0)$ be bounded and $L$-Lipschitz continuous, $M$ be the feedback budget, and $T$ be the maximum tree depth. Under standard regularity assumptions, the Total Variation distance between the BFMT sampling policy $\hat{q}_M(x_0)$ and the target optimal policy $\pi^*(x_0)$ is bounded:$$D_{TV}(\hat{q}_M \parallel \pi^*) \le \mathcal{O}\left( \beta T^2 M^{-1/4} \right)$$
\end{tcolorbox}

\begin{proof}
Let $M$ be the total number of trajectory rollouts starting at the root, and $T$ be the maximum depth of the tree.  We assume that the reward function is bounded, $r(x) \in [0, R_{max}]$. Consequently, the true soft value function is also bounded: $V_t(x) \in [0, R_{max}]$ for all $t$. We also assume that the transition kernel $p_\theta(x_{t-1} \mid x_t)$ and the reward function $r(x)$ are $L$-Lipschitz continuous with respect to the state space. 
\paragraph{Step 1: Bound
The Single-Node Value Estimation Error: At any intermediate node $x_t$ visited
$N$ times:}
BFMT utilizes progressive widening to limit the continuous branching factor to $B = \lceil C N^\alpha \rceil$ for some constant $C$ and $\alpha \in (0,1)$. The empirical soft value estimate is defined as:$$\hat{V}_t(x_t) = \frac{1}{\beta} \log \left( \frac{1}{B} \sum_{i=1}^B \exp\left(\beta \> \hat{V}_{t-1}(x_{t-1}^{(i)})\right) \right)$$
The true soft value is defined as: $$V_t(x_t) = \frac{1}{\beta} \log \mathbb{E}_{x_{t-1} \sim \mathrm{BSS}^{1}\left( x_0 \sim p_{0|t}(\cdot \mid x_t) \right)} \left[ \exp\left(\beta \> V_{t-1}(x_{t-1})\right) \right]$$The error $\epsilon(N) = |\hat{V}_t(x_t) - V_t(x_t)|$ at this node stems from two distinct sources:
\begin{itemize}
    \item \emph{Discretization Error ($\epsilon_{disc}$): Approximating the continuous integral over $\mathrm{BSS}^1(x_0 \sim p_{0|t} (\cdot | x_t))$ with a finite sum of $B$ branches.} \emph{Under the Lipschitz assumption, standard Monte Carlo integration error scales as $\mathcal{O}(B^{-1/2})$}. Next, we prove this statement. \\
    Firstly, the Monte-Carlo estimator is unbiased. To show this, let's assume, $f(x_{t-1}) = \exp\left(\beta \> V_{t-1}(x_{t-1})\right)$, $\hat{p}_{\hat{v}}(x_{t-1}|x_t):= \mathrm{BSS}^1(x_0 \sim p_{0|t} (\cdot | x_t))$. Specifically, we are looking at the integral:$$I = \mathbb{E}_{x_{t-1} \sim \hat{p}_{\hat{v}}(\cdot)} [f(x_{t-1})] = \int f(x_{t-1}) \hat{p}_{\hat{v}}(x_{t-1}) dx_{t-1}$$
    Now, our Monte Carlo estimator using $B$ samples $x_{t-1}^{(1)}, x_{t-1}^{(2)}, \dots, x_{t-1}^{(B)} \sim \hat{p}_{\hat{v}}(\cdot)$ is:$$\hat{I}_B = \frac{1}{B} \sum_{i=1}^B f(x_{t-1}^{(i)})$$
    $$\mathbb{E}[\hat{I}_B] = \mathbb{E} \left[ \frac{1}{B} \sum_{i=1}^B f(x_{t-1}^{(i)}) \right]$$By linearity of expectation:$$\mathbb{E}[\hat{I}_B] = \frac{1}{B} \sum_{i=1}^B \mathbb{E}[f(x_{t-1}^{(i)})] = \frac{1}{B} \sum_{i=1}^B I = I$$
    which shows that the Monte-Carlo estimator is unbiased. \\
    Secondly, because \emph{the estimator is unbiased, the mean squared error (MSE) equals the variance of the estimator}. We must first ensure that the variance of a single sample, $\text{Var}(f(x_{t-1}))$, is finite. As we assume bounded rewards: $V_{t-1}(x_{t-1}) \in [0, R_{max}]$, therefore, $f(x_{t-1}) = \exp(\beta V_{t-1}(x_{t-1}))$ is also bounded:$$1 \le f(x_{t-1}) \le \exp(\beta \> R_{max})$$
    Both the transition kernel and the reward function are Lipschitz continuous. Since $V_{t-1}$ is derived from Lipschitz rewards and transitions, it is Lipschitz continuous. The composition of a Lipschitz function with an exponential function (over a bounded domain) remains Lipschitz. Because $f(x_{t-1})$ is bounded on a compact domain, its true variance $\sigma^2$ must be finite:$$\sigma^2 = \text{Var}(f(x_{t-1})) = \mathbb{E}[f(x_{t-1})^2] - (\mathbb{E}[f(x_{t-1})])^2 < \infty$$
    Specifically, since $f(x_{t-1}) \in [1, \exp(\beta \> R_{max})]$, we can crudely bound the variance using Popoviciu's inequality:$$\sigma^2 \le \frac{(\exp(\beta \> R_{max}) - 1)^2}{4}$$
    We now calculate the variance of the estimator using the average of our $B$ samples.$$\text{Var}(\hat{I}_B) = \text{Var} \left( \frac{1}{B} \sum_{i=1}^B f(x_{t-1}^{(i)}) \right)$$Because the $B$ samples are drawn independently, the variance of the sum is the sum of the variances:$$\text{Var}(\hat{I}_B) = \frac{1}{B^2} \sum_{i=1}^B \text{Var}(f(x_{t-1}^{(i)}))$$
    Since all samples are drawn from the same distribution, they all have the same variance $\sigma^2$:$$\text{Var}(\hat{I}_B) = \frac{1}{B^2} (B \cdot \sigma^2) = \frac{\sigma^2}{B}$$
    The expected estimation error is typically quantified by the Root Mean Square Error (RMSE), which is the standard deviation of the estimator.$$\text{RMSE}(\hat{I}_B) = \sqrt{\text{Var}(\hat{I}_B)} = \sqrt{\frac{\sigma^2}{B}} = \frac{\sigma}{\sqrt{B}}$$Because $\sigma$ is a finite constant independent of the number of samples $B$, we can express the asymptotic convergence rate in Big-O notation:$$\epsilon_{disc} = \text{Error} = \mathcal{O}\left(\frac{1}{\sqrt{B}}\right) = \mathcal{O}(B^{-1/2})$$
    \item \emph{Statistical Estimation Error ($\epsilon_{stat}$): The downstream values $\hat{V}_{t-1}$ are estimated using finite rollouts. Leveraging Pigeonhole Principle~\citep{rosen1999discrete}, with $N$ total visits distributed across $B$ branches, each child receives a minimum polynomial budget fraction $N(x_t) \ge \Omega(B^\gamma)$, where $\gamma = (1 - \alpha)^t$, which implies $\gamma  > 0$. However, under standard regularity assumptions, we relax the polynomial node visitation count relation and let each child receive $n \approx N/B$ visits. By Hoeffding's inequality on the bounded log-sum-exp operator, this error scales as $\mathcal{O}(n^{-1/2})$}. Next, we prove this statement. \\
    For a specific child node, let $X_1, X_2, \dots, X_n$ represent the outcomes (returns) of the $n$ independent rollouts starting from that node. The empirical value estimate $\hat{V}_{t-1}$ is simply the sample average of these returns:$$\hat{V}_{t-1} = \frac{1}{n} \sum_{i=1}^n X_i$$
    We assume that the reward function—and consequently the true value function $V_{t-1}$—is strictly bounded within the interval $[0, R_{max}]$. Therefore, every individual rollout return $X_i$ is a bounded random variable:$$0 \le X_i \le R_{max} \quad \text{for all } i$$
    Now, Hoeffding's inequality provides a strict upper bound on the probability that the sample mean of independent, bounded random variables deviates from its true expected value by more than a certain margin, $\epsilon$. For random variables $X_i \in [a, b]$, Hoeffding's inequality states:$$P(|\hat{V}_{t-1} - V_{t-1}| \ge \epsilon) \le 2 \exp\left( - \frac{2n\epsilon^2}{(b-a)^2} \right)$$Substituting our specific bounds $a = 0$ and $b = R_{max}$:$$P(|\hat{V}_{t-1} - V_{t-1}| \ge \epsilon) \le 2 \exp\left( - \frac{2n\epsilon^2}{R_{max}^2} \right)$$
    We aim to find the error bound $\epsilon$ that holds with a high probability $1 - \delta$ (where $\delta$ is a very small failure probability, e.g., 0.05).To do this, we set the right side of the inequality to $\delta$:$$2 \exp\left( - \frac{2n\epsilon^2}{R_{max}^2} \right) = \delta$$Now, we solve for $\epsilon$ algebraically, which yields: $$\epsilon = R_{max} \sqrt{\frac{\ln(2/\delta)}{2n}}$$
    For any chosen confidence level $\delta$ (which is fixed), the terms $R_{max}$ and $\sqrt{\ln(2/\delta)/2}$ are just constants. Let's denote this combined constant $K$. We can rewrite the equation as:  $$\epsilon = K \cdot \frac{1}{\sqrt{n}}$$
    Therefore, with high probability $1-\delta$, the estimation error for any single child node shrinks proportionally to the inverse square root of the number of visits $n$, i.e.,$$\epsilon_{stat} = \mathcal{O}(n^{-1/2})$$
    In the BFMT framework, the values of the $B$ children are aggregated using a soft-maximum (the log-sum-exp function). The standard log-sum-exp function, defined as $f(z) = \frac{1}{\beta} \log \sum \exp(\beta z_i)$, is a well-known non-expansive operator.Specifically, it is 1-Lipschitz continuous with respect to the maximum norm (L-infinity norm). This implies that if the inputs (the children's $\hat{V}_{t-1}$ estimates) have a maximum error of $\mathcal{O}(n^{-1/2})$, passing them through the log-sum-exp operator will not amplify the error. The final aggregated statistical error at the parent node remains bounded by $\mathcal{O}(n^{-1/2})$.
\end{itemize}

Combining these via the triangle inequality:$$\epsilon(N) \le \mathcal{O}\left( \frac{1}{\sqrt{B}} \right) + \mathcal{O}\left( \frac{1}{\sqrt{N/B}}  \right); n \approx \frac{N}{B}$$Substituting $B = C N^\alpha$:$$\epsilon(N) \le \mathcal{O}\left( N^{-\alpha/2} \right) + \mathcal{O}\left( N^{-(1-\alpha)/2} \right)$$To find the tightest bound, we balance the two error terms by setting $\alpha = 0.5$. This yields a single-node convergence rate of:
\begin{equation}
  \epsilon(N) \le \mathcal{O}\left( N^{-1/4} \right).  
\end{equation}

\paragraph{Step 2: Backward Induction for Error Propagation:}
We must now propagate this error from the leaves ($t=0$) up to the root ($t=T$). Let $\epsilon_t$ denote the maximum absolute error in the value estimate at depth $t$. By the definition of the Bellman backup, the error at depth $t$ is bounded by the intrinsic estimation error at that node plus the maximum inherited error from its children at depth $t-1$:$$\epsilon_t \le \epsilon_{t-1} + \mathcal{O}\left( N_t^{-1/4} \right)$$
Because BFMT focuses its search budget $M$ along the highest-value trajectories (according to the BASE selection mechanism), the visitation count $N_t$ along the sampled trajectory remains proportional to $M$, i.e., $N_t \approx cM$ for some constant $c \in (0,1]$. Unrolling the recursion from $t=T$ to $t=0$, we sum the errors across the horizon:$$\sup_{x_{t}} |\hat{V}_t(x_t) - V_t(x_t)| \le \sum_{t=1}^T \mathcal{O}\left( (cM)^{-1/4} \right) = \mathcal{O}\left( T M^{-1/4} \right)$$

\paragraph{Step 3: Bound the Step-Wise Policy Divergence:} 
The BFMT sampling policy at timestep $t$ is an empirically perturbed Boltzmann distribution:$$\hat{q}(x_{t-1} \mid x_t) = \frac{\hat{p}_{\hat{v}}(x_{t-1} \mid x_t) \exp(\beta \> \hat{V}_{t-1}(x_{t-1}))}{\int \hat{p}_{\hat{v}}(x'_{t-1} \mid x_t) \exp(\beta \> \hat{V}_{t-1}(x'_{t-1})) dx'}$$We bound the Total Variation distance between this empirical policy and the optimal target policy $\pi^*(x_{t-1} \mid x_t)$. For any two Boltzmann distributions over the same base measure differing only by their potential functions, the TV distance is bounded by the maximum absolute difference between those potentials multiplied by $\beta$: $$D_{TV}(\hat{q}(x_{t-1}| x_t) \parallel \pi^*_t(x_{t-1}|x_t)) \le 2 \beta \sup_{x_{t-1}} |\hat{V}_{t-1}(x_{t-1}) - V_{t-1}(x_{t-1})|$$. Next, we present the detailed derivation of the above statement. 

Let $\hat{p}_{\hat{v}}(x_{t-1}|x_t)$ be a shared base measure. We define two Boltzmann distributions, the empirical policy $\hat{q}(x_{t-1}|x_t)$ and the target policy $\pi^*(x_{t-1}|x_t)$, parameterized by a temperature scalar $\beta$ and potential functions $\hat{V}_{t-1}(x_{t-1})$ and $V_{t-1}(x_{t-1})$, respectively:$$\hat{q}(x_{t-1}|x_t) = \frac{\hat{p}_{\hat{v}}(x_{t-1}|x_t) \exp(\beta \hat{V}_{t-1}(x_{t-1}))}{\hat{Z}} \quad \text{and} \quad \pi^*(x_{t-1}|x_t) = \frac{\hat{p}_{\hat{v}}(x_{t-1}|x_t) \exp(\beta V_{t-1}(x_{t-1}))}{Z^*}$$where $\hat{Z}$ and $Z^*$ are the respective normalizing constants (partition functions). Let $\epsilon = \sup_{x_{t-1}} |\hat{V}_{t-1}(x_{t-1}) - V_{t-1}(x_{t-1})|$ be the maximum absolute difference between the potential functions. We aim to prove that:$$D_{TV}(\hat{q}(x_{t-1}| x_t) \parallel \pi^*_t(x_{t-1}|x_t)) \le 2 \beta \epsilon$$
To this end, Let $\Delta(x_{t-1}) = \hat{V}(x_{t-1}) - V(x_{t-1})$ be the difference between the potential functions at any point $x_{t-1}$. By definition, $-\epsilon \le \Delta(x_{t-1}) \le \epsilon$.
Multiplying by the temperature parameter $\beta$ and exponentiating gives us strict bounds on the ratio of the unnormalized exponential terms:$$e^{-\beta \epsilon} \le \exp(\beta \Delta(x_{t-1})) \le e^{\beta \epsilon}$$
\begin{equation}\label{eq:ratio}
    e^{-\beta \epsilon} \le \frac{\exp(\beta \hat{V}_{t-1}(x_{t-1}))}{\exp(\beta V_{t-1}(x_{t-1}))} \le e^{\beta \epsilon}
\end{equation}

The partition function $Z^*$ for the true distribution is defined as $Z^* = \int \hat{p}_{\hat{v}}(x_{t-1}|x_t) \exp(\beta V_{t-1}(x_{t-1})) dx_{t-1}$. The partition function $\hat{Z}$ for the empirical distribution is:$$\hat{Z} = \int \hat{p}_{\hat{v}}(x_{t-1}|x_t) \exp(\beta \hat{V}_{t-1}(x_{t-1})) dx_{t-1} = \int \hat{p}_{\hat{v}}(x_{t-1}|x_t) \exp(\beta V_{t-1}(x_{t-1})) \exp(\beta \Delta(x_{t-1})) dx_{t-1}$$
Using our bounds from Eqn.~\ref{eq:ratio}, we can factor the maximum deviations out of the integral:$$\hat{Z} \le \int \hat{p}_{\hat{v}}(x_{t-1}|x_t) \exp(\beta V_{t-1}(x_{t-1})) e^{\beta \epsilon} dx_{t-1} = e^{\beta \epsilon} Z^*$$$$\hat{Z} \ge \int \hat{p}_{\hat{v}}(x_{t-1}|x_t) \exp(\beta V_{t-1}(x_{t-1})) e^{-\beta \epsilon} dx_{{t-1}} = e^{-\beta \epsilon} Z^*$$Rearranging these inequalities gives us bounds on the ratio of the two normalizing constants:$$e^{-\beta \epsilon} \le \frac{Z^*}{\hat{Z}} \le e^{\beta \epsilon}$$
We can now evaluate the ratio of the two full probability distributions $\frac{\hat{q}(x_{t-1}|x_t)}{\pi^*(x_{t-1}|x_t)}$:$$\frac{\hat{q}(x_{t-1}|x_t)}{\pi^*(x_{t-1}|x_t)} = \frac{\exp(\beta \hat{V}_{t-1}(x_{t-1}))}{\exp(\beta V_{t-1}(x_{t-1}))} \cdot \frac{Z^*}{\hat{Z}}$$
Since we have lower bounds for both fractions on the right side, we can multiply them to find the minimum possible ratio between the two distributions:$$\frac{\hat{q}(x_{t-1}|x_t)}{\pi^*(x_{t-1}|x_t)} \ge e^{-\beta \epsilon} \cdot e^{-\beta \epsilon} = e^{-2\beta \epsilon}$$
This implies that for all $x_{t-1}$:
\begin{equation}\label{eq:tv}
   \hat{q}(x_{t-1}|x_t) \ge e^{-2\beta \epsilon} \pi^*(x_{t-1}|x_t) 
\end{equation}
An alternative but mathematically equivalent definition of Total Variation distance between two probability measures is:$$D_{TV}(\hat{q}(x_{t-1}|x_t) \parallel \pi^*(x_{t-1}|x_t)) = 1 - \int \min(\hat{q}(x_{t-1}|x_t), \pi^*(x_{t-1}|x_t)) dx_{t-1}$$
Using the inequality established in~\ref{eq:tv}, we know that $\min(\hat{q}(x_{t-1}|x_t), \pi^*(x_{t-1}|x_t))$ must be at least $e^{-2\beta \epsilon} \pi^*(x_{t-1}|x_t)$. 
Therefore:$$\int \min(\hat{q}(x_{t-1}|x_t), \pi^*(x_{t-1}|x_t)) dx_{t-1} \ge \int e^{-2\beta \epsilon} \pi^*(x_{t-1}|x_t) dx_{t-1} = e^{-2\beta \epsilon} \int \pi^*(x_{t-1}|x_t) dx_{t-1}$$Because $\pi^*(x_{t-1}|x_t)$ is a valid probability distribution, its integral over the entire domain is exactly 1.
$$\int \min(\hat{q}(x_{t-1}|x_t), \pi^*(x_{t-1}|x_t)) dx \ge e^{-2\beta \epsilon}$$Substitute this back into the TV distance definition:$$D_{TV}(\hat{q}(x_{t-1}|x_t) \parallel \pi^* (x_{t-1}|x_t)) \le 1 - e^{-2\beta \epsilon}$$
For any $y \ge 0$, the exponential inequality $1 - e^{-y} \le y$ holds true. Setting $y = 2\beta \epsilon$:$$1 - e^{-2\beta \epsilon} \le 2\beta \epsilon$$
Thus, we conclude that:$$D_{TV}(\hat{q} (x_{t-1}|x_t)\parallel \pi^*(x_{t-1}|x_t)) \le 2\beta \epsilon = 2\beta \sup_{x_{t-1}} |\hat{V}_{t-1}(x_{t-1}) - V_{t-1}(x_{t-1})|$$

Substituting our propagated value error from \textbf{Step 2}:$$D_{TV}(\hat{q}(x_{t-1}|x_t) \parallel \pi^*(x_{t-1}|x_t)) \le \mathcal{O}\left( \beta T M^{-1/4} \right)$$  

\paragraph{Step 4: Full Trajectory Divergence:} 
Finally, we bound the divergence of the entire generated sample $x_0$. Following~\citep{benton2024nearly}, and by using the chain rule for Total Variation distance over Markov chains, the total error is at most the sum of the step-wise errors over the $T$ steps of the diffusion horizon:$$D_{TV}(\hat{q}_M(x_0) \parallel \pi^*(x_0)) \le \sum_{t=1}^T D_{TV}(\hat{q}(x_{t-1}|x_t)) \parallel \pi^*
(x_{t-1}|x_t))$$$$D_{TV}(\hat{q}_M \parallel \pi^*) \le T \times \mathcal{O}\left( \beta T M^{-1/4} \right) = \mathcal{O}\left( \beta T^2 M^{-1/4} \right)$$
This completes the proof.
\end{proof}

\section{Proof of Proposition 4.1}\label{sec:prop2-a}

\begin{tcolorbox}[colback=blue!5!white, colframe=gray!80!black, title={\textbf{Proposition 4.1} (Bootstrap Sufficient Statistic Realizes DDPM-Like Stochastic Path)}]
Let $0< t''<t' < t$ be any two consecutive transition time steps in a trajectory, and $p^{\text{DDPM}}_{t''|t'}$ be the DDPM transition kernel. Then for $x_{t''}$ obtained via~\ref{eq:bss}, starting from $x_t$:
\begin{equation}
    x_{t''} \sim p^{\text{DDPM}}_{t''|t'}(\cdot \mid x_t)
\end{equation}
\end{tcolorbox}

\begin{proof}
We present the proof in two steps.
\paragraph{Step 1: We first show that starting at $x_t$, the children of $x_t$ ( i.e., $x_{t^{child}}$) (i.e., equivalent to a Single transition) obtained via~\ref{eq:bss}, is a valid DDPM transition.} 

Given $x_t$ as the starting node of the trajectory, and $x_{t^{child}}$ is the next node in the trajectory. We first compute $r^{*} = g(x_t, x_{t^{child}})$, then for $\epsilon \sim \mathcal{N}(0, I_d)$,  we define the random variable:
\begin{equation}
    \tilde{X}_{t^{child}} = \alpha_{t^{child}} S_{r^*,t}(x_{r^*},x_t)
\end{equation}

We aim to prove that:
\begin{equation}\label{eq:obj}
    \tilde{p}(\tilde{X}_{t^{child}} = x_{t^{child}} | X_t = x_t) = p^{\text{DDPM}}_{t^{child}|t}(X_{t^{child}} = x_{t^{child}} | X_t = x_t)
\end{equation}
Where the forward process of DDPM is defined via a transition kernel $q_{t|t'}$ in reverse-time given by:
\begin{equation}
    q_{t|t^{child}}(x_t|x_{t^{child}}) = \mathcal{N} \left( \frac{\alpha_t}{\alpha_{t^{child}}} x_{t^{child}}; \left( \sigma_t^2 - \frac{\alpha_t^2}{\alpha_{t^{child}}^2} \sigma_{t^{child}}^2 \right) I \right), \quad (t < t^{child})
\end{equation}

The time-reversal SDE or DDPM transitions are then given by:
\begin{equation}
    p^{\text{DDPM}}_{t^{child}|t}(x_{t^{child}}|x_t) = q_{t|t^{child}}(x_t|x_{t^{child}}) \frac{p_{t^{child}}(x_{t^{child}})}{p_t(x_t)}
\end{equation}
In other words, we must demonstrate that the conditional distribution of $\tilde{X}_{t^{child}}$ given $X_t = x_t$ exactly matches the DDPM transition kernel $p^{\text{DDPM}}_{t^{child}|t}$. To establish this, it suffices to show that their corresponding joint distributions are identical:
\begin{equation} \label{eq:joint_coincide} 
    \tilde{p}(\tilde{X}_{t^{child}} = x_{t^{child}}, X_t = x_t) = p_{t^{child},t}(X_{t^{child}} = x_{t^{child}}, X_t = x_t)
\end{equation}
for $x_t \sim p_t$. As the DDPM transition of $q_{t|t^{child}}$, we know that:
\begin{equation}
    p_{t,t^{child}}(x_t, x_{t^{child}}) = \int q_{t|t^{child}}(x_t|x_{t^{child}}) p_{t^{child}}(x_{t^{child}}|x_1) p_{\text{data}}(x_1) dx_1
\end{equation}
where $q_{t|t^{child}}(x_t|x_{t^{child}}) = \mathcal{N} \left( \frac{\alpha_t}{\alpha_{t^{child}}} x_{t^{child}}; \left( \sigma_t^2 - \frac{\alpha_t^2}{\alpha_{t^{child}}^2} \sigma_{t^{child}}^2 \right) I \right)$ and $p_{t^{child}}(x_{t^{child}}|x_1) = \mathcal{N}(\alpha_{t^{child}} x_1; \sigma_{t^{child}}^2 I)$.

Now, we get a sample $(X_t, X_{t^{child}}) \sim p_{t,t^{child}}(x_t, x_{t^{child}})$ via:
\begin{align*}
    X_1 &\sim p_{\text{data}}, \\
    X_{t^{child}} &= \alpha_{t^{child}} X_1 + \sigma_{t^{child}} \epsilon_1, \\
    X_t &= \frac{\alpha_t}{\alpha_{t^{child}}} X_{t^{child}} + \sqrt{\sigma_t^2 - \frac{\alpha_t^2}{\alpha_{t^{child}}^2} \sigma_{t^{child}}^2} \, \epsilon_2, \quad \epsilon_1, \epsilon_2 \sim \mathcal{N}(0, I_d)
\end{align*}

Next, to sample from $\tilde{p}(\tilde{x}_{t^{child}}, x_t)$, we can perform the following steps : Sample $X_1 \sim p_{\text{data}}$ first and then renoise to obtain $X_t$ and $X_{r^*}$ independently and then apply stochastic interpolant to obtain $X_{t^{child}}$. Specifically, we perform the following steps:
\begin{align}
    X_1 &\sim p_{\text{data}}, \label{eq:44}  \\
    X_{r^*} &= \bar{\alpha}_{r^*} X_1 + \bar{\sigma}_{r^*} \tilde{\epsilon}_1 \\
    X_t &= \alpha_t X_1 + \sigma_t \tilde{\epsilon}_2   \\
    X_{t^{child}} &= \alpha_{t^{child}} \frac{\bar{\alpha}_{r^*} \sigma_t^2 X_{r^*} + \alpha_t \bar{\sigma}_{r^*}^2 X_t}{\bar{\alpha}_{r^*}^2 \sigma_t^2 + \alpha_t^2 \bar{\sigma}_{r^*}^2} \label{eq:gt}
\end{align}

We obtain~\ref{eq:gt} by following the definition of the GLASS velocity field~\citep{holderrieth2025glass} and GLASS probability path~\citep{holderrieth2025glass}. 
$\text{Here,} \quad \bar{\gamma} = \rho\sigma_{t^{child}}\sigma_t^{-1}, \quad \bar{\alpha} = \alpha_{t^{child}} - \bar{\gamma}\alpha_t, \quad \bar{\sigma}^2 = \sigma_{t^{child}}^2(1 - \rho^2)$. $\rho$ is the correlation parameter.

Next, we aim to show that $(X_t, X_{t^{child}})$ sampled via this procedure results in the same distribution as sampling them via the SDE/DDPM. But it is sufficient if we can show that their distribution coincides conditioned on $X_1$. Therefore, let's fix $X_1 = x_1$. Then, for the DDPM, it holds that:
\begin{equation}
    (X_t, X_{t^{child}}) | X_1 = x_1 \sim \mathcal{N} \left( 
    \begin{pmatrix} 
    \alpha_t x_1 \\ 
    \alpha_{t^{child}} x_1 
    \end{pmatrix}, 
    \begin{pmatrix} 
    \sigma_t^2 & \frac{\alpha_t}{\alpha_{t^{child}}} \sigma_{t^{child}}^2 \\ 
    \frac{\alpha_t}{\alpha_{t^{child}}} \sigma_{t^{child}}^2 & \sigma_{t^{child}}^2 
    \end{pmatrix} 
    \right)
\end{equation}

Similarly, covariance can be derived conditioned on $X_1$ using \eqref{eq:gt} and the fact that $X_t, X_{r^*}$ are independent given $X_1$:
\begin{align*}
    \text{Cov}(X_t, X_{t^{child}} | X_1) &= \alpha_{t^{child}} \frac{\bar{\alpha}_{r^*} \sigma_t^2 \text{Cov}(X_{r^*}, X_t | X_1) + \alpha_t \bar{\sigma}_{r^*}^2 \text{Cov}(X_t, X_t | X_1)}{\bar{\alpha}_{r^*}^2 \sigma_t^2 + \alpha_t^2 \bar{\sigma}_{r^*}^2} \\
    &= \alpha_{t^{child}} \frac{\alpha_t \bar{\sigma}_{r^*}^2 \text{Cov}(X_t, X_t | X_1)}{\bar{\alpha}_{r^*}^2 \sigma_t^2 + \alpha_t^2 \bar{\sigma}_{r^*}^2} \\
    &= \frac{\alpha_{t^{child}} \alpha_t \bar{\sigma}_{r^*}^2 \text{Var}(X_t | X_1)}{\bar{\alpha}_{r^*}^2 \sigma_t^2 + \alpha_t^2 \bar{\sigma}_{r^*}^2} \\
    &= \frac{\alpha_{t^{child}} \alpha_t \bar{\sigma}_{r^*}^2 \sigma_t^2}{\bar{\alpha}_{r^*}^2 \sigma_t^2 + \alpha_t^2 \bar{\sigma}_{r^*}^2}
\end{align*}

We know by definition:
\begin{align*}
    \sigma_{t^{child}}^2 &= \alpha_{t^{child}}^2 g(t^{child}) = \alpha_{t^{child}}^2 g(t^*(r^*, t)) = \alpha_{t^{child}}^2 \frac{\sigma_t^2 \bar{\sigma}_{r^*}^2}{\bar{\alpha}_{r^*}^2 \sigma_t^2 + \alpha_t^2 \bar{\sigma}_{r^*}^2} \\
    &= \frac{\alpha_{t^{child}}}{\alpha_t} \left( \frac{\alpha_{t^{child}} \alpha_t \bar{\sigma}_{r^*}^2 \sigma_t^2}{\bar{\alpha}_{r^*}^2 \sigma_t^2 + \alpha_t^2 \bar{\sigma}_{r^*}^2} \right) \\
    &= \frac{\alpha_{t^{child}}}{\alpha_t} \text{Cov}(X_t, X_{t^{child}} | X_1)
\end{align*}

Similarly, we can show that:
\begin{align*}
    \text{Var}(X_{t^{child}} | X_1) &= \alpha_{t^{child}}^2 \frac{\bar{\alpha}_{r^*}^2 \sigma_t^4 \bar{\sigma}_{r^*}^2 + \alpha_t^2 \bar{\sigma}_{r^*}^4 \sigma_t^2}{(\bar{\alpha}_{r^*}^2 \sigma_t^2 + \alpha_t^2 \bar{\sigma}_{r^*}^2)^2} \\
    &= \alpha_{t^{child}}^2 \left( \frac{\bar{\alpha}_{r^*}^2 \sigma_t^2 + \alpha_t^2 \bar{\sigma}_{r^*}^2}{\bar{\alpha}_{r^*}^2 \sigma_t^2 + \alpha_t^2 \bar{\sigma}_{r^*}^2} \right) \frac{\sigma_t^2 \bar{\sigma}_{r^*}^2}{\bar{\alpha}_{r^*}^2 \sigma_t^2 + \alpha_t^2 \bar{\sigma}_{r^*}^2} \\
    &= \alpha_{t^{child}}^2 \frac{\sigma_t^2 \bar{\sigma}_{r^*}^2}{\bar{\alpha}_{r^*}^2 \sigma_t^2 + \alpha_t^2 \bar{\sigma}_{r^*}^2} \\
    &= \alpha_{t^{child}}^2 g(t^{child}) = \sigma_{t^{child}}^2
\end{align*}

Hence, we can write that:
\begin{equation}
    (X_t, X_{t^{child}}) | X_1 = x_1 \sim \mathcal{N} \left( 
    \begin{pmatrix} 
    \alpha_t x_1 \\ 
    \alpha_{t^{child}} x_1 
    \end{pmatrix}, 
    \begin{pmatrix} 
    \sigma_t^2 & \frac{\alpha_t}{\alpha_{t^{child}}} \sigma_{t^{child}}^2 \\ 
    \frac{\alpha_t}{\alpha_{t^{child}}} \sigma_{t^{child}}^2 & \sigma_{t^{child}}^2 
    \end{pmatrix} 
    \right)
\end{equation}
which precisely matches the distribution of the memoryless SDE conditioned on $X_1$. By marginalizing over $X_1 \sim p_{\text{data}}$, we establish that the joint distributions are identical, fulfilling \eqref{eq:joint_coincide}. This joint equality directly implies the equivalence of the target distributions conditioned on $X_t = x_t$.
\paragraph{Step 2: Proof by Induction}

Having established the relationship between $x_t$ and $x_{t^{child}}$, we now aim to show that the child of $x_{t^{child}}$ obtained via~\ref{eq:bss} also follows the DDPM transition kernel. To this end, we simply consider 

$x_t = x_{t^{child}}$, and the next transition time step in the trajectory is:

$t^{child}$ = time step of the \textbf{child of} $x_{t^{child}}$. 

Assuming this, we can compute the updated $r^* = g(x_t, x_{t^{child}})$, and as before, we can compute $\tilde{X}_{t^{child}} = \alpha_{t^{child}} S_{r^*,t}(x_{r^*},x_t)$. Then, following the results in Step 1, we can conclude that the child of $x_{t^{child}}$ is also a valid DDPM transition from $x_{t^{child}}$. 

By iteratively repeating this procedure as defined in~\ref{eq:bss}, we can conclude that any intermediate node in the trajectory (say $x_{t'}$) has a corresponding child $x_{t''}$ that also follows a valid DDPM transition. In other words, $x_{t''}$ obtained via~\ref{eq:bss}, starting from $x_t$ follows:
\begin{equation}
    x_{t''} \sim p^{\text{DDPM}}_{t''|t'}(\cdot \mid x_t)
\end{equation}

This concludes the proof.

\end{proof}

\section{Derivation of Equation~\ref{eq:policy}}\label{sec:proof-dts}
Let's assume $\hat{p}_{\hat{v}}(x_{t-1}| x_t) = \mathrm{BSS}^{1} (x_0 \sim p_{0|t}(\cdot |x_t))$.

The joint target density over the full chain $(x_0, \dots, x_{t-1}, x_t)$ is given by:
\begin{equation}
    \pi^*(x_0, \dots, x_{t-1}, x_t) = \frac{1}{Z} \hat{p}_{\hat{v}}(x_0, \dots, x_{t-1}, x_t) \exp(\beta r(x_0)),
\end{equation}
where $Z$ represents the normalization constant of this joint density.

The marginal joint density of $(x_t, x_{t-1})$ under $\pi^*$ is:
\begin{equation}
\begin{aligned}
    \pi^*(x_t, x_{t-1}) &= \frac{1}{Z} \int \hat{p}_{\hat{v}}(x_0, \dots, x_{t-1}, x_t) \exp(\beta r(x_0)) \, dx_0 \dots dx_{t-2} \\
    &= \frac{1}{Z} \hat{p}_{\hat{v}}(x_t) \hat{p}_{\hat{v}}(x_{t-1} \mid x_t) \left( \int \hat{p}_{\hat{v}}(x_0, \dots, x_{t-2} \mid x_{t-1}) \exp(\beta r(x_0)) \, dx_0 \dots dx_{t-2} \right) \\
    &= \frac{1}{Z} \hat{p}_{\hat{v}}(x_t) \mathrm{BSS}^{1} (x_0 \sim p_{0|t}(\cdot \mid x_t)) \exp(\beta V_{t-1}(x_{t-1})).
\end{aligned}
\end{equation}

Similarly, the marginal density of $x_t$ under $\pi^*$ is:
\begin{equation}
    \pi^*(x_t) = \frac{1}{Z} \hat{p}_{\hat{v}}(x_t) \exp(\beta V_t(x_t)).
\end{equation}

By dividing these two marginals, we get the transitions under the optimal policy:
\begin{equation}
\begin{aligned}
    \pi^*(x_{t-1} \mid x_t) &= \frac{\pi^*(x_t, x_{t-1})}{\pi^*(x_t)} \\
    &= \frac{\hat{p}_{\hat{v}}(x_{t-1} \mid x_t) \exp(\beta V_{t-1}(x_{t-1}))}{\exp(\beta V_t(x_t))} \\
    &= \frac{\hat{p}_{\hat{v}}(x_{t-1} \mid x_t) \exp(\beta V_{t-1}(x_{t-1}))}{\int \hat{p}_{\hat{v}}(x_{t-1} \mid x_t) \exp(\beta V_{t-1}(x_{t-1})) \, dx_{t-1}} \\
    &= \frac{\mathrm{BSS}^{1} (x_0 \sim p_{0|t}(\cdot |x_t)) \exp(\beta V_{t-1}(x_{t-1}))}{\int \mathrm{BSS}^{1} (x_0 \sim p_{0|t}(\cdot |x_t)) \exp(\beta V_{t-1}(x_{t-1})) \, dx_{t-1}}.
\end{aligned}
\end{equation}

The above relation gives the optimal policy from Equation~\ref{eq:policy}.

\newpage

\section{Detailed Results on the Quantity Aware Alignment Task}\label{sec:se-quant}
\begin{figure}[h]
    \centering
\includegraphics[width=0.60\linewidth]{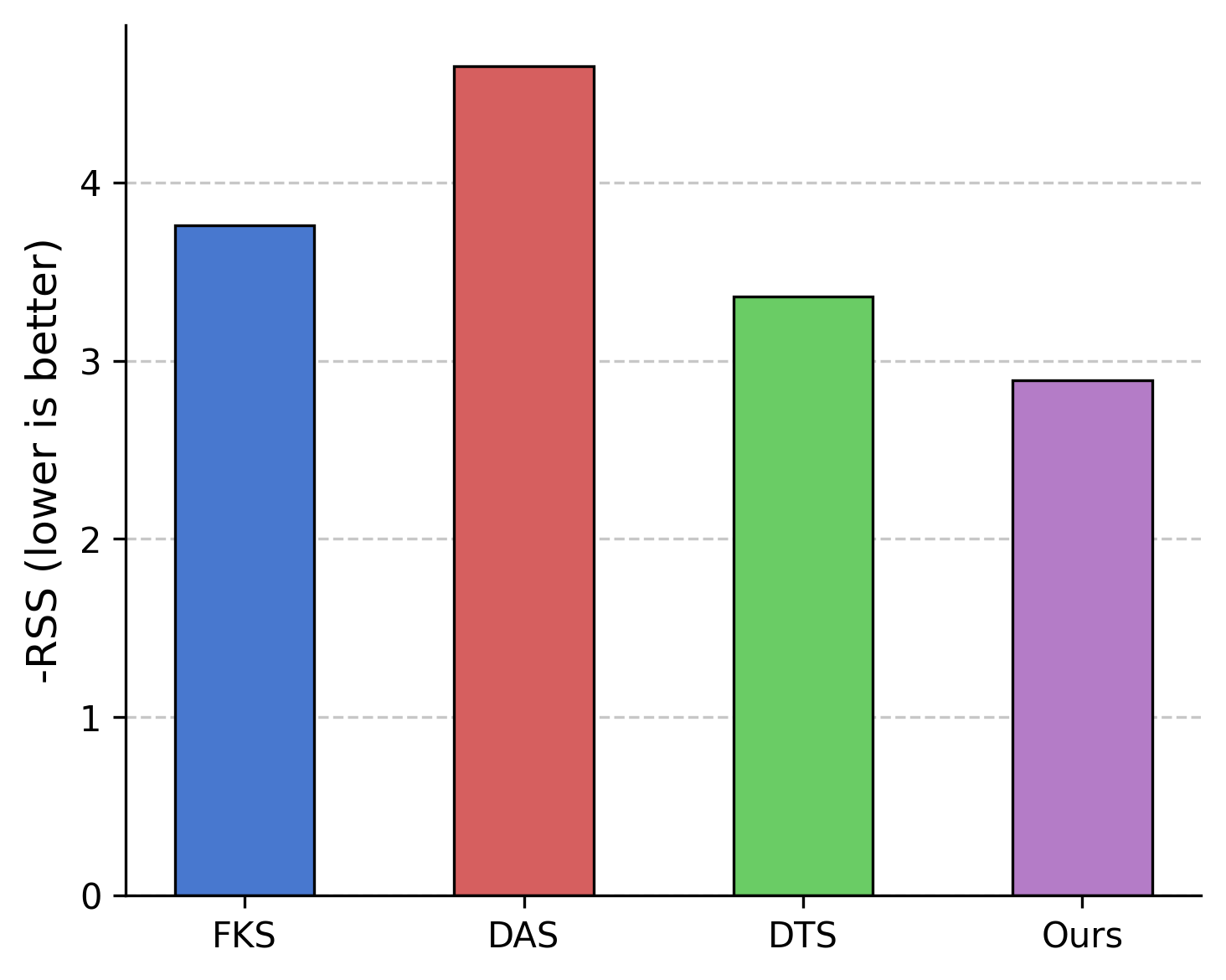}
    \caption{Performance Comparisons on Quantity-Aware Alignment Task.}
    \label{fig:quantity-result}
\end{figure}

Here, we present results for the quantity-aware alignment task. Quantitative comparisons are detailed in Fig.~\ref{fig:quantity-result}. Mirroring the patterns observed in the compositional alignment task, BFMT achieves high reward scores (corresponding to lower RSS scores) using significantly fewer feedback budgets. These findings further validate the efficacy of BFMT for complex, online feedback-driven alignment tasks.

\vspace{4pt}
\paragraph{Additional Quantitative Comparisons with Baselines on the Quantity-aware Alignment Task}\label{sec:mean-quant}
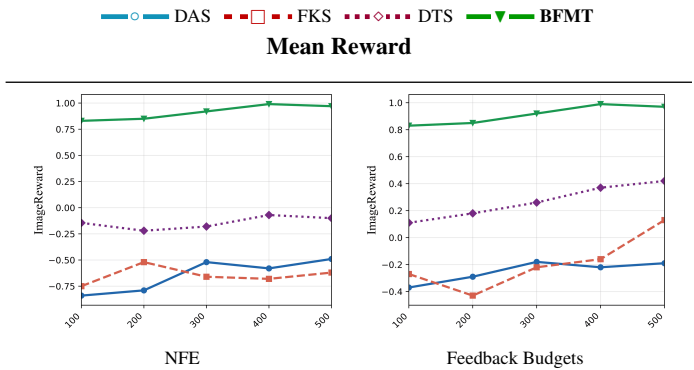
\begin{wrapfigure}{r}{0.65\textwidth}
\centering
\newcommand{\legenditem}[4]{%
  \begin{tikzpicture}[baseline=-0.6ex]
    \draw[#1, line width=1.8pt, #2] (0,0) -- (0.9,0);
    \node[text=#1, fill=white, inner sep=0.6pt] at (0.45,0) {\scriptsize $#3$};
  \end{tikzpicture}\,#4%
}

\newcommand{\methodlegend}{%
  {\scriptsize
  \setlength{\tabcolsep}{2pt}%
  \begin{tabular}{@{}ccccc@{}}
    \legenditem{cyan!70!black}{solid}{\circ}{DAS} &
    \legenditem{red!75!black}{dashed}{\square}{FKS} &
    \legenditem{purple!80!black}{dotted}{\diamond}{DTS} &
    \legenditem{green!60!black}{solid}{\blacktriangledown}{\textbf{BFMT}}
  \end{tabular}%
  }%
}

\newcommand{\barpanel}[2]{%
  \begin{minipage}[t]{0.46\linewidth}
    \centering
    \IfFileExists{#1}{\includegraphics[width=\linewidth,height=0.30\textheight,keepaspectratio]{#1}}{\fbox{\begin{minipage}[c][0.30\textheight][c]{\linewidth}\centering\scriptsize Add\\\detokenize{#1}\end{minipage}}}\par
    \vspace{-0.3ex}
    {\scriptsize #2}
  \end{minipage}%
}

\methodlegend
\vspace{0.2ex}

\noindent
\begin{minipage}[t]{0.90\linewidth}
  \hspace{0.08\linewidth}\centering
  {\small\textbf{Mean Reward}}
\end{minipage}%
\hfill
\begin{minipage}[t]{0.10\linewidth}
  \hspace{-0.08\linewidth}\centering
\end{minipage}

\noindent\rule{\linewidth}{0.5pt}

\barpanel{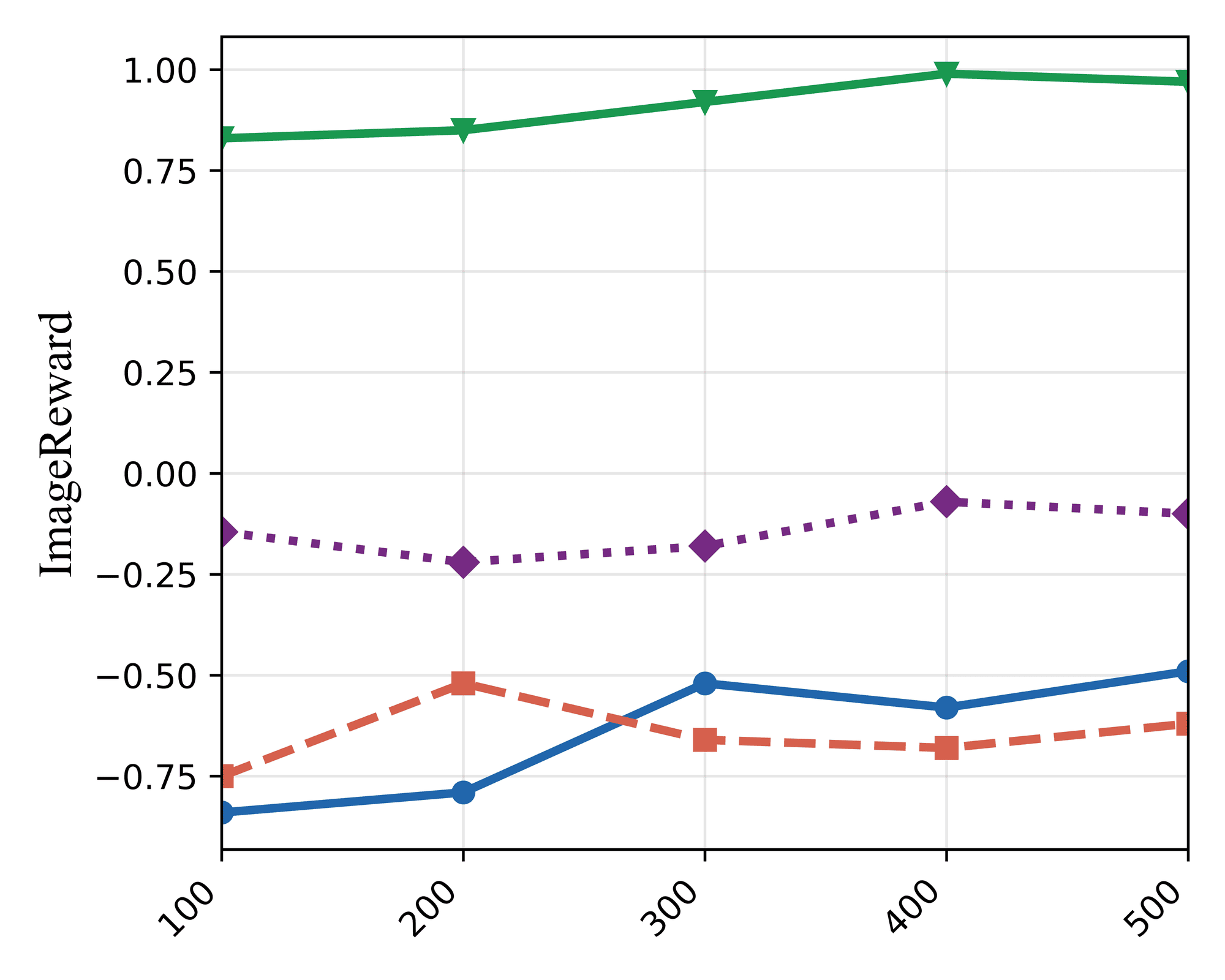}{NFE}\hspace{0.025\linewidth}%
\barpanel{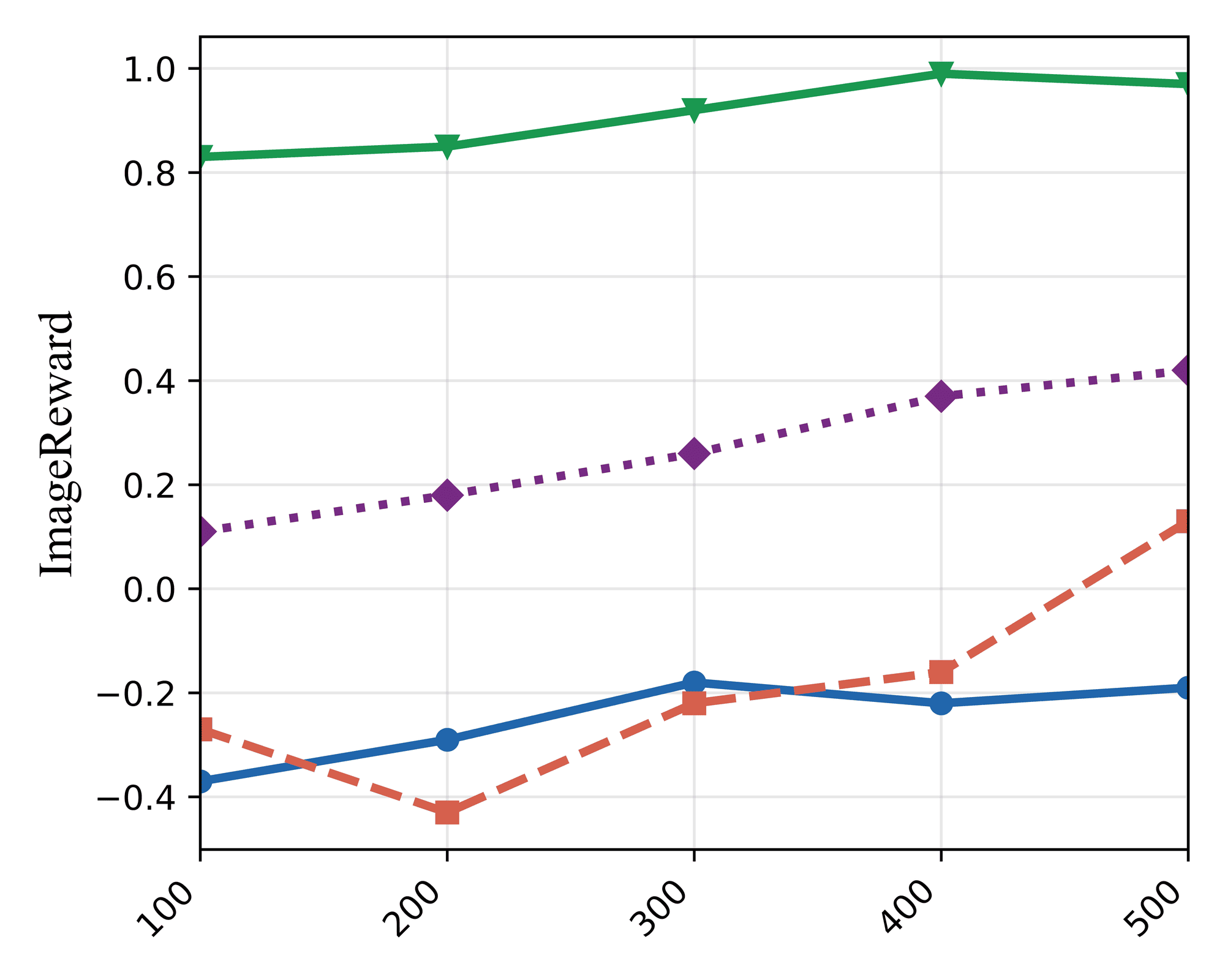}{Feedback Budgets}

\caption{\small{Alignment Performance Analysis on Imagenet Classes.}}
\label{fig:quant-result-mean1}
\end{wrapfigure}
  We also evaluate the alignment performance of BFMT against several baselines in Figure~\ref{fig:quant-result-mean1}, using ImageReward as the evaluation metric. BFMT consistently outperforms all baselines across both NFE and feedback budgets, achieving higher mean reward and further underscoring the effectiveness of the BFMT framework in the online feedback-driven quantity-aware alignment settings. These findings further strengthen the impact of BFMT. 

\vspace{6pt}

\section{Visualizing the Impact of Bootstrap Sufficient Statistic}\label{sec:BSS-a}

\begin{figure*}[h]
\centering
\newcommand{\doublepanel}[2]{%
  \setlength{\tabcolsep}{2pt}%
  \begin{tabular}{@{}c@{\hspace{2pt}}c@{}}
    \includegraphics[width=0.47\linewidth]{#1} &
    \includegraphics[width=0.47\linewidth]{#2}
  \end{tabular}%
}
\newcommand{\panelblock}[3]{%
  \begin{minipage}[t]{\linewidth}
    \centering
    {\itshape ``#1''\par}
    \vspace{0pt}
    \doublepanel{#2}{#3}
  \end{minipage}%
}
\newcommand{\topheaders}{%
  {\fontsize{9.5}{10.5}\selectfont
  \begin{tabular}{@{}c@{\hspace{2pt}}c@{}}
    \makebox[0.47\linewidth][c]{\textbf{BFMT-ISI}} &
    \makebox[0.47\linewidth][c]{\textbf{BFMT}}
  \end{tabular}%
  }
}
\noindent
\begin{minipage}[t]{0.32\textwidth}
\centering
\topheaders
\end{minipage}\hfill%
\begin{minipage}[t]{0.32\textwidth}
\centering
\topheaders
\end{minipage}\hfill%
\begin{minipage}[t]{0.32\textwidth}
\centering
\topheaders
\end{minipage}
\vspace{-0.8ex}
\noindent\rule{\textwidth}{0.6pt}
\vspace{-1.2ex}
\begin{minipage}[t]{0.32\textwidth}
\centering
\panelblock{a red sports car.}{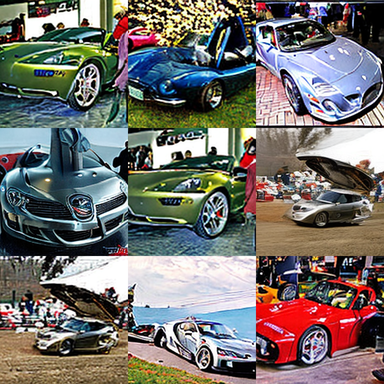}{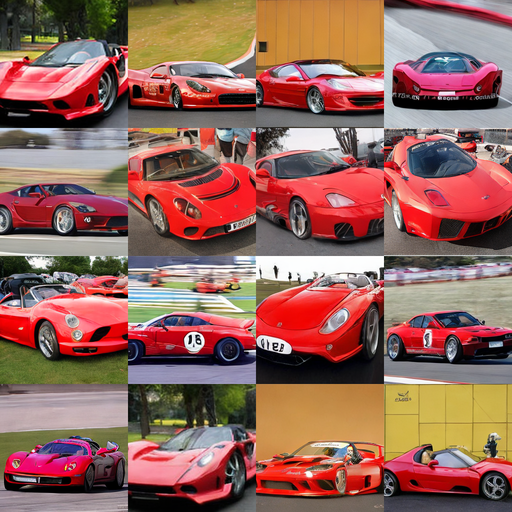}
\end{minipage}\hfill%
\begin{minipage}[t]{0.32\textwidth}
\centering
\panelblock{a black cat.}{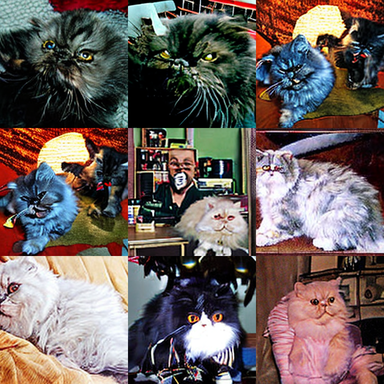}{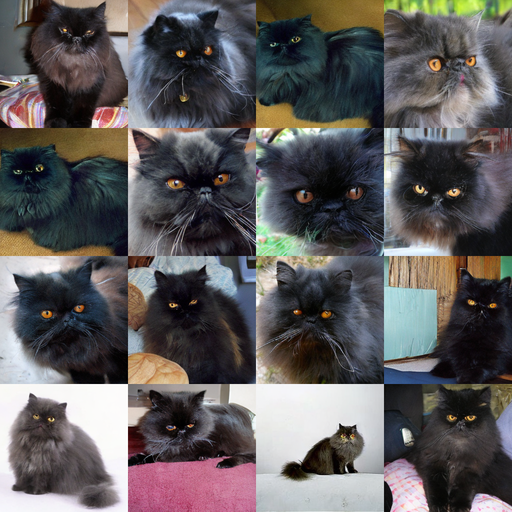}
\end{minipage}\hfill%
\begin{minipage}[t]{0.32\textwidth}
\centering
\panelblock{a red daisy.}{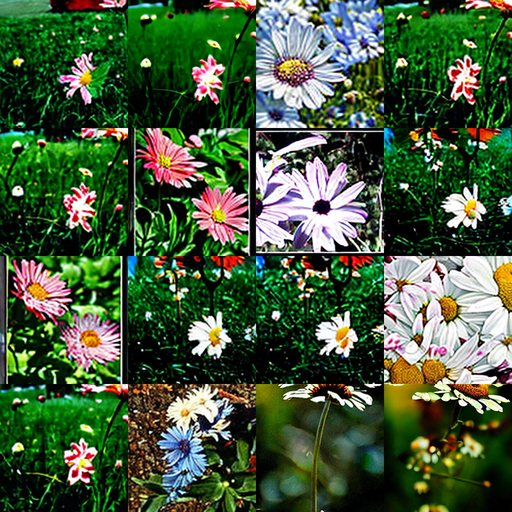}{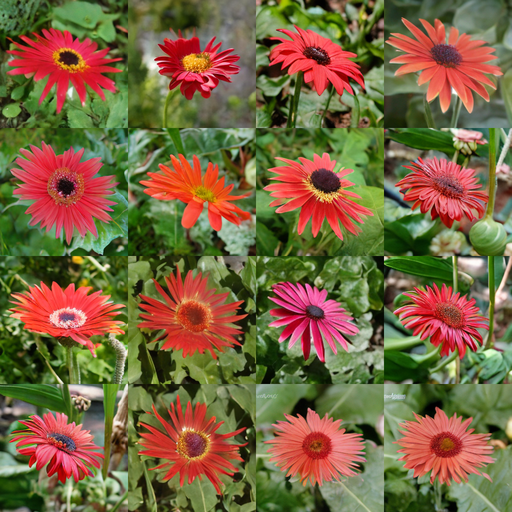}
\end{minipage}%
\vspace{0.6ex}
\noindent\rule{\textwidth}{0.6pt}
\caption{\small{Additional Visualizations on the impact of BSS.}}
\label{fig:bss-add}
\end{figure*}

In the main paper (Section~\ref{pg:bss}), we analyze the impact of the Bootstrap Sufficient Statistic (BSS) both qualitatively and quantitatively. Here, we present additional visualizations highlighting that the BSS mechanism is crucial for generating high-quality, target-aligned samples, particularly under strict NFE constraints. As depicted in Fig.~\ref{fig:bss-add}, an alternative method relying on an iterative stochastic interpolant fails to produce high-quality, aligned images. In contrast, the BSS-based approach succeeds because it guarantees a DDPM-like trajectory rollout, thereby ensuring accurate sampling from the true target distribution. 

\vspace{-4pt}
\section{A Visual Illustration of BFMT's Exploratory Search Strategy}\label{sec:vis-hi}
\begin{figure*}[h]
\centering
{\small\itshape Prompt: ``A Black Cat''\par}
\vspace{0.8ex}

\newcommand{\gridimage}[1]{%
  \IfFileExists{#1}{%
    \includegraphics[width=0.16\textwidth,height=0.16\textwidth,keepaspectratio]{#1}%
  }{%
    \fbox{\begin{minipage}[c][0.16\textwidth][c]{0.16\textwidth}\centering\scriptsize Add\\#1\end{minipage}}%
  }%
}
\newcommand{\feedbackarrow}[1]{%
  \begin{tikzpicture}[x=\textwidth,y=1cm]
    \draw[blue!35!black,line width=1.2pt,-{Stealth[length=2.4mm]}] (0.13,0) -- (0.87,0);
    \node[fill=white,inner sep=1.2pt,font=\scriptsize\bfseries] at (0.50,0) {#1};
  \end{tikzpicture}%
}
\setlength{\tabcolsep}{7pt}%
\renewcommand{\arraystretch}{1.0}%
\begin{tabular}{@{}cccc@{}}
  \gridimage{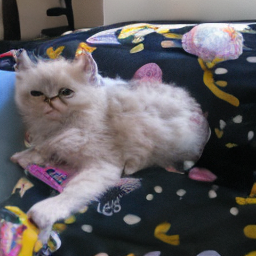} &
  \gridimage{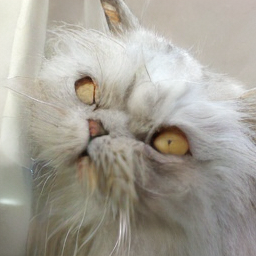} &
  \gridimage{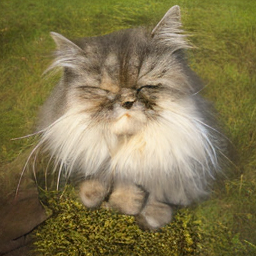} &
  \gridimage{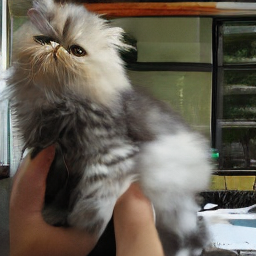} \\
  \multicolumn{4}{@{}c@{}}{\feedbackarrow{Initial Feedback Steps}} \\[0.8ex]
  \gridimage{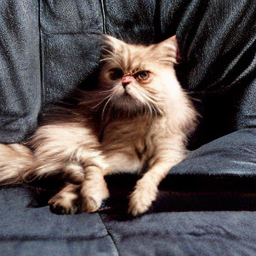} &
  \gridimage{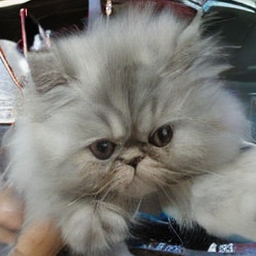} &
  \gridimage{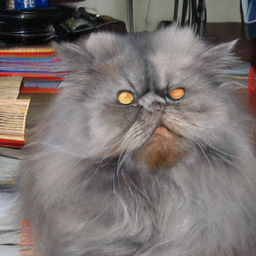} &
  \gridimage{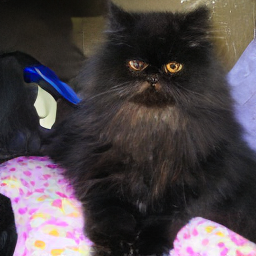} \\
  \multicolumn{4}{@{}c@{}}{\feedbackarrow{Early-Intermediate Feedback Steps}} \\[0.8ex]
  \gridimage{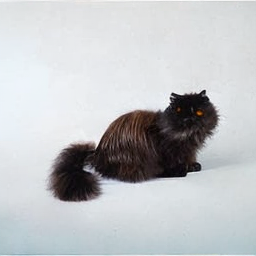} &
  \gridimage{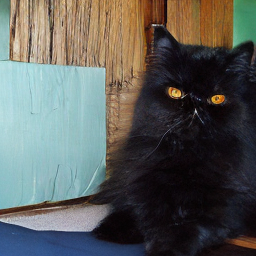} &
  \gridimage{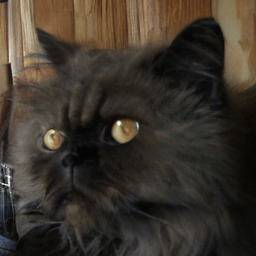} &
  \gridimage{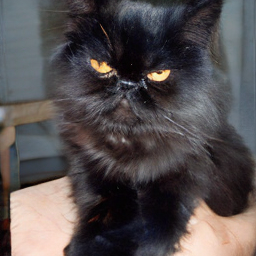} \\
  \multicolumn{4}{@{}c@{}}{\feedbackarrow{Late-Intermediate Feedback Steps}} \\[0.8ex]
  \gridimage{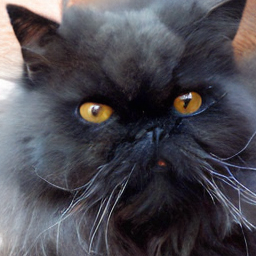} &
  \gridimage{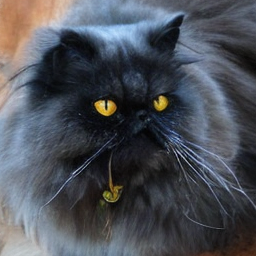} &
  \gridimage{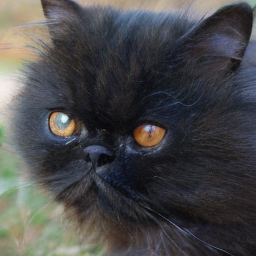} &
  \gridimage{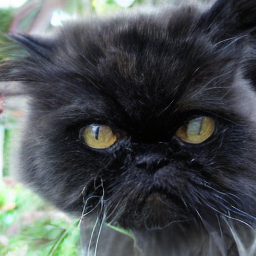} \\
  \multicolumn{4}{@{}c@{}}{\feedbackarrow{Final Feedback Steps}}
\end{tabular}
\caption{\small{Exploration Strategy of BFMT in Online Feedback-Driven Search Settings.}}
\label{fig:spiral-search-trajectory}
\end{figure*}
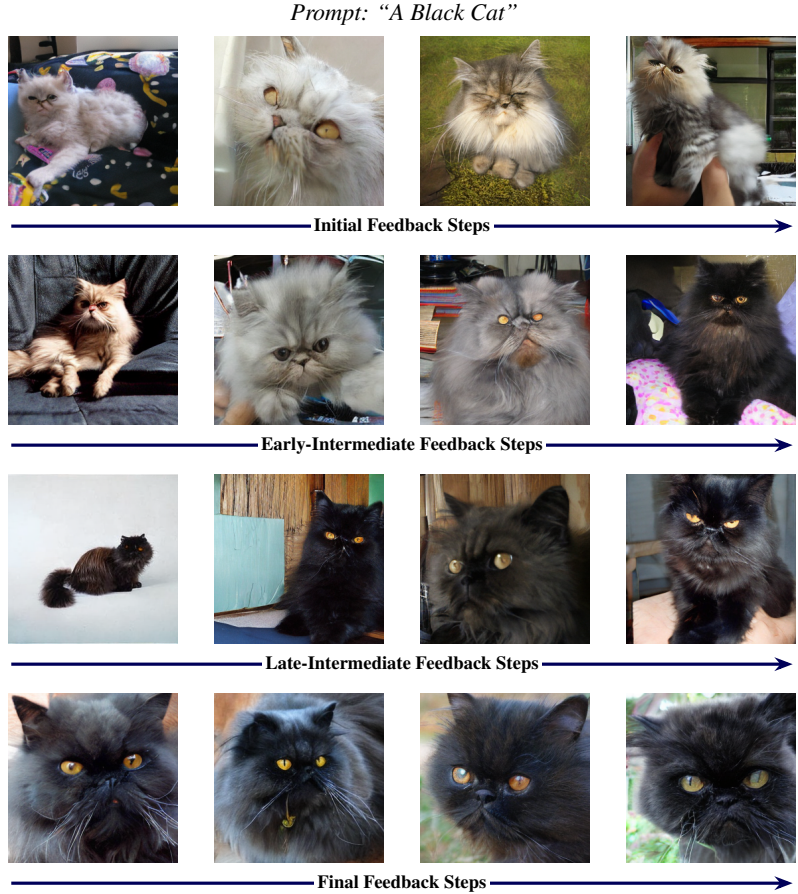

Figure~\ref{fig:spiral-search-trajectory} visualizes the evolution of BFMT's search strategy across different feedback stages. In the early stages, BFMT explores broadly, producing semantically diverse samples within the target category, albeit without precise alignment to the fine-grained target prompt. As feedback accumulates and the search matures, BFMT progressively converges toward samples that closely match the target prompt — reflecting an efficient search process. This exploration-then-exploitation dynamic underscores BFMT's strong exploratory capability, which is fundamental to its success in online feedback-driven settings. 
\vspace{10pt}

\section{Additional Visual Illustration of BFMT's Exploratory Search Strategy}\label{sec:exp-a}
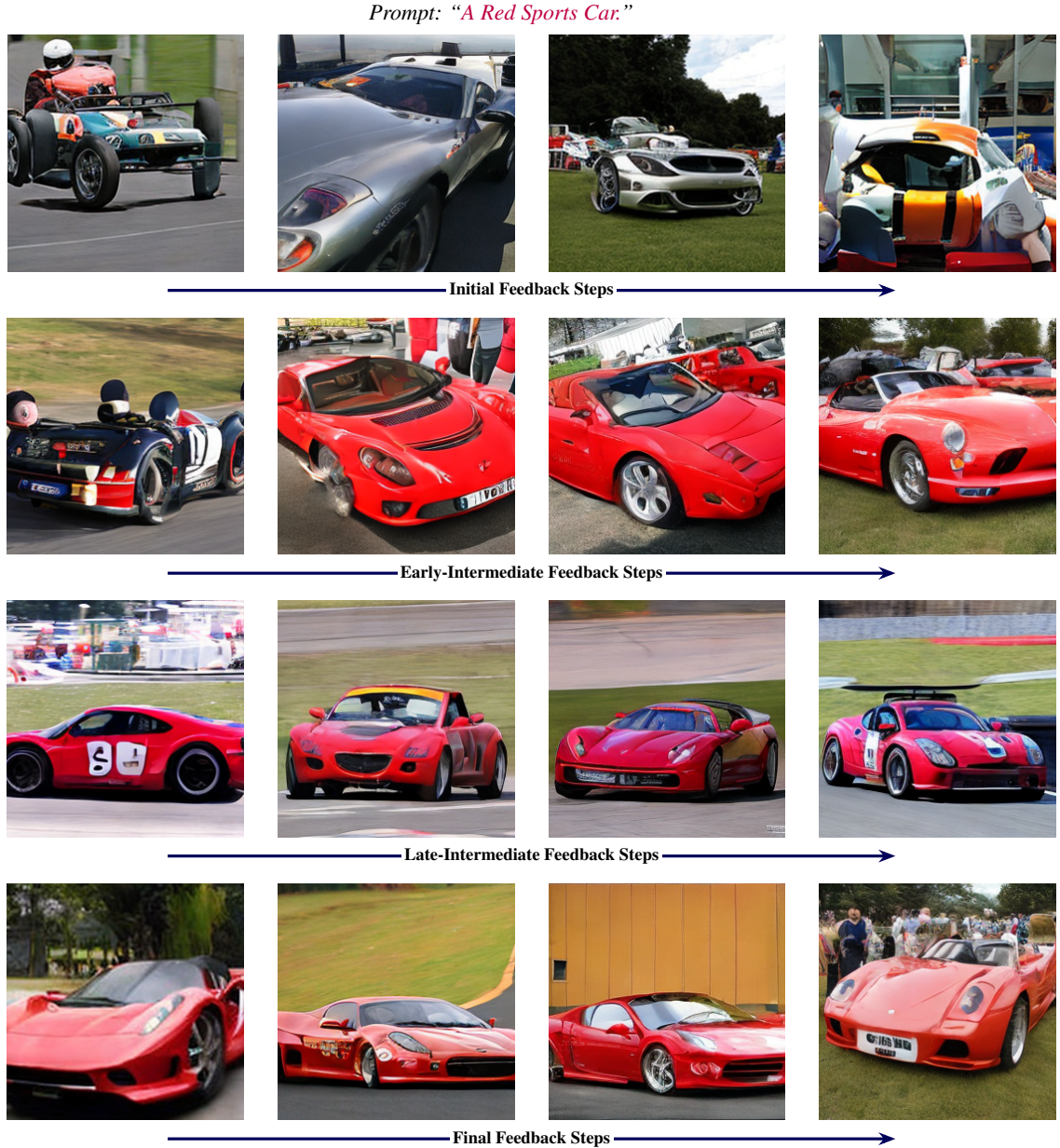
\begin{figure*}[h]
\centering
{\small\itshape Prompt: ``\textcolor{purple}{A Red Sports Car.}''\par}
\vspace{0.8ex}

\newcommand{\gridimage}[1]{%
  \IfFileExists{#1}{%
    \includegraphics[width=0.24\textwidth,height=0.24\textwidth,keepaspectratio]{#1}%
  }{%
    \fbox{\begin{minipage}[c][0.16\textwidth][c]{0.16\textwidth}\centering\scriptsize Add\\#1\end{minipage}}%
  }%
}
\newcommand{\feedbackarrow}[1]{%
  \begin{tikzpicture}[x=\textwidth,y=1cm]
    \draw[blue!35!black,line width=1.2pt,-{Stealth[length=2.4mm]}] (0.13,0) -- (0.87,0);
    \node[fill=white,inner sep=1.2pt,font=\scriptsize\bfseries] at (0.50,0) {#1};
  \end{tikzpicture}%
}
\setlength{\tabcolsep}{7pt}%
\renewcommand{\arraystretch}{1.0}%
\begin{tabular}{@{}cccc@{}}
  \gridimage{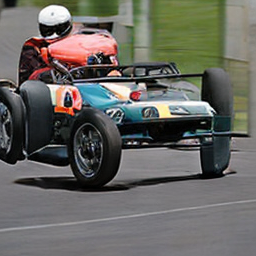} &
  \gridimage{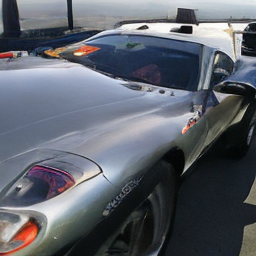} &
  \gridimage{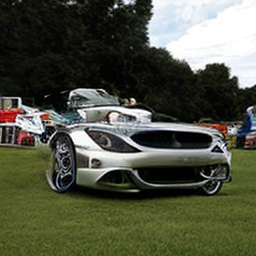} &
  \gridimage{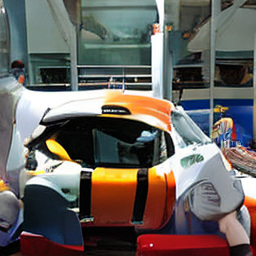} \\
  \multicolumn{4}{@{}c@{}}{\feedbackarrow{Initial Feedback Steps}} \\[0.8ex]
  \gridimage{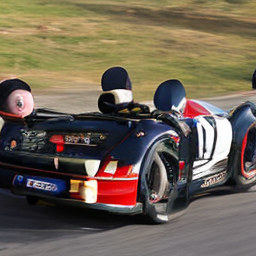} &
  \gridimage{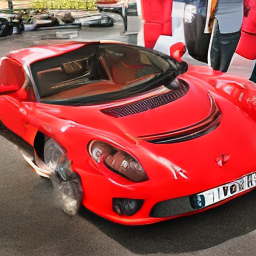} &
  \gridimage{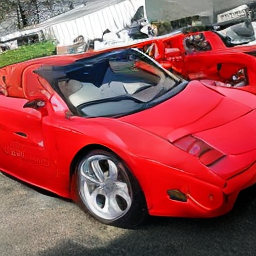} &
  \gridimage{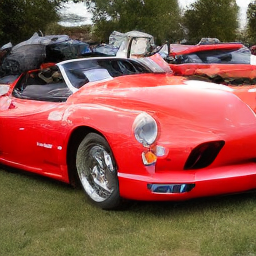} \\
  \multicolumn{4}{@{}c@{}}{\feedbackarrow{Early-Intermediate Feedback Steps}} \\[0.8ex]
  \gridimage{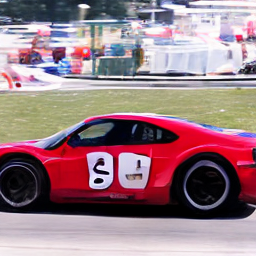} &
  \gridimage{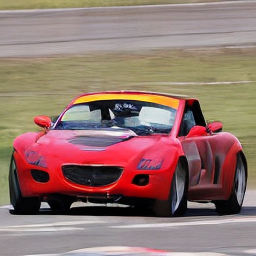} &
  \gridimage{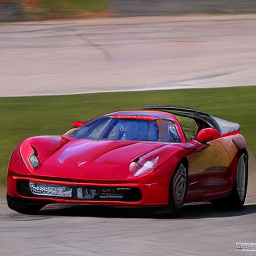} &
  \gridimage{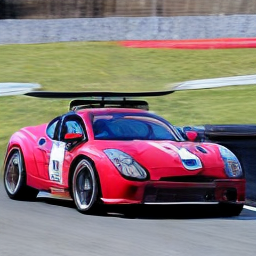} \\
  \multicolumn{4}{@{}c@{}}{\feedbackarrow{Late-Intermediate Feedback Steps}} \\[0.8ex]
  \gridimage{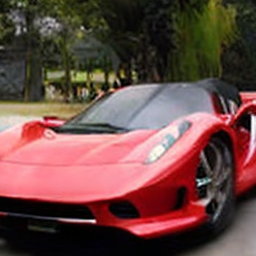} &
  \gridimage{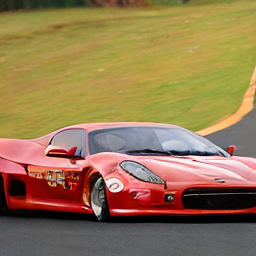} &
  \gridimage{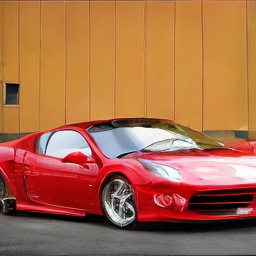} &
  \gridimage{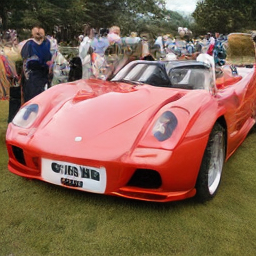} \\
  \multicolumn{4}{@{}c@{}}{\feedbackarrow{Final Feedback Steps}}
\end{tabular}
\vspace{-2pt}
\caption{\small{Exploration Strategy of BFMT on Online Feedback-Driven Search Settings.}}
\label{fig:spiral-search-trajectory1}
\end{figure*}

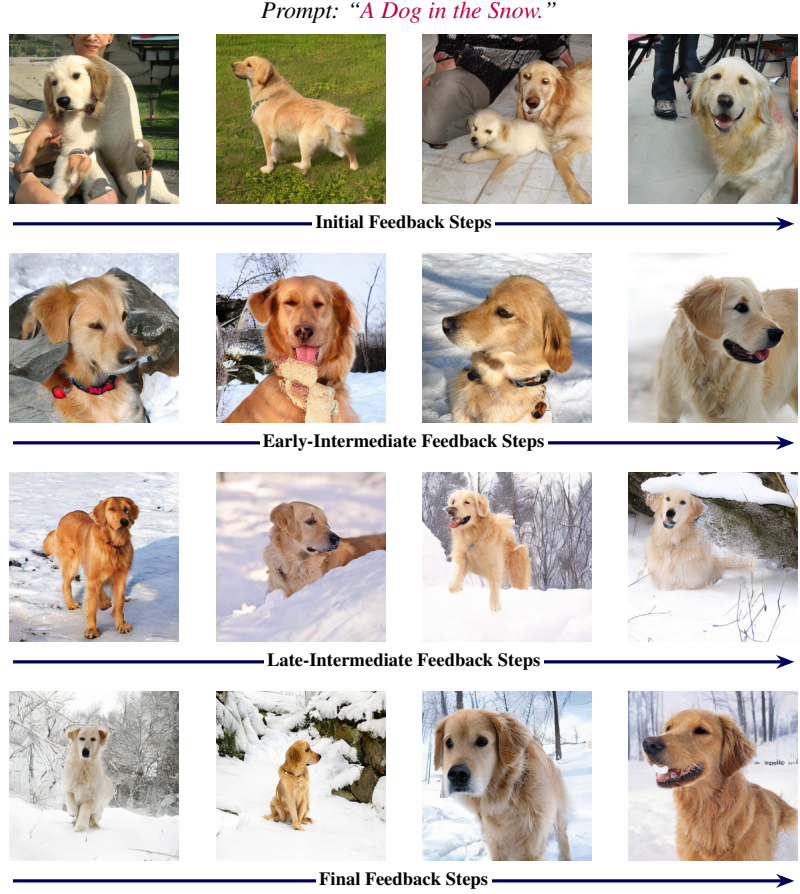
\begin{figure*}[h]
\centering
{\small\itshape Prompt: ``\textcolor{purple}{A Dog in the Snow.}''\par}
\vspace{0.8ex}

\newcommand{\gridimage}[1]{%
  \IfFileExists{#1}{%
    \includegraphics[width=0.16\textwidth,height=0.16\textwidth,keepaspectratio]{#1}%
  }{%
    \fbox{\begin{minipage}[c][0.16\textwidth][c]{0.16\textwidth}\centering\scriptsize Add\\#1\end{minipage}}%
  }%
}
\newcommand{\feedbackarrow}[1]{%
  \begin{tikzpicture}[x=\textwidth,y=1cm]
    \draw[blue!35!black,line width=1.2pt,-{Stealth[length=2.4mm]}] (0.13,0) -- (0.87,0);
    \node[fill=white,inner sep=1.2pt,font=\scriptsize\bfseries] at (0.50,0) {#1};
  \end{tikzpicture}%
}
\setlength{\tabcolsep}{7pt}%
\renewcommand{\arraystretch}{1.0}%
\begin{tabular}{@{}cccc@{}}
  \gridimage{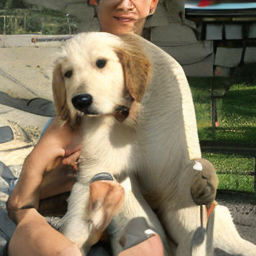} &
  \gridimage{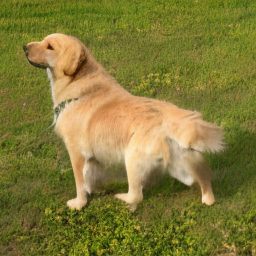} &
  \gridimage{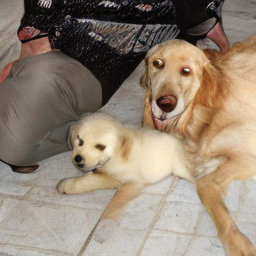} &
  \gridimage{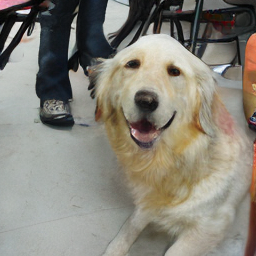} \\
  \multicolumn{4}{@{}c@{}}{\feedbackarrow{Initial Feedback Steps}} \\[0.8ex]
  \gridimage{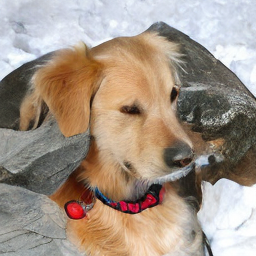} &
  \gridimage{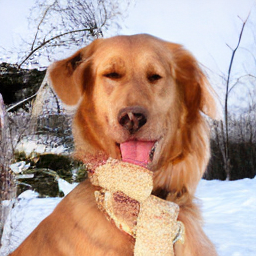} &
  \gridimage{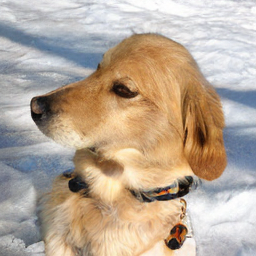} &
  \gridimage{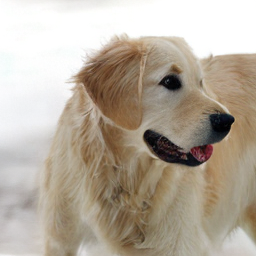} \\
  \multicolumn{4}{@{}c@{}}{\feedbackarrow{Early-Intermediate Feedback Steps}} \\[0.8ex]
  \gridimage{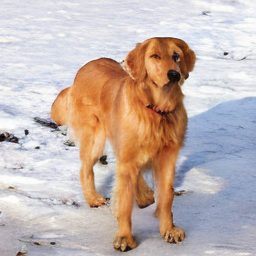} &
  \gridimage{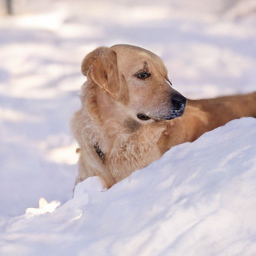} &
  \gridimage{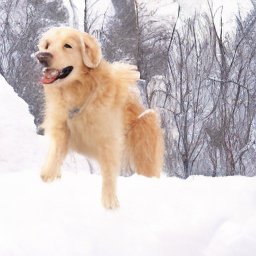} &
  \gridimage{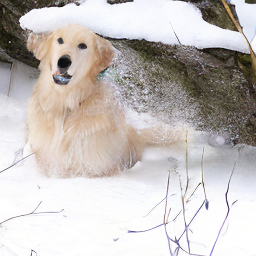} \\
  \multicolumn{4}{@{}c@{}}{\feedbackarrow{Late-Intermediate Feedback Steps}} \\[0.8ex]
  \gridimage{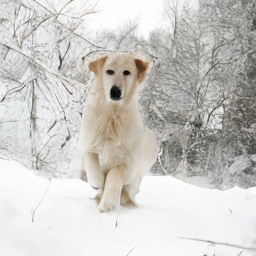} &
  \gridimage{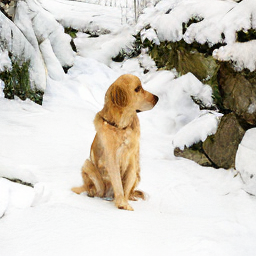} &
  \gridimage{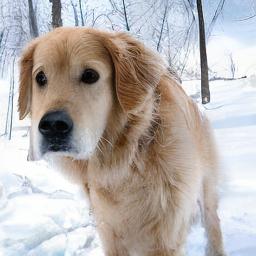} &
  \gridimage{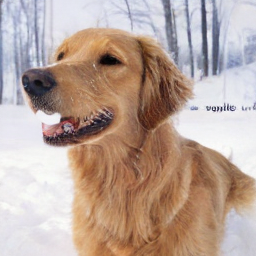} \\
  \multicolumn{4}{@{}c@{}}{\feedbackarrow{Final Feedback Steps}}
\end{tabular}
\vspace{-2pt}
\caption{\small{Exploration Strategy of BFMT on Online Feedback-Driven Search Settings.}}
\label{fig:spiral-search-trajectory2}
\end{figure*}

Figures~\ref{fig:spiral-search-trajectory1} and~\ref{fig:spiral-search-trajectory2} illustrate how BFMT's search strategy evolves across feedback stages. Initially, BFMT explores broadly, generating semantically diverse samples within the target category, though not yet precisely aligned with the fine-grained target prompt. As feedback accumulates, it progressively converges toward well-aligned samples, showcasing an effective exploration-to-exploitation dynamic that is central to its success in online feedback-driven settings.

\vspace{10pt}

\section{Hierarchical Exploration Strategy of BFMT}\label{sec:hie-vis-b}
Here, we illustrate BFMT's hierarchical search strategy and its enhanced exploration capability, enabled by the dynamic transition schedule. 
\begin{figure}[h]
    \centering
    \includegraphics[width=0.75\linewidth]{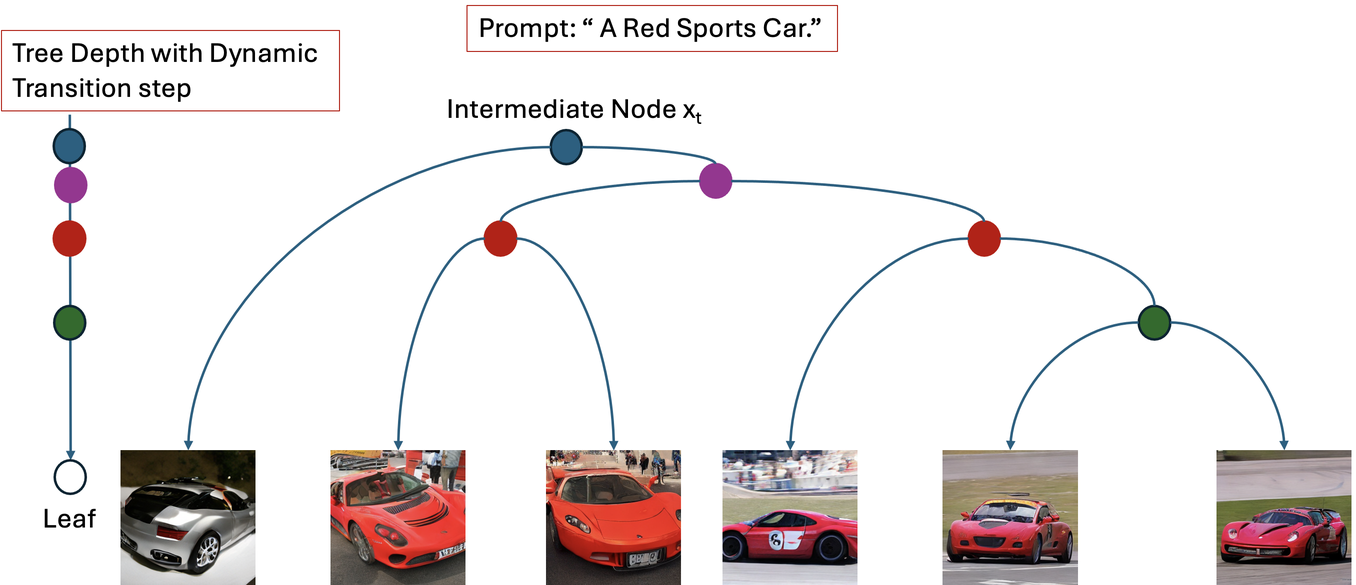}
    \caption{\small{BFMT's Hierarchical Search Strategy.}}
    \label{fig:bfmt_tree_search}
\end{figure}
As shown in the figure~\ref{fig:bfmt_tree_search}, smaller transition steps in the early stages allow BFMT to explore the space more broadly (e.g., BFMT samples a car with a different color and a different background), while larger transition steps in the later stages enable it to focus on and sample from high-utility modes (e.g., BFMT samples a red sports car). 

\section{Additional Visualizations of Hierarchical Exploration Strategy of BFMT}\label{sec:hie-vis-a}
\begin{figure}[h]
    \centering
    \includegraphics[width=\linewidth]{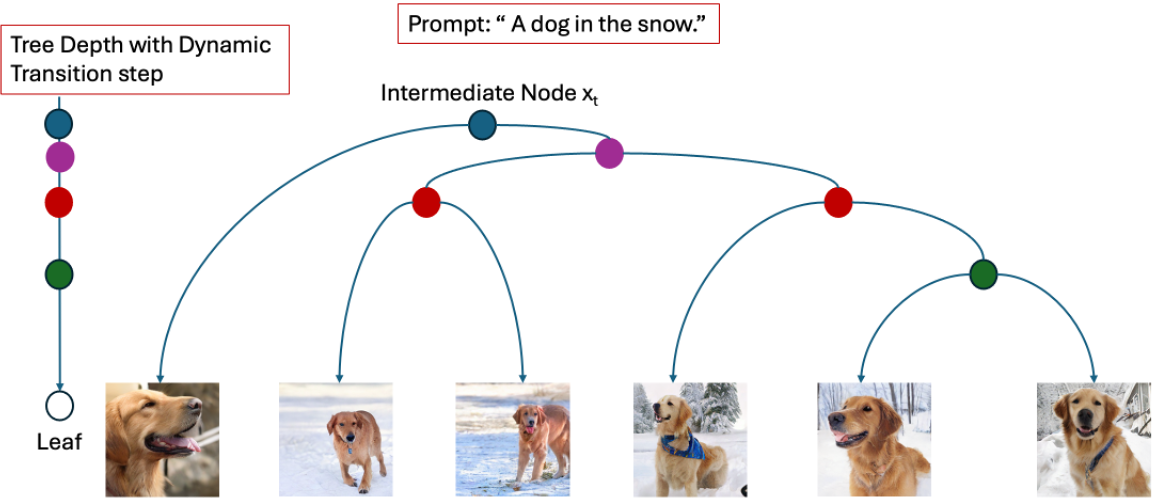}
    \caption{\small{BFMT's Hierarchical Search Strategy.}}
    \label{fig:bfmt_tree_search1}
\end{figure}

\begin{figure}[h]
    \centering
    \includegraphics[width=\linewidth]{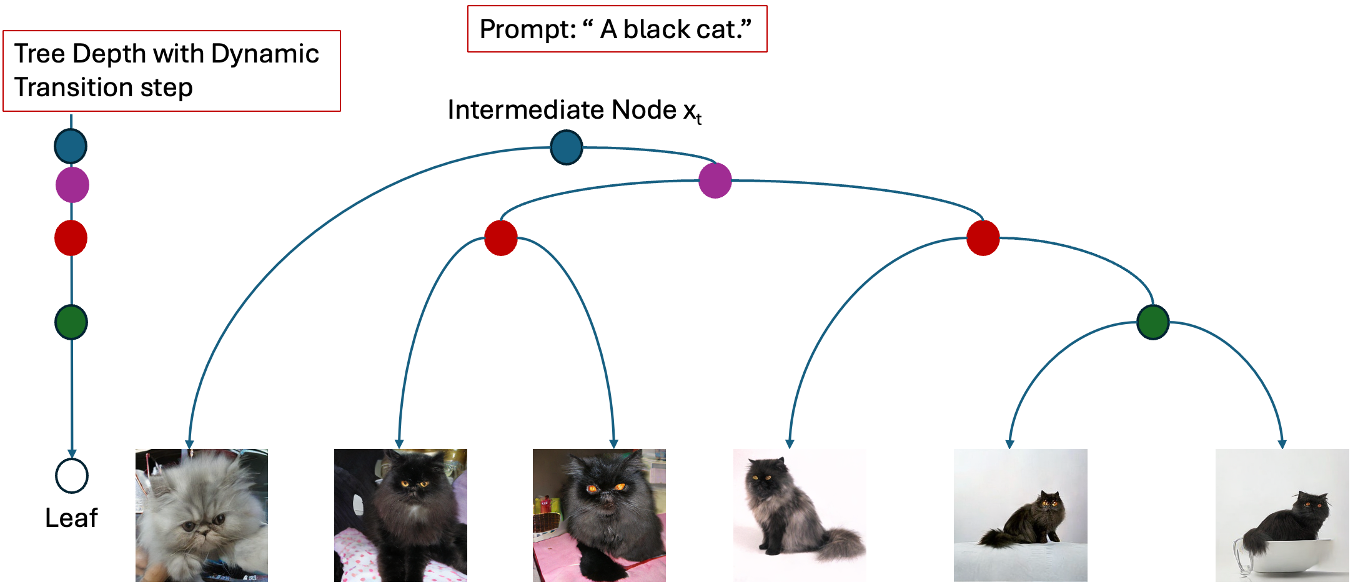}
    \caption{\small{BFMT's Hierarchical Search Strategy.}}
    \label{fig:bfmt_tree_search2}
\end{figure}

Figures~\ref{fig:bfmt_tree_search1} and~\ref{fig:bfmt_tree_search2} present additional visualizations of BFMT's hierarchical exploration strategy, further reinforcing its effectiveness in enabling efficient online feedback-driven search.

\vspace{10pt}

\section{Adapting the TrigFlow Noise Schedule to the BFMT Framework}\label{sec:mean-a}
\label{sec:trigflow_adaptation}
 
In this section, we describe how we adapt the trigonometric noise schedule 
from TrigFlow~\cite{chen2025sana} and the consistency model (SANA-Sprint)~\cite{chen2025sana} to our BFMT framework, which operates under a different noise schedule. We 
first review the TrigFlow parameterization, then describe the consistency model 
reparameterization, and finally derive the adaptation of the noise schedule 
to our setting.
 
\subsection{Background: TrigFlow Transformation}
\label{subsec:trigflow_background}
 
TrigFlow introduces a trigonometric reparameterization of the Flow Matching 
timestep, establishing a bijective mapping between the standard FM timestep 
$t_{FM} \in [0, 1]$ and a cosine-based schedule parameterized by $t_{cm}$. 
Specifically, the mapping is defined as:
\begin{equation}
    t_{FM} = \frac{\sin(t_{cm})}{\sin(t_{cm}) + \cos(t_{cm})},
    \label{eq:tFM}
\end{equation}
where $t_{cm}$ controls the balance between noise and signal through 
trigonometric interpolation. This mapping ensures that $t_{FM} = 0$ 
corresponds to a clean sample and $t_{FM} = 1$ corresponds to pure noise, 
while providing a smoother, geometry-aware interpolation path compared to 
a linear schedule.
 
Given the mapping in Eq.~\eqref{eq:tFM}, the noisy latent $X_{t_{FM}}$ is 
expressed in terms of the original latent $X_{t_{cm}}$ as:
\begin{equation}
    X_{t_{FM}} = \frac{X_{t_{cm}}}{\sigma_d} \sqrt{t_{FM}^2 + (1 - t_{FM})^2},
    \label{eq:Xtfm}
\end{equation}
where $\sigma_d$ is a scaling factor that normalizes the latent representation. 
The factor $\sqrt{t_{FM}^2 + (1-t_{FM})^2}$ accounts for the norm of the 
trigonometric interpolation coefficients, ensuring consistent scaling 
across timesteps.
 
Under this reparameterization, the score network $\hat{F}_\theta$ is 
reformulated to operate on the rescaled latent $X_{t_{cm}} / \sigma_d$ at 
timestep $t_{cm}$. The relationship between the TrigFlow score network 
and the velocity field $V_\theta$ of the standard Flow Matching model is:
\begin{equation}
    \begin{split}
        \hat{F}_\theta\!\left(\frac{X_{t_{cm}}}{\sigma_d},\, t_{cm},\, y\right) 
        &= \frac{1}{\sqrt{t_{FM}^2 + (1 - t_{FM})^2}} \\
        &\quad \times \Big[ (1 - 2t_{FM})\, X_{t,FM} + \left(1 - 2t_{FM} + 2t_{FM}^2\right) V_\theta\!\left(X_{t_{FM}},\, t_{FM},\, y\right) \Big],
    \end{split}
    \label{eq:Ftheta}
\end{equation}
where $y$ is the conditioning input (e.g., a text prompt). This 
reformulation shows that the TrigFlow score network can be expressed 
as a linear combination of the noisy latent and the predicted velocity 
field, weighted by time-dependent coefficients that are derived analytically 
from the trigonometric schedule.
 
\subsection{SANA-Sprint Reparameterization}
\label{subsec:sana_sprint}
 
SANA-Sprint builds upon TrigFlow by introducing a consistency-style 
denoising function $f_\theta$ that directly maps a noisy latent $X_t$ 
to a clean estimate $\hat{X}_0$ in a single step. Under the TrigFlow 
time parameterization, this function takes the form:
\begin{equation}
    f_\theta(X_t, t) = \cos(t_{cm}) \cdot X_t 
    - \sin(t_{cm}) \cdot \sigma_d \cdot F_\theta\!\left(\frac{X_t}{\sigma_d},\, t_{cm}\right),
    \label{eq:ftheta_sana}
\end{equation}
where $X_{cm} = X_t / \text{Scale}$ is the rescaled latent. Intuitively, 
Eq.~\eqref{eq:ftheta_sana} can be understood as a trigonometric 
decomposition: the $\cos(t_{cm})$ term retains the signal component of 
$X_t$, while the $\sin(t_{cm})$ term removes the noise component as 
predicted by $F_\theta$. This yields the denoised estimate:
\begin{equation}
    \hat{X}_0 = \cos(t_{cm})\, X_{t_{cm}} - \sin(t_{cm})\, \hat{F}_\theta,
    \label{eq:X0hat}
\end{equation}
which provides a clean, geometry-consistent reconstruction of the 
original sample from any noisy intermediate state.
 
\subsection{Adapting to Our BFMT Framework with the Sufficient Statistic ($S$) Schedule}
\label{subsec:schedule_adaptation}
 
Our BFMT framework uses a different noise schedule compared to 
TrigFlow. Specifically, we adopt the $S$ schedule, which defines a 
signal-to-noise ratio (SNR) function that differs from the trigonometric 
parameterization used in TrigFlow. To adapt the SANA-Sprint framework 
to our setting, we must establish a consistent mapping between the two 
schedules.
 
\paragraph{Setting up the Schedule Correspondence.}
Under the $S$ schedule, the signal and noise coefficients at timestep 
$t = t_{cm}$ are defined as:
\begin{equation}
    \alpha_t = \cos t = \sqrt{s}, 
    \qquad 
    \sigma_t = \sin t = \sqrt{1 - s},
    \label{eq:alpha_sigma}
\end{equation}
where $s \in [0, 1]$ is the schedule variable. This gives the forward 
noising process:
\begin{equation}
    X_t = \left(\alpha_t\, \hat{X}_0 + \sigma_t\, \frac{\varepsilon}{\sigma_d}\right) \cdot \sigma_d
        = \cos(t)\, \hat{X}_0 \cdot \sigma_d + \sin(t)\, \varepsilon,
    \label{eq:forward_process}
\end{equation}
which is a standard affine interpolation between the clean sample 
$\hat{X}_0$ and Gaussian noise $\varepsilon \sim \mathcal{N}(0, I)$, 
scaled by the latent normalization factor $\sigma_d$.
 
\paragraph{Defining the SNR Function.}
Under the $S$ schedule, the SNR function $g(s)$ and its transformed 
counterpart $\tilde{g}(\alpha_t)$ are defined as:
\begin{equation}
    g(s) = \frac{1}{s} - 1,
    \qquad
    \tilde{g}(\alpha_t) = \frac{1}{\alpha_t} - 1 
    = \frac{1}{\cos^2(t_{cm})} - 1,
    \label{eq:snr_functions}
\end{equation}
where $\tilde{g}(\alpha_t)$ is the SNR expressed as a function of the 
signal coefficient $\alpha_t = \cos(t_{cm})$. This function is monotonically 
decreasing in $t_{cm}$, consistent with the intuition that noise increases 
as the timestep progresses.
 
\paragraph{Deriving the Optimal Transition Timestep $r^* = r^*_{cm}$ for computing DDPM transition via Sufficient Statistic.}
A key step in adapting the noise schedule is determining the optimal 
intermediate timestep $r^*$ at which to perform the consistency transition 
between two noise levels $t_{cm}$ and $t_{cm'}$ (with $t_{cm'} < t_{cm}$, 
i.e., $t_{cm'}$ is closer to the clean data). The optimal $r^*$ is derived 
by matching the SNR levels under the $S$ schedule, yielding:
\begin{equation}
    \tilde{g}(r^*) = \frac{\tilde{g}(\cos t_{cm'}) \cdot \tilde{g}(\cos t_{cm})}
    {\tilde{g}(\cos t_{cm}) - \tilde{g}(\cos t_{cm'})},
    \label{eq:r_star_snr}
\end{equation}
which can be solved by applying the inverse of $g$ to both sides:
\begin{equation}
    \hat{r}^* = g^{-1}\!\left(
    \frac{\tilde{g}(\cos t_{cm'}) \cdot \tilde{g}(\cos t_{cm})}
    {\tilde{g}(\cos t_{cm}) - \tilde{g}(\cos t_{cm'})}
    \right).
    \label{eq:r_star}
\end{equation}
Eq.~\eqref{eq:r_star} gives the optimal SNR level $r^*$ in the $s$-domain. 
To recover the corresponding trigonometric timestep $r^*_{cm}$, we use 
the relation $\alpha_t = \cos(t_{cm}) = \sqrt{s}$, which gives:
\begin{equation}
    \cos(r^*_{cm}) = d\sqrt{r^*} 
    \quad \Longrightarrow \quad 
    r^*_{cm} = \arccos\!\left(\sqrt{r^*}\right).
    \label{eq:r_star_cm}
\end{equation}
This recovers the optimal intermediate timestep $r^*_{cm}$ in the 
trigonometric domain, which can be directly used within the 
TrigFlow/SANA-Sprint framework.
 
\paragraph{Computing the DDPM Transition via Sufficient Statistic}
Given the optimal transition timestep $r^*_{cm}$, the DDPM transition sample 
$X_{t_{cm'}}$ at the new timestep $t_{cm'}$ is obtained by applying 
the Sufficient Statistic scheme as defined in Equation 9 in the main paper. 
%
%

\section{Details of Class and Prompt Detail in the Search Experiment}\label{sec:prompt-a}
In the following table~\ref{tb:st}, we detail the randomly selected ImageNet target classes and their fine-grained specifications used to evaluate the competing methods on the online feedback-driven search task.
\begin{table}[h]
    \centering
    \begin{tabular}{|p{0.22\textwidth}|p{0.68\textwidth}|}
        \hline
        \textbf{Target Class} & \textbf{Fine-Grained Prompt} \\
        \hline
        Car & A Red Sports car. \\
        \hline
        Cat & A Black Cat. \\
        \hline
        Dog & A Dog in the Snow. \\
        \hline
        Pepper & A Green Bell Pepper. \\
        \hline
        Daisy & A Red Daisy. \\
        \hline
    \end{tabular}
    \vspace{2pt}
    \caption{Target classes and their corresponding fine-grained prompts for the Online Feedback-driven Search task.}
    \label{tb:st}
\end{table}

\section{Details about the Evaluation Dataset for Alignment Tasks}\label{sec:align-eval-data}
We provide the evaluation prompts for both the compositional and quantity-aware alignment tasks in separate Excel files, available via the zip file. 
\section{Additional Results with Molecule Generation Task}\label{sec:mole-add}
Tables~\ref{tab:mole_K} and~\ref{tab:mole_S} report the mean validity, uniqueness, QED, and composite QED and SA scores for samples generated under feedback budgets 200 and 600, respectively. These results further highlight BFMT's ability to efficiently generate valid, target-aligned molecules even under tightly constrained feedback budgets.
\begin{table}[!h]
\centering
\caption{Validity Check for the Generated Molecules.}
\label{tab:mole_K}
\setlength{\tabcolsep}{5pt}
\renewcommand{\arraystretch}{0.90}
\begin{tabular}{lccc|ccc}
\toprule
Method & Uniqueness & Validity & Mean QED & Uniqueness & Validity & Mean QED+SA \\
\midrule
CFM & 0.9993 & 0.9410 & 0.4876 & 0.9993 & 0.9410 & 0.5148\\
\rowcolor{blue!20}BFMT & 0.9823 & 0.9620 & \textbf{0.5489} & 0.9778 & 0.9500 & \textbf{0.5627}\\
\bottomrule
\end{tabular}
\end{table}
\begin{table}[!h]
\centering
\caption{Validity Check for the Generated Molecules.}
\label{tab:mole_S}
\setlength{\tabcolsep}{5pt}
\renewcommand{\arraystretch}{0.90}
\begin{tabular}{lccc|ccc}
\toprule
Method & Uniqueness & Validity & Mean QED & Uniqueness & Validity & Mean QED+SA \\
\midrule
CFM & 0.9989 & 0.9410 & 0.4925 & 0.9989 & 0.9410 & 0.5202\\
\rowcolor{blue!20}BFMT & 0.9774 & 0.9579 & \textbf{0.5575} & 0.9766 & 0.9588 & \textbf{0.5702}\\
\bottomrule
\end{tabular}
\end{table}
\section{Implementation Details}\label{sec:imp-a}
Implementation details are available in the provided GitHub link. Finally, all experiments are implemented in PyTorch and conducted
on 3 NVIDIA A100 GPUs. Our training and inference code will be made public.
\section{Additional Comparative Search Visualizations}\label{sec:search-vis-a}
\begin{figure*}[h]
\centering
\newcommand{\triplepanel}[3]{%
  \setlength{\tabcolsep}{2pt}%
  \begin{tabular}{@{}c@{\hspace{4pt}}c@{\hspace{4pt}}c@{}}
    \includegraphics[width=0.315\linewidth]{#1} &
    \includegraphics[width=0.315\linewidth]{#2} &
    \includegraphics[width=0.315\linewidth]{#3}
  \end{tabular}%
}
\newcommand{\panelblock}[4]{%
  \begin{minipage}[t]{\linewidth}
    \centering
    {\itshape ``#1''\par}
    \vspace{0pt}
    \triplepanel{#2}{#3}{#4}
  \end{minipage}%
}
\newcommand{\topheaders}{%
  {\fontsize{9.5}{10.5}\selectfont
  \begin{tabular}{@{}c@{\hspace{4pt}}c@{\hspace{4pt}}c@{}}
    \makebox[0.315\linewidth][c]{\textbf{DTS}} &
    \makebox[0.315\linewidth][c]{\textbf{MFM}} &
    \makebox[0.315\linewidth][c]{\textbf{BFMT}}
  \end{tabular}%
  }
}
\noindent
\topheaders
\vspace{-0.4ex}
\noindent\rule{\textwidth}{0.6pt}
\vspace{-0.8ex}
\begin{minipage}[t]{\textwidth}
\centering
\panelblock{a black cat.}{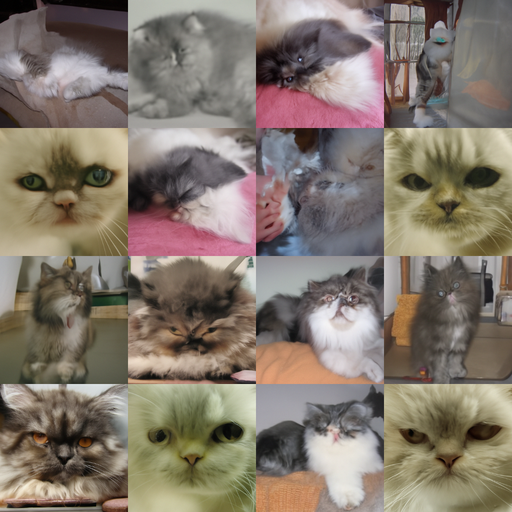}{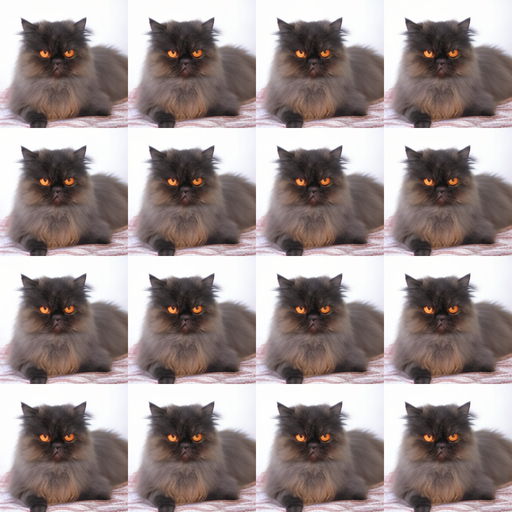}{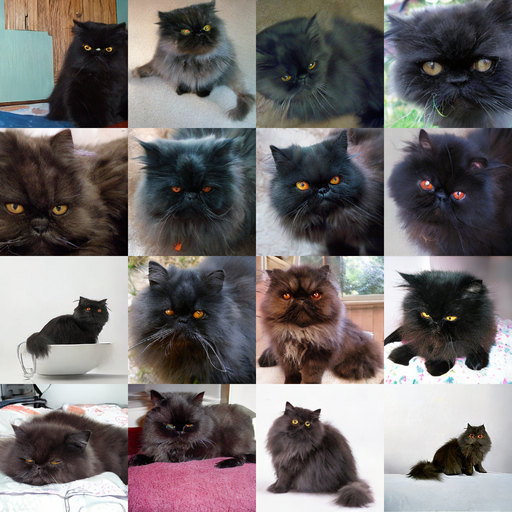}
\end{minipage}
\vspace{0.8ex}
\noindent\rule{\textwidth}{0.6pt}
\vspace{-0.4ex}
\topheaders
\vspace{-0.4ex}
\noindent\rule{\textwidth}{0.6pt}
\vspace{-0.8ex}
\begin{minipage}[t]{\textwidth}
\centering
\panelblock{A red daisy.}{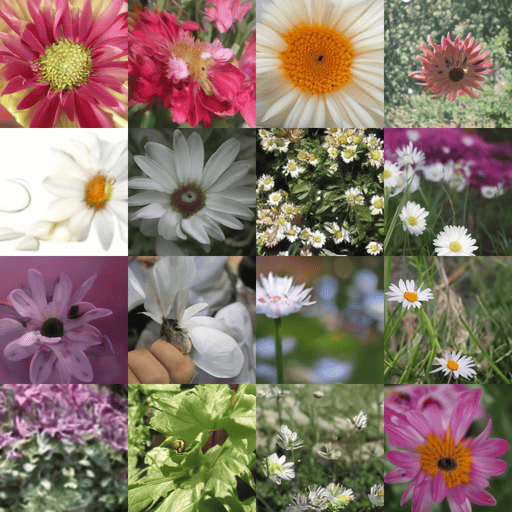}{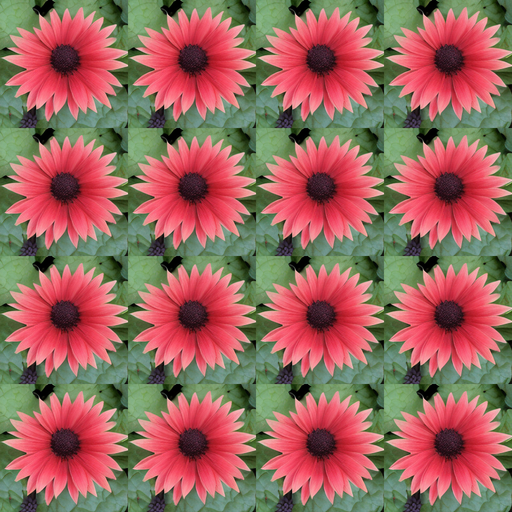}{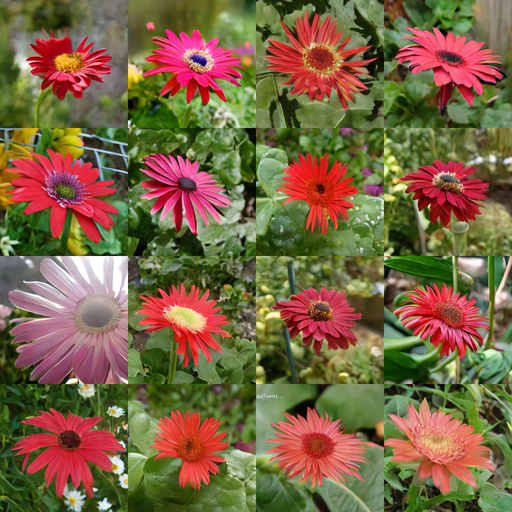}
\end{minipage}%
\vspace{0.6ex}
\noindent\rule{\textwidth}{0.6pt}
\vspace{-0.4ex}
\topheaders
\vspace{-0.4ex}
\noindent\rule{\textwidth}{0.6pt}
\vspace{-0.8ex}
\begin{minipage}[t]{\textwidth}
\centering
\panelblock{A green pepper.}{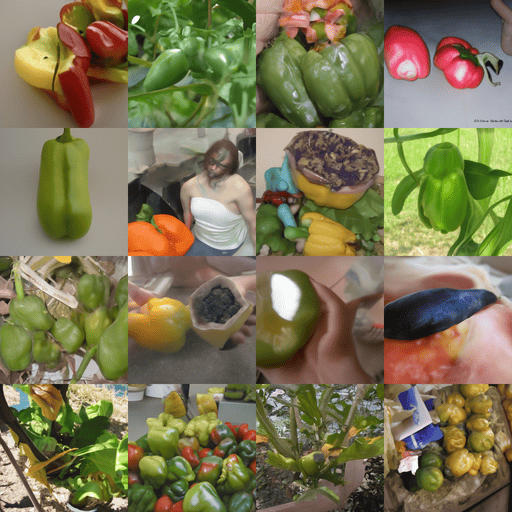}{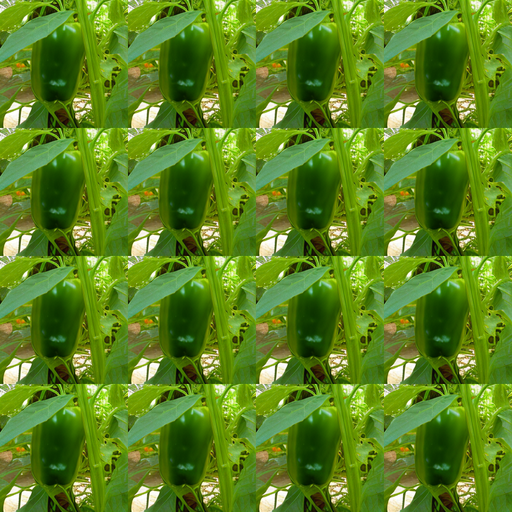}{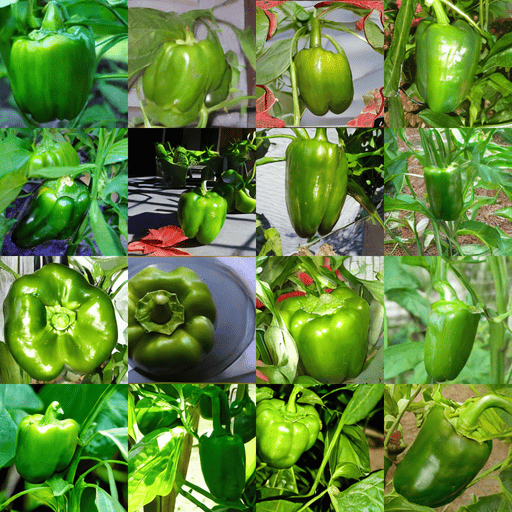}
\end{minipage}%
\vspace{0.6ex}
\noindent\rule{\textwidth}{0.6pt}
\caption{\small{Additional Search Visualizations.}}
\label{fig:search-visu-add}
\vspace{-18pt}
\end{figure*}

We provide additional qualitative results for the online feedback-driven search task in Figure~\ref{fig:search-visu-add}, comparing BFMT against the most competitive baselines. These visualizations further corroborate the ability of BFMT to produce diverse, target-aligned samples under online feedback.
\pagebreak

\section{Additional Competitive Alignment Visualization}\label{sec:align-vis-a}
\begin{figure*}[h]
\centering
\newcommand{\triplepanel}[3]{%
  \setlength{\tabcolsep}{2pt}%
  \begin{tabular}{@{}c@{\hspace{4pt}}c@{\hspace{4pt}}c@{}}
    \includegraphics[width=0.315\linewidth]{#1} &
    \includegraphics[width=0.315\linewidth]{#2} &
    \includegraphics[width=0.315\linewidth]{#3}
  \end{tabular}%
}
\newcommand{\panelblock}[4]{%
  \begin{minipage}[t]{\linewidth}
    \centering
    {\itshape ``#1''\par}
    \vspace{0pt}
    \triplepanel{#2}{#3}{#4}
  \end{minipage}%
}
\newcommand{\topheaders}{%
  {\fontsize{9.5}{10.5}\selectfont
  \begin{tabular}{@{}c@{\hspace{4pt}}c@{\hspace{4pt}}c@{}}
    \makebox[0.315\linewidth][c]{\textbf{DTS}} &
    \makebox[0.315\linewidth][c]{\textbf{FKS}} &
    \makebox[0.315\linewidth][c]{\textbf{BFMT}}
  \end{tabular}%
  }
}
\resizebox{0.9\textwidth}{!}{%
\begin{minipage}{\textwidth}
\noindent
\topheaders
\vspace{-0.4ex}
\noindent\rule{\textwidth}{0.6pt}
\vspace{-0.8ex}
\begin{minipage}[t]{\textwidth}
\centering
\panelblock{Four Trucks.}{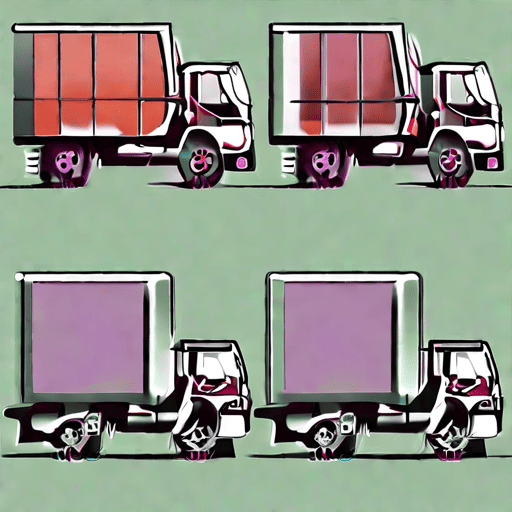}{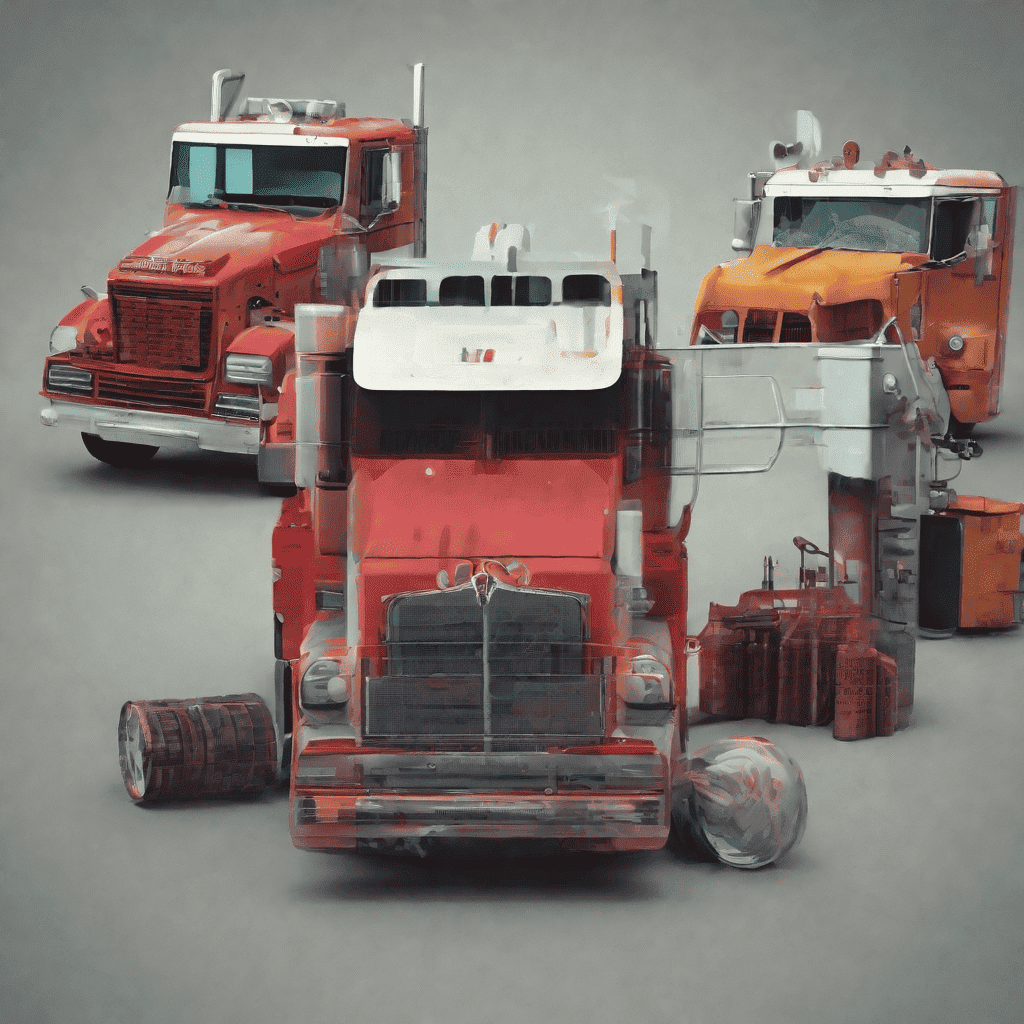}{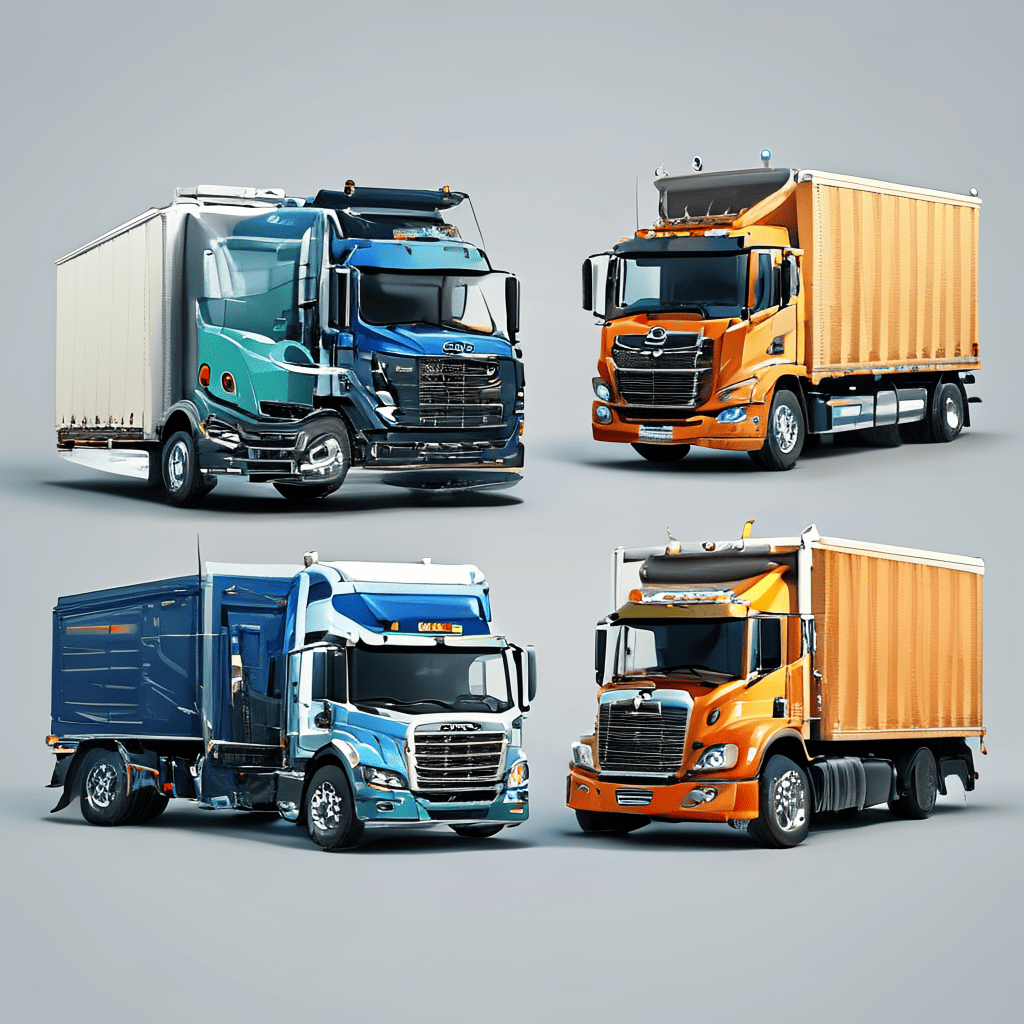}
\end{minipage}
\vspace{0.6ex}
\noindent\rule{\textwidth}{0.6pt}
\vspace{-0.4ex}
\topheaders
\vspace{-0.4ex}
\noindent\rule{\textwidth}{0.6pt}
\vspace{-0.8ex}
\begin{minipage}[t]{\textwidth}
\centering
\panelblock{Six bowls.}{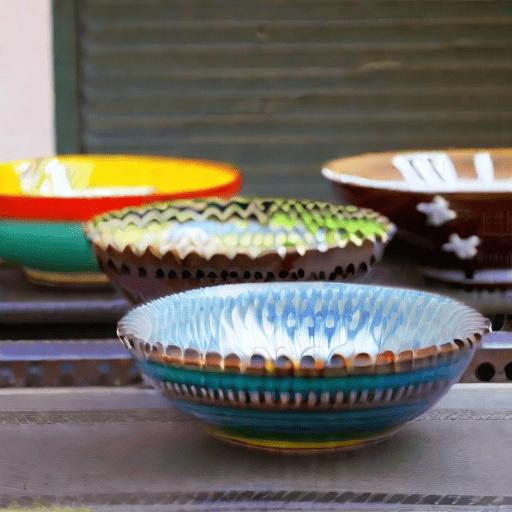}{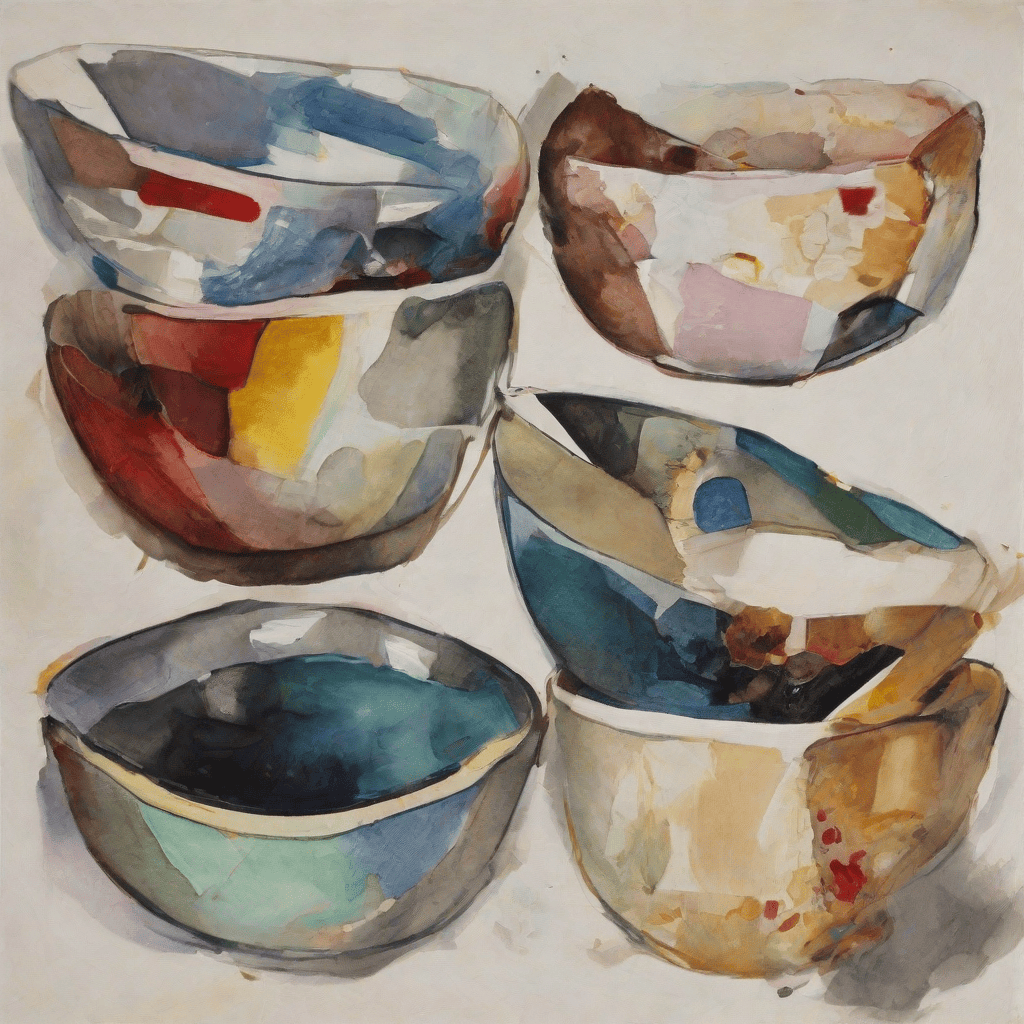}{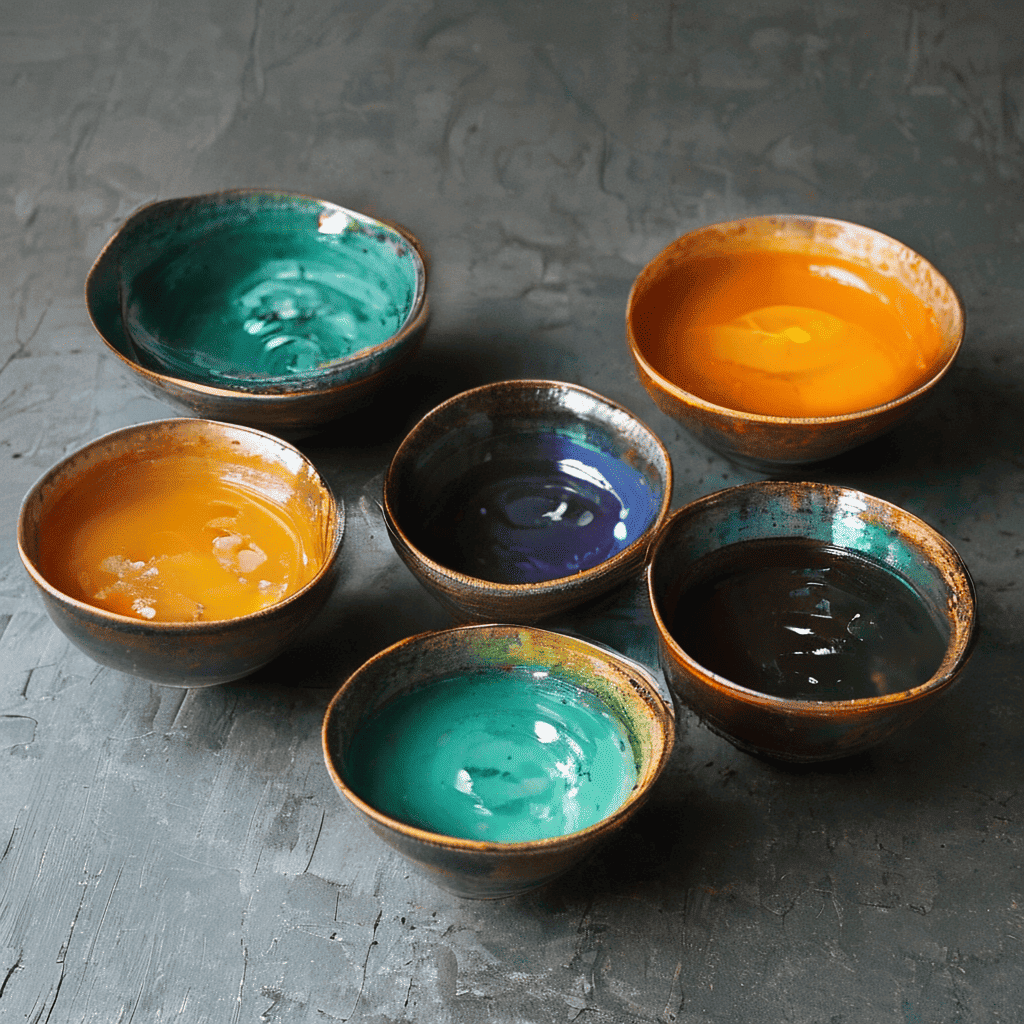}
\end{minipage}
\vspace{0.6ex}
\noindent\rule{\textwidth}{0.6pt}
\vspace{-0.4ex}
\topheaders
\vspace{-0.4ex}
\noindent\rule{\textwidth}{0.6pt}
\vspace{-0.8ex}
\begin{minipage}[t]{\textwidth}
\centering
\panelblock{A lantern casting dim light in a haunted forest.}{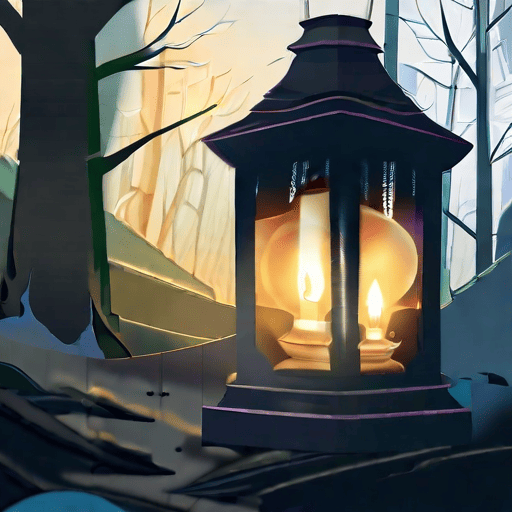}{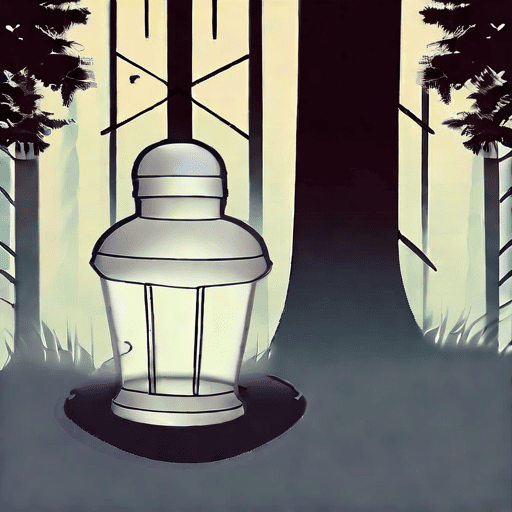}{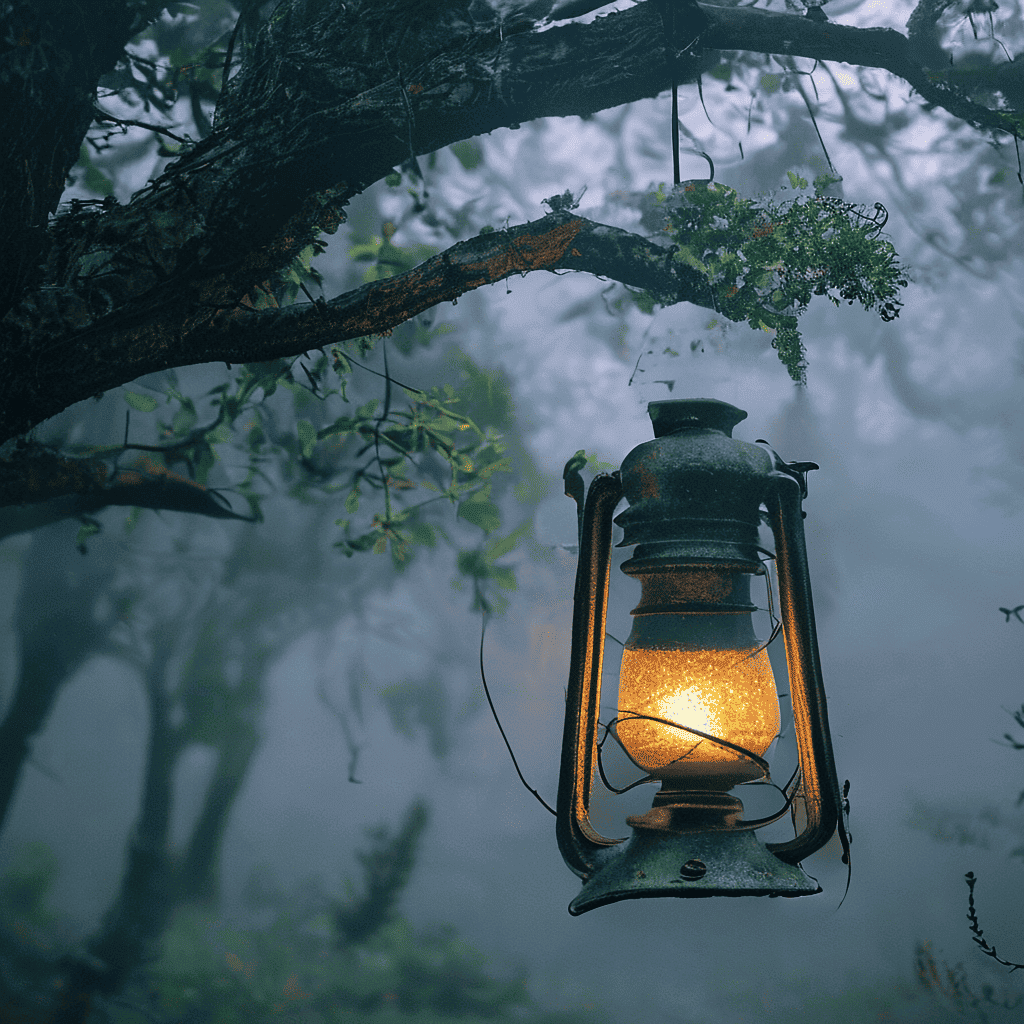}
\end{minipage}
\vspace{0.6ex}
\noindent\rule{\textwidth}{0.6pt}
\vspace{-0.4ex}
\topheaders
\vspace{-0.4ex}
\noindent\rule{\textwidth}{0.6pt}
\vspace{-0.8ex}
\begin{minipage}[t]{\textwidth}
\centering
\panelblock{A sorcerer's hat casting shadows over a cluttered, enchanted desk.}{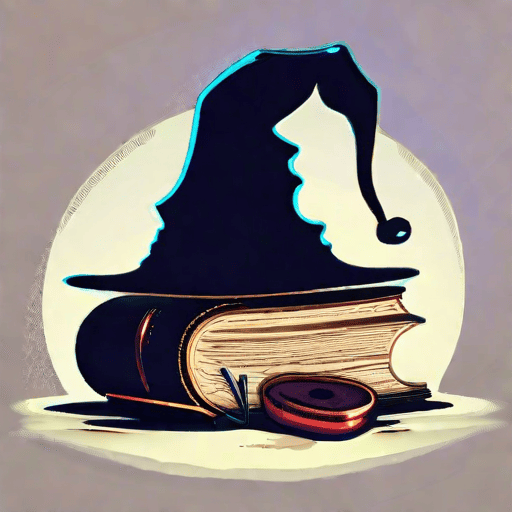}{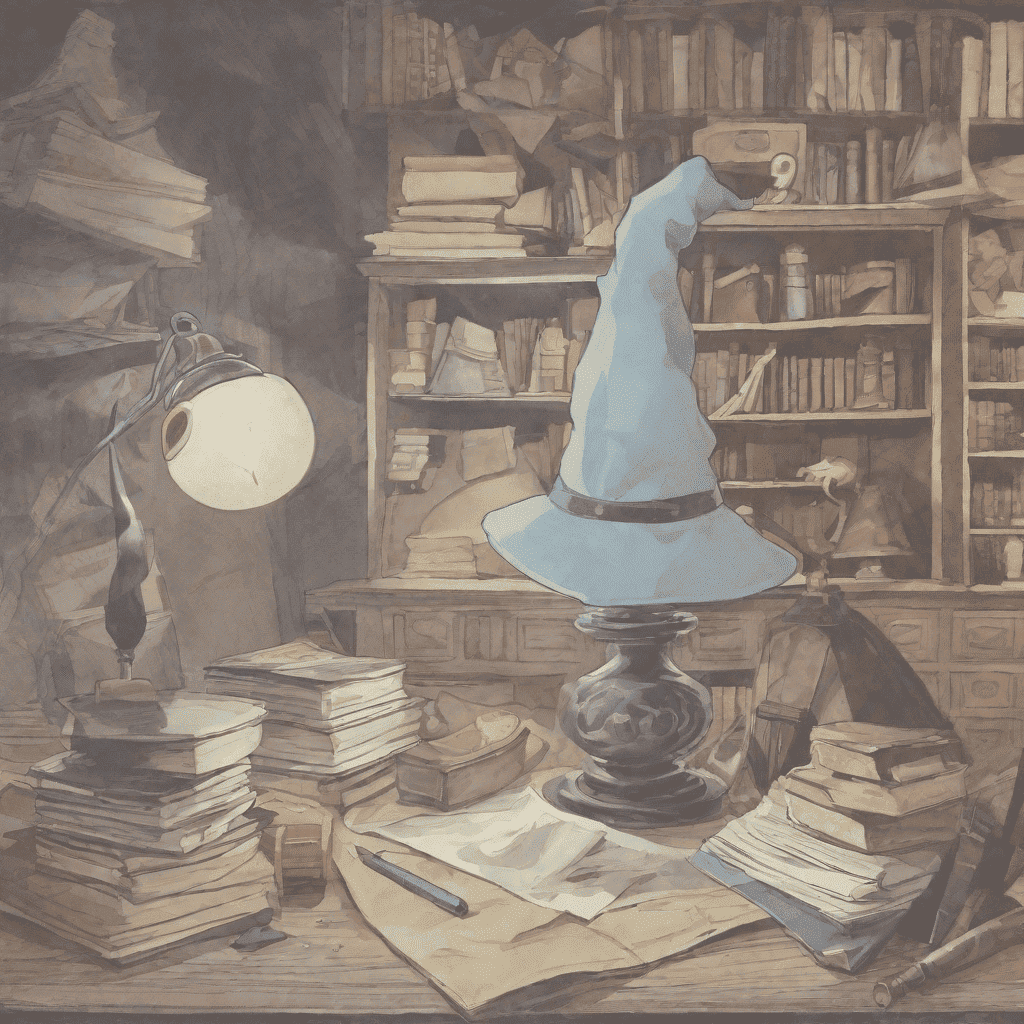}{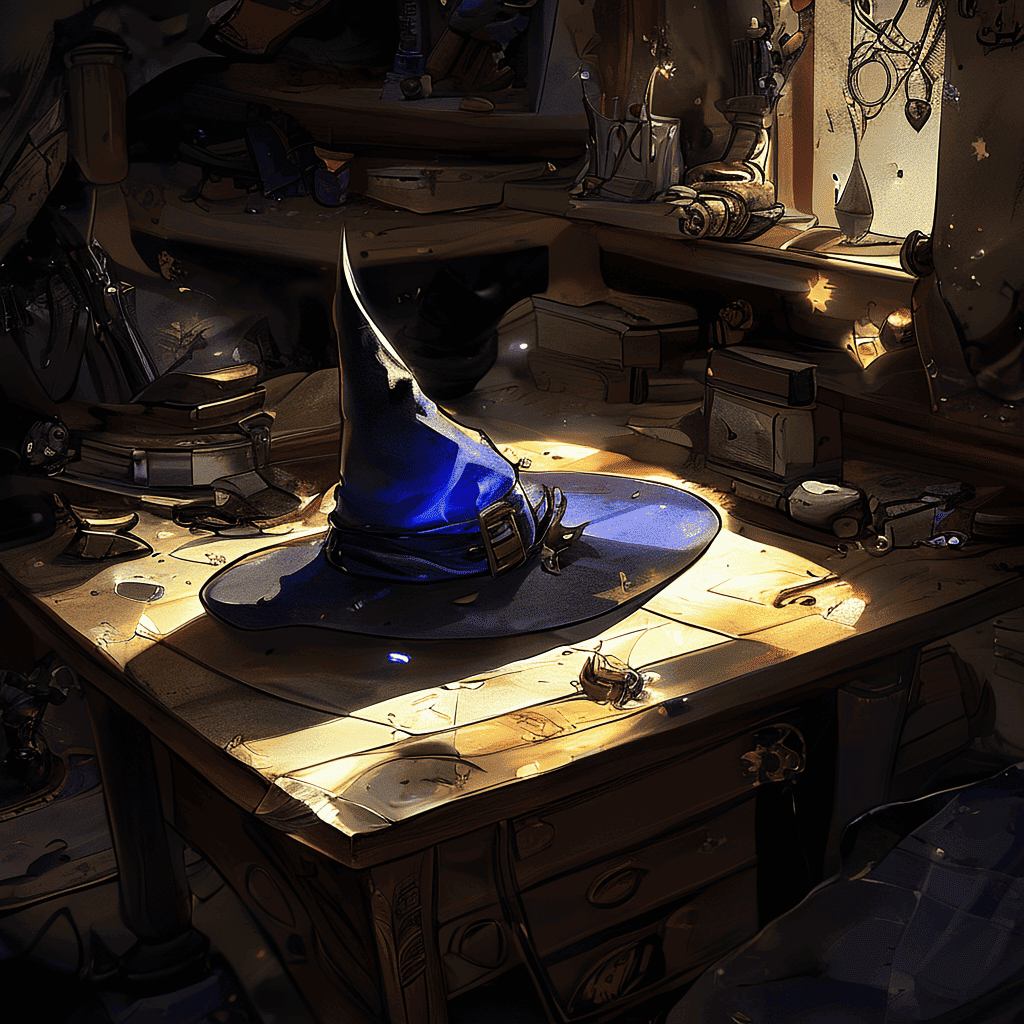}
\end{minipage}
\vspace{0.6ex}
\noindent\rule{\textwidth}{0.6pt}
\end{minipage}%
}
\caption{\small{Additional Alignment Visualization.}}
\label{fig:align-visu-add1}
\vspace{-9pt}
\end{figure*}

\begin{figure*}[h]
\centering
\newcommand{\methodrowpanel}[2]{%
  \begin{minipage}[t]{0.49\textwidth}
    \centering
    \includegraphics[width=\linewidth,height=0.205\textheight,keepaspectratio]{#1}\par
  \end{minipage}%
}
\noindent\rule{\textwidth}{0.6pt}
\vspace{0.2ex}
\begin{center}
\setlength{\tabcolsep}{2pt}%
\renewcommand{\arraystretch}{1.0}%
\begin{tabular}{@{}c@{\hspace{1pt}}c@{\hspace{4pt}\vrule width 0.6pt\hspace{4pt}}c@{}}
  & \small \textcolor{purple}{\shortstack{Three deer.
}}
  & \small \textcolor{purple}{\shortstack{Two breads and two strawberries.}} \\[0.2ex]
  \raisebox{9.5ex}{\textbf{\scriptsize DTS}}
  & \methodrowpanel{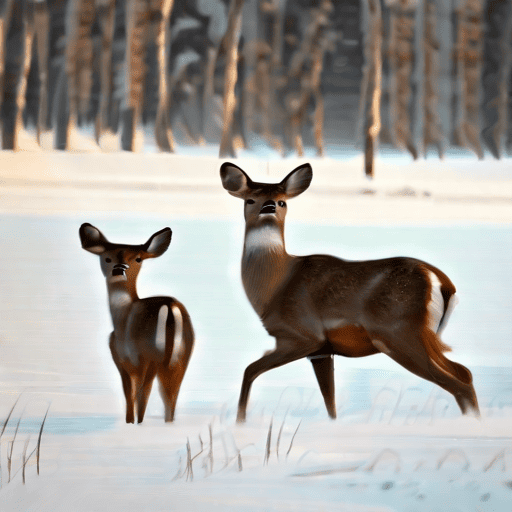}{}
  & \methodrowpanel{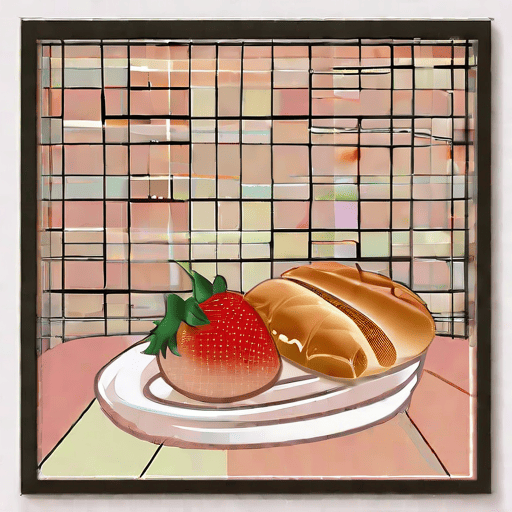}{} \\[0.25ex]
  \raisebox{9.5ex}{\textbf{\scriptsize BFMT}}
  & \methodrowpanel{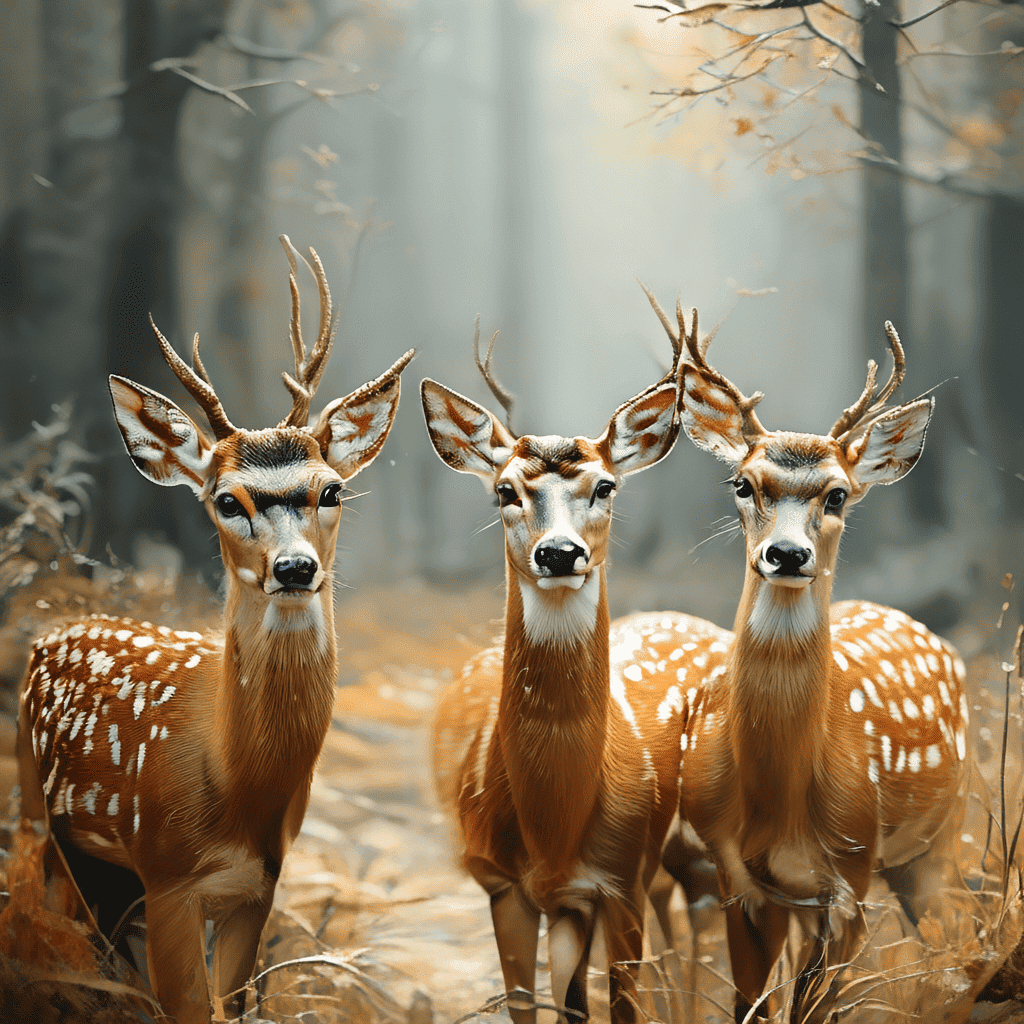}{}
  & \methodrowpanel{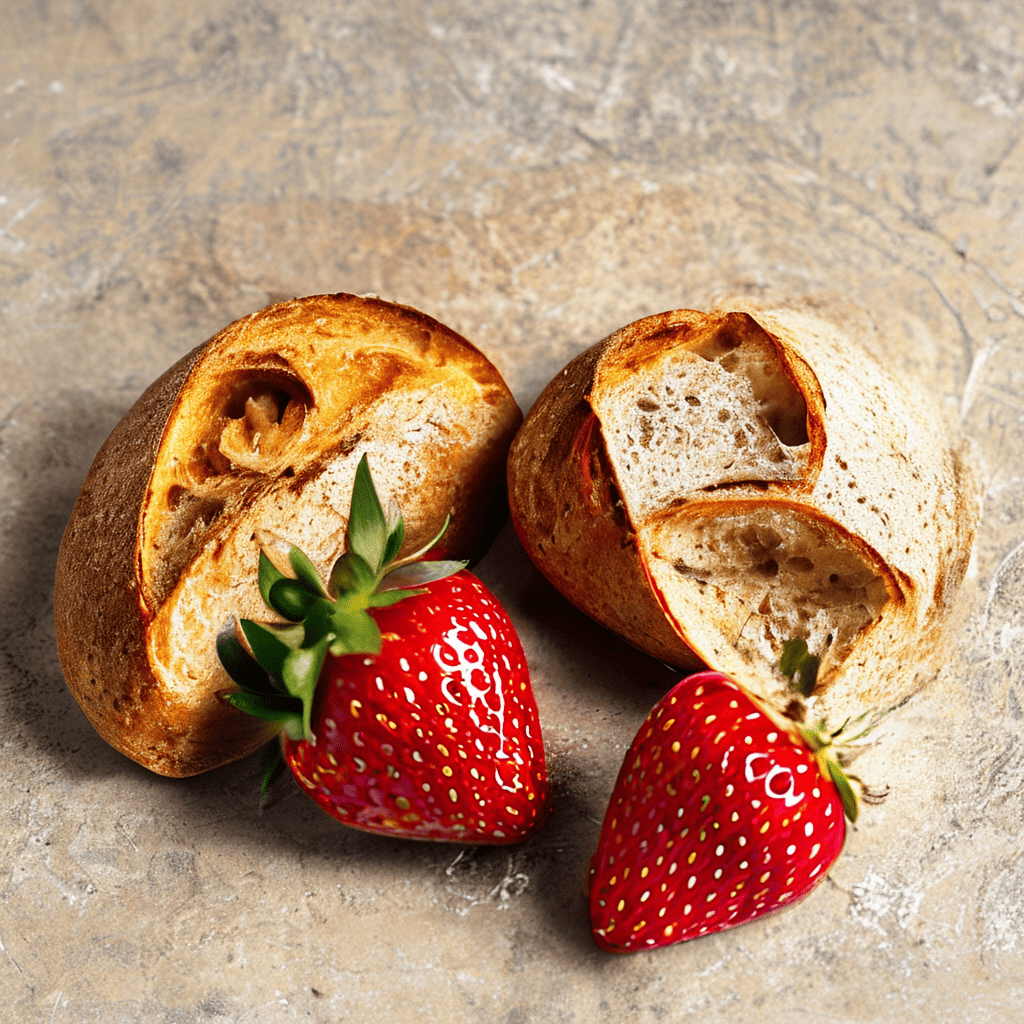}{} \\[0.35ex]
  \multicolumn{3}{@{}c@{}}{\rule{\textwidth}{0.8pt}} \\[0.35ex]
  & \small \textcolor{purple}{\shortstack{A ghostly ship sailing on \\a fog-shrouded, moonlit sea.}}
  & \small \textcolor{purple}{\shortstack{A phoenix soaring above a city, \\aglow with golden flames.}} \\[0.2ex]
  \raisebox{9.5ex}{\textbf{\scriptsize DTS}}
  & \methodrowpanel{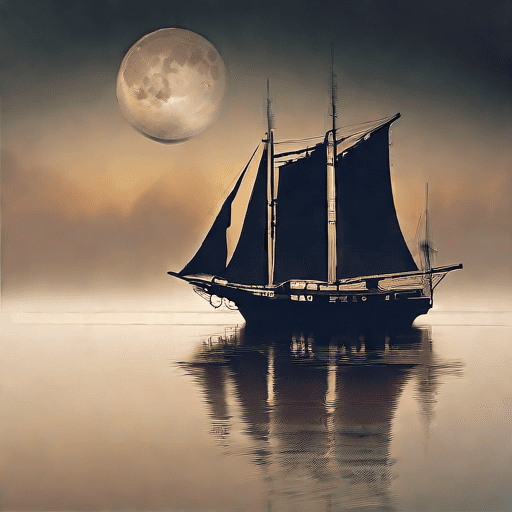}{}
  & \methodrowpanel{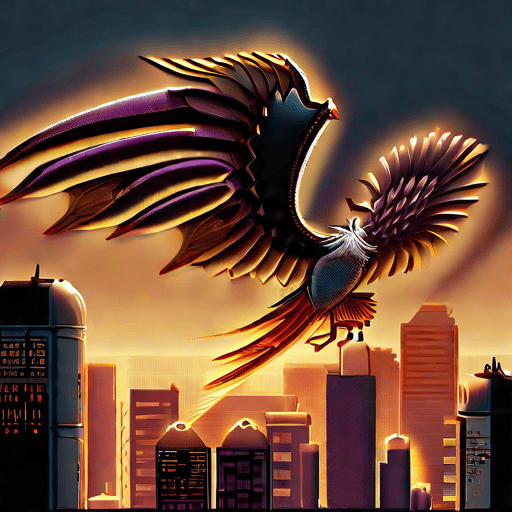}{} \\[0.25ex]
  \raisebox{9.5ex}{\textbf{\scriptsize BFMT}}
  & \methodrowpanel{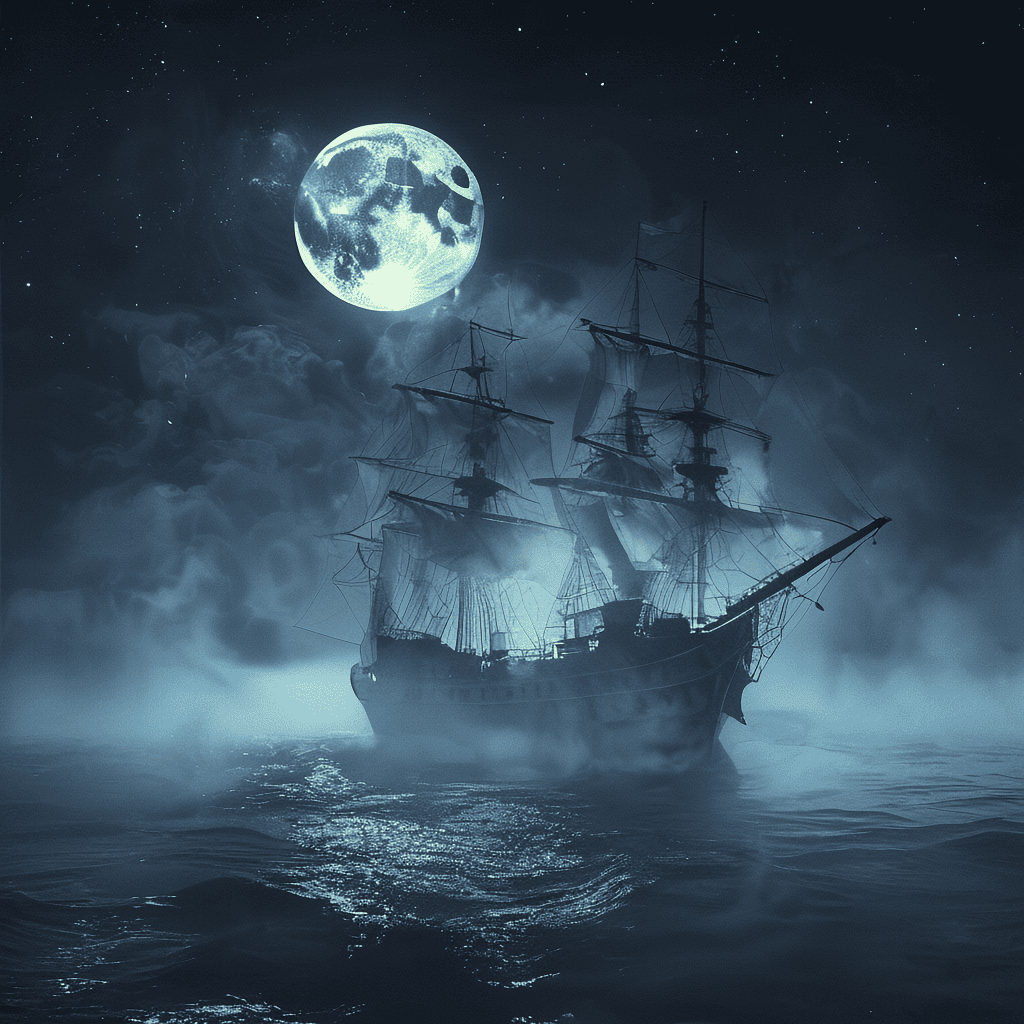}{}
  & \methodrowpanel{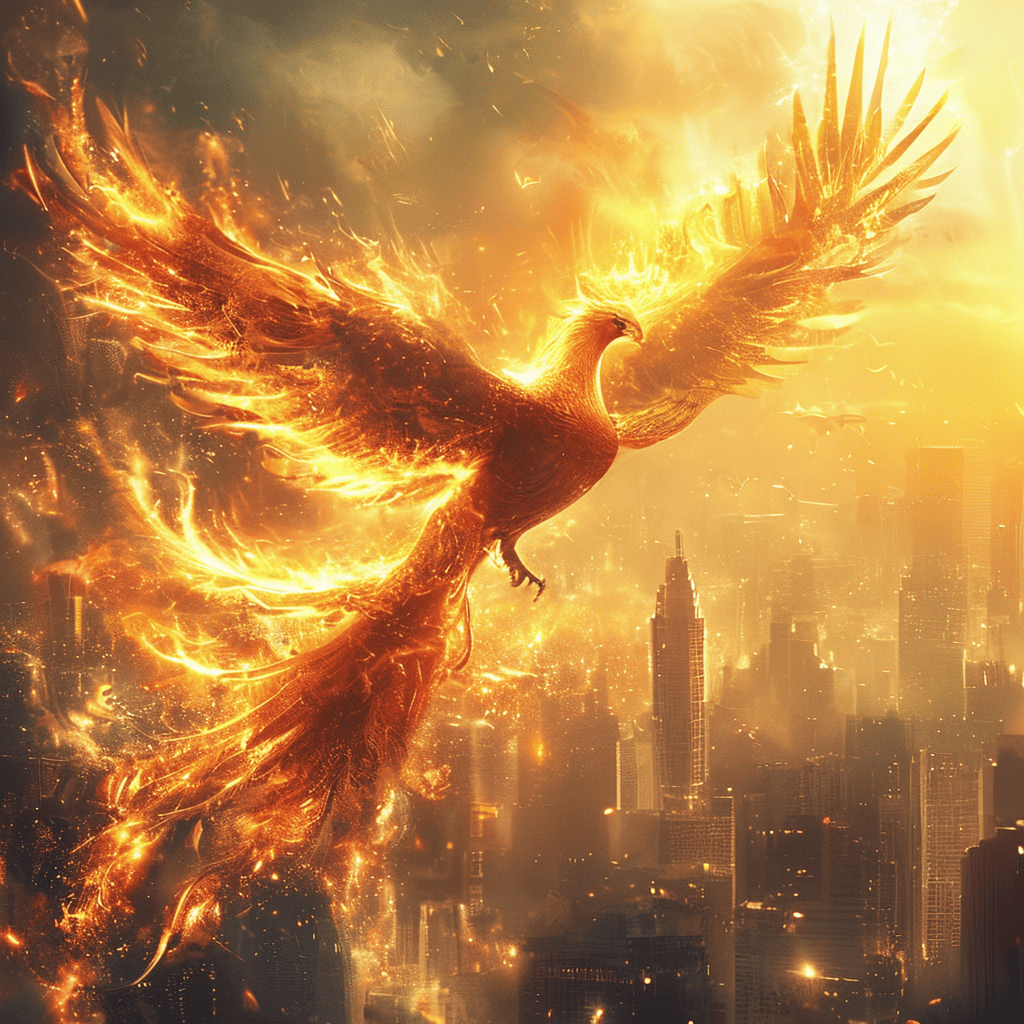}{}
\end{tabular}
\end{center}
\vspace{-0.2ex}
\noindent\rule{\textwidth}{0.6pt}
\caption{\small{Additional Alignment Visualization. }}
\label{fig:align-visu3-ADD}
\end{figure*}

\begin{figure*}[h]
\centering
\newcommand{\methodrowpanel}[2]{%
  \begin{minipage}[t]{0.49\textwidth}
    \centering
    \includegraphics[width=\linewidth,height=0.165\textheight,keepaspectratio]{#1}\par
  \end{minipage}%
}
\noindent\rule{\textwidth}{0.6pt}
\vspace{0.2ex}
\begin{center}
\setlength{\tabcolsep}{2pt}%
\renewcommand{\arraystretch}{1.0}%
\begin{tabular}{@{}c@{\hspace{1pt}}c@{\hspace{4pt}\vrule width 0.6pt\hspace{4pt}}c@{}}
  & \small \textcolor{purple}{\shortstack{Three cups and one cow.}}
  & \small \textcolor{purple}{One plate and four bears.} \\[0.2ex]
  \raisebox{9.5ex}{\textbf{\scriptsize DTS}}
  & \methodrowpanel{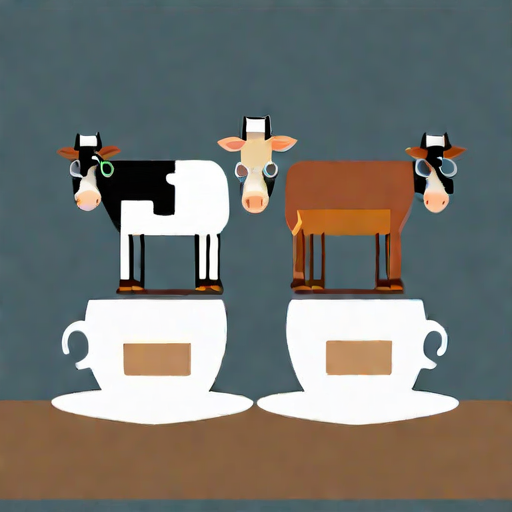}{}
  & \methodrowpanel{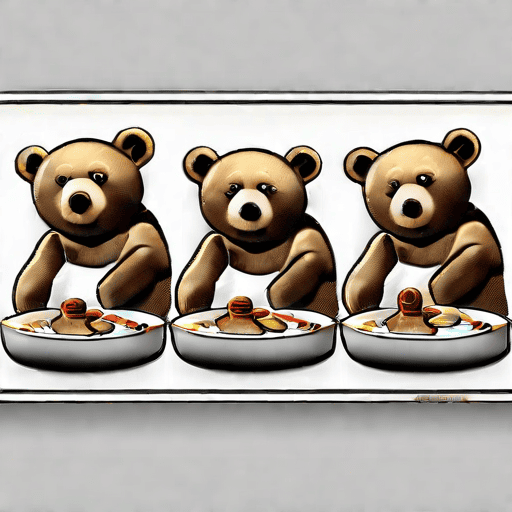}{} \\[0.25ex]
  \raisebox{9.5ex}{\textbf{\scriptsize BFMT}}
  & \methodrowpanel{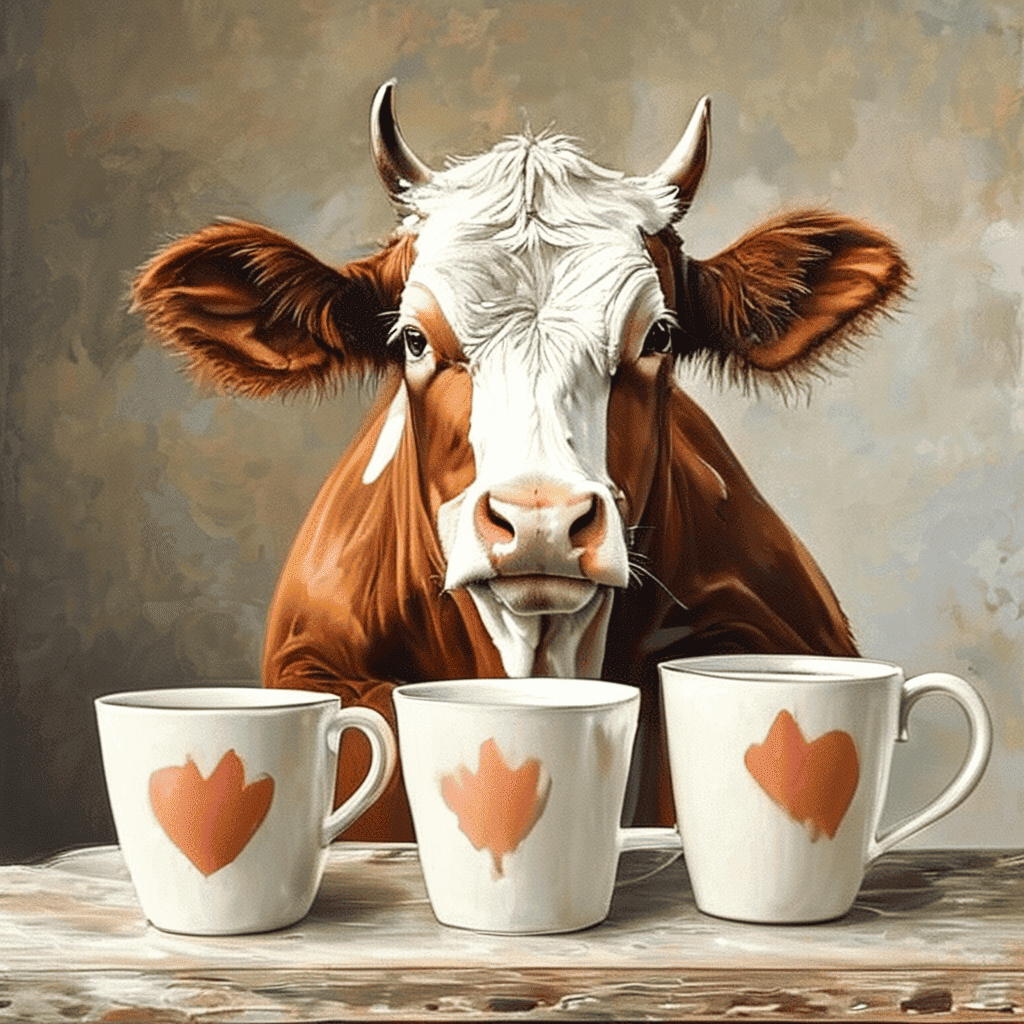}{}
  & \methodrowpanel{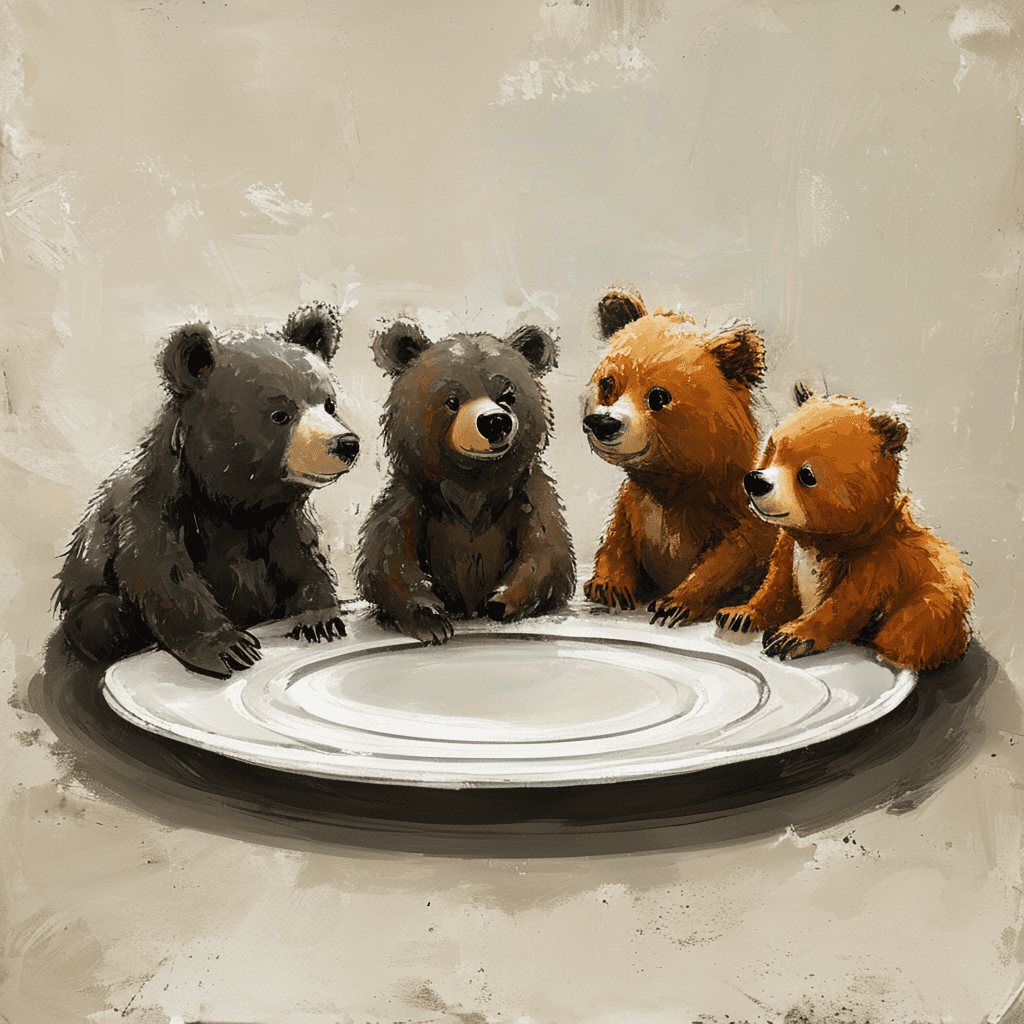}{}
\end{tabular}
\end{center}
\vspace{-0.2ex}
\noindent\rule{\textwidth}{0.6pt}
\caption{\small{Additional Alignment Visualization. }}
\label{fig:align-visunew-ADD}
\end{figure*}

\begin{figure*}[h]
\centering
\newcommand{\triplepanel}[3]{%
  \setlength{\tabcolsep}{2pt}%
  \begin{tabular}{@{}c@{\hspace{1pt}}c@{\hspace{1pt}}c@{}}
    \includegraphics[width=0.305\linewidth]{#1} &
    \includegraphics[width=0.305\linewidth]{#2} &
    \includegraphics[width=0.305\linewidth]{#3}
  \end{tabular}%
}
\newcommand{\panelblock}[4]{%
  \begin{minipage}[t]{\linewidth}
    \centering
    {\itshape ``#1''\par}
    \vspace{0pt}
    \triplepanel{#2}{#3}{#4}
  \end{minipage}%
}
\newcommand{\topheaders}{%
  {\fontsize{9.5}{10.5}\selectfont
  \begin{tabular}{@{}c@{\hspace{1pt}}c@{\hspace{1pt}}c@{}}
    \makebox[0.305\linewidth][c]{\textbf{DTS}} &
    \makebox[0.305\linewidth][c]{\textbf{FKS}} &
    \makebox[0.305\linewidth][c]{\textbf{BFMT}}
  \end{tabular}%
  }
}
\noindent
\begin{minipage}[t]{0.48\textwidth}
\centering
\topheaders
\end{minipage}\hfill%
\begin{minipage}[t]{0.48\textwidth}
\centering
\topheaders
\end{minipage}
\vspace{-0.8ex}
\noindent\rule{\textwidth}{0.6pt}
\vspace{-1.2ex}
\begin{minipage}[t]{0.48\textwidth}
\centering
\panelblock{Two people and \\ two helicopters.}{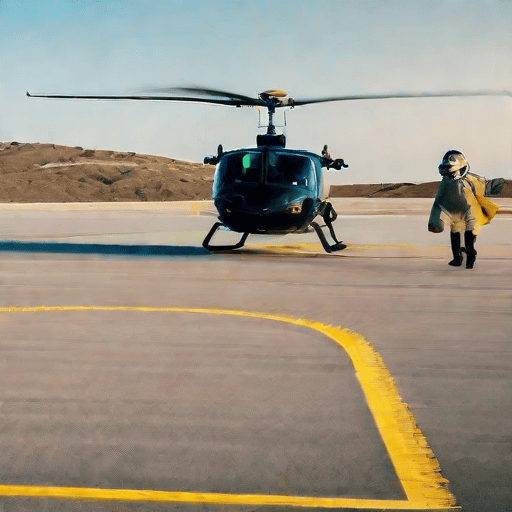}{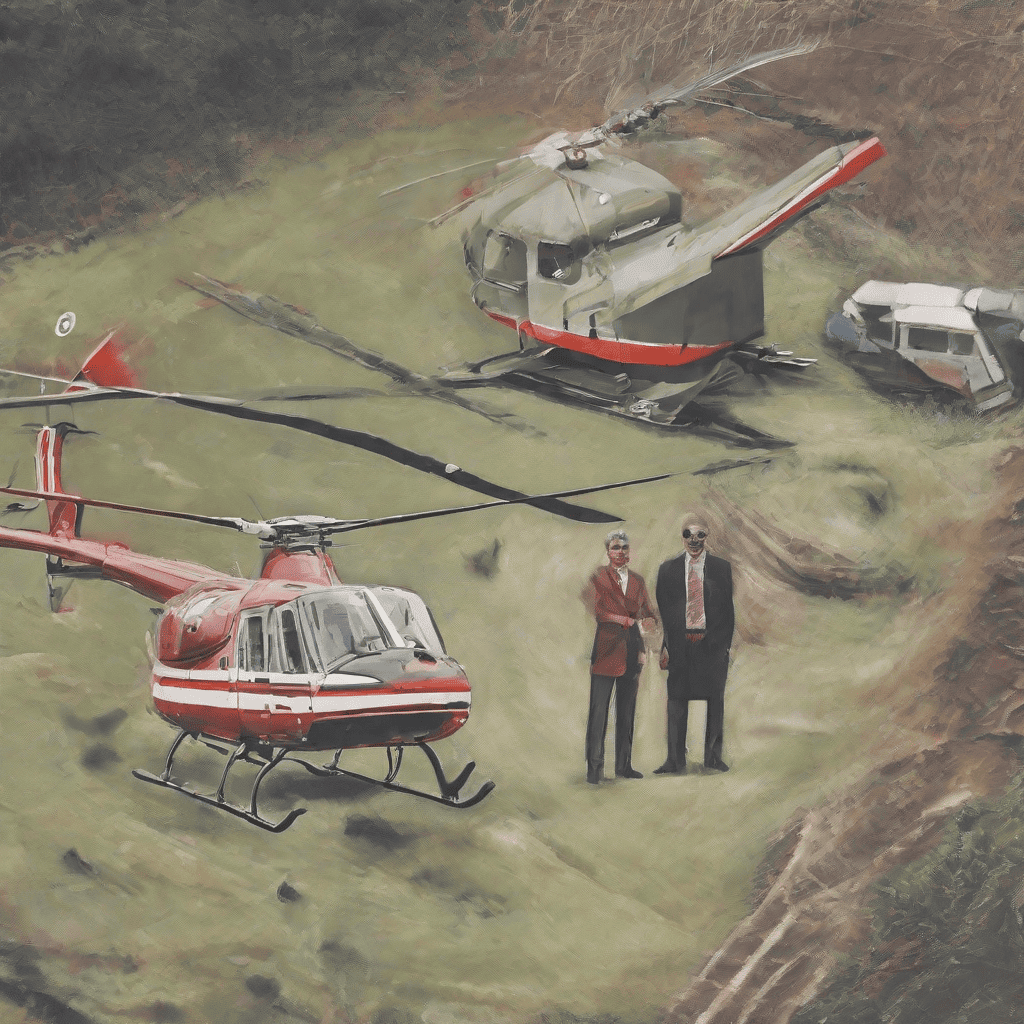}{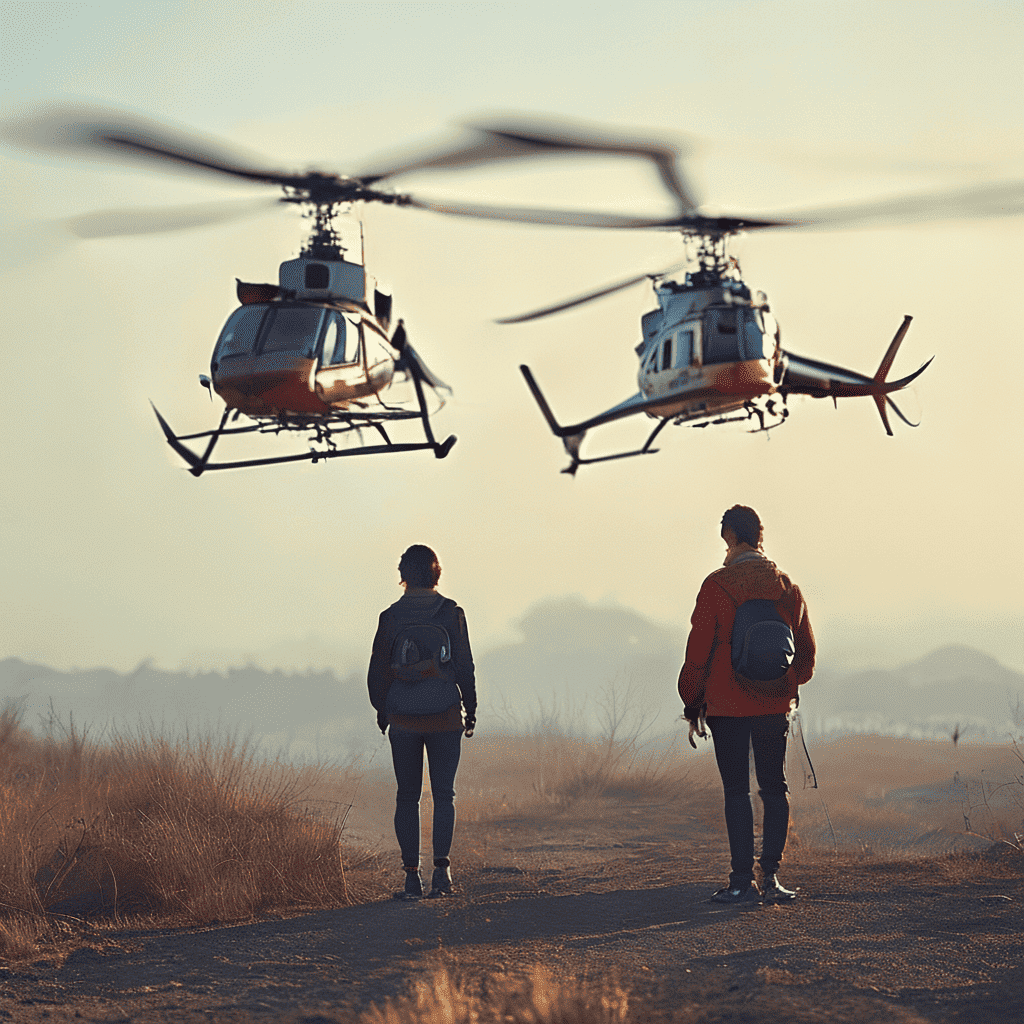}
\end{minipage}\hfill%
\begin{minipage}[t]{0.48\textwidth}
\centering
\panelblock{A dragon perched majestically on a craggy, smoke-wreathed mountain.}{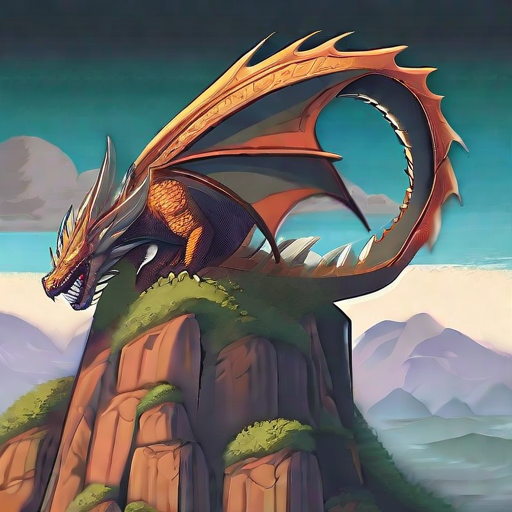}{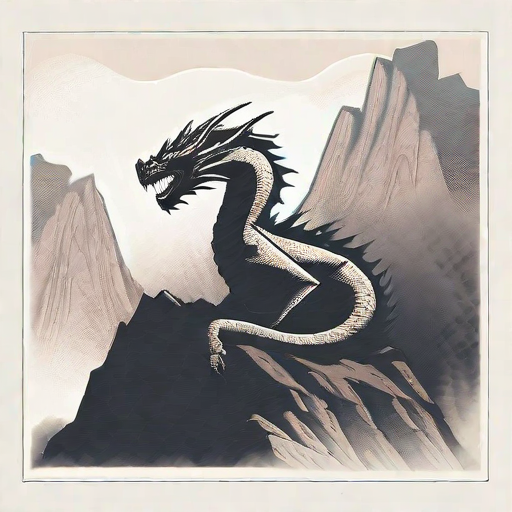}{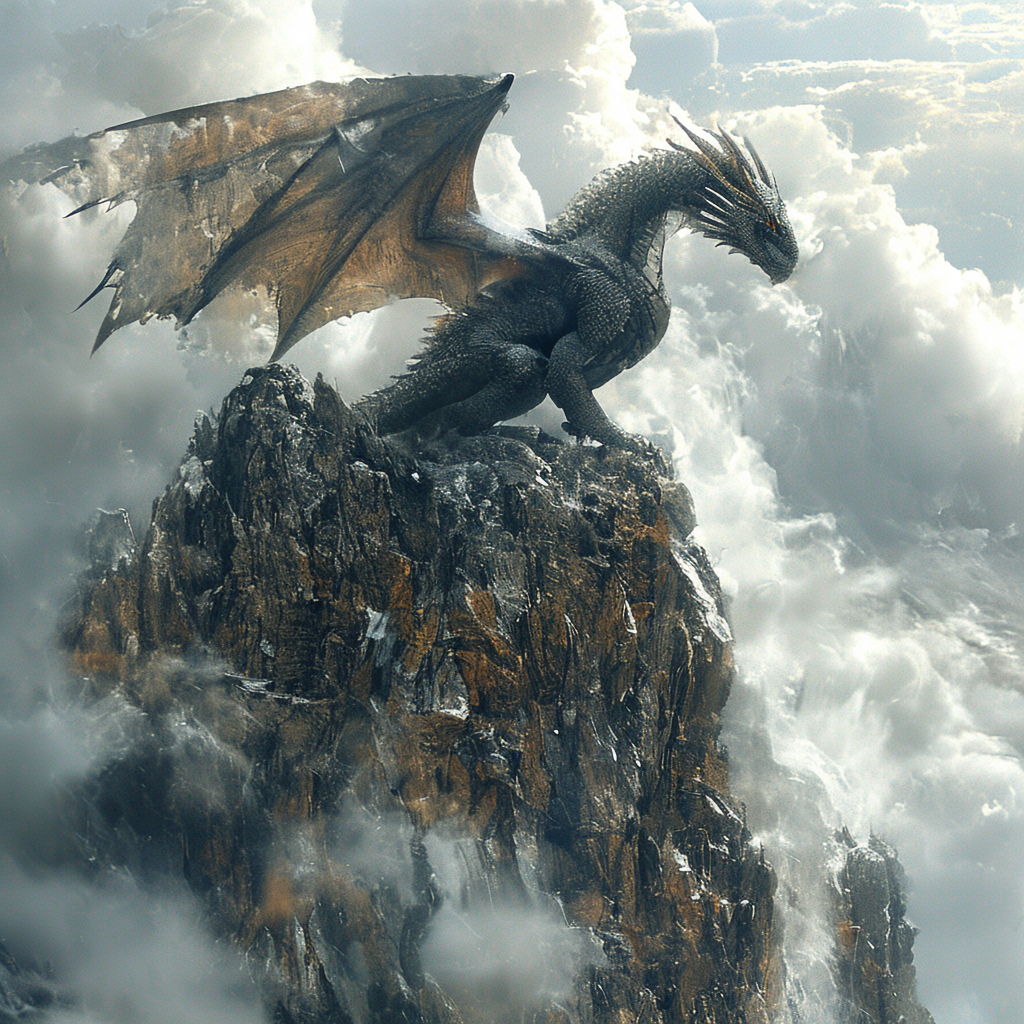}
\end{minipage}%
\vspace{0.6ex}
\noindent\rule{\textwidth}{0.6pt}
\vspace{-0.25ex}
\vspace{-0.45ex}
\noindent\rule{\textwidth}{0.6pt}
\caption{\small{Alignment Visualization. (left) Quantity Prompt. (right) Compositional Prompt. }}
\label{fig:align-vis-add1}
\vspace{-9pt}
\end{figure*}

Figures~\ref{fig:align-vis-add1}, ~\ref{fig:align-visu-add1},~\ref{fig:align-visu3-ADD},~\ref{fig:align-visunew-ADD} present additional qualitative comparisons in the online feedback-driven alignment setting, further validating the effectiveness of BFMT over competitive baselines.

\end{document}